\documentclass{article}
\usepackage{microtype}
\usepackage{graphicx}
\usepackage{booktabs}
\usepackage{hyperref}
\usepackage{multirow}

\usepackage[accepted]{icml2020}
\usepackage{chngcntr}
\usepackage{apptools}
\AtAppendix{\counterwithin{lemma}{section}}
\usepackage{url}  
\usepackage{graphicx}  
\usepackage{amsmath,amsfonts}
\usepackage{bm}
\usepackage{mathtools}
\usepackage{amssymb}
\usepackage{enumitem}
\usepackage{nicefrac}       
\usepackage{microtype}      
\usepackage{amsmath}
\usepackage{mathtools}
\usepackage{ntheorem}
\usepackage{multirow}
\usepackage{cleveref}
\usepackage{color}
\usepackage{comment}
\usepackage[subrefformat=parens,labelformat=parens]{subfig}
\usepackage{bm}
\usepackage[title]{appendix}
\newtheorem{definition}{Definition}
\newtheorem{lemma}{Lemma}
\newtheorem{remark}{Remark}
\newtheorem{corollary}{Corollary}

\newtheorem{prop}{Proposition}

\newtheorem*{myproof}{Proof}
\newcommand{\proofname}{Proof}
\DeclareMathOperator*{\argmax}{argmax} 
\DeclareMathOperator{\E}{\mathbb{E}}
\DeclareMathOperator{\Prob}{\mathbb{P}}
\newcommand{\inv}{^{\raisebox{.2ex}{$\scriptscriptstyle-1$}}}
\newcommand{\pch}[1]{\textcolor{black}{#1}}
\newcommand*{\QED}{\hfill\ensuremath{\square}}%

\makeatletter
\DeclareRobustCommand{\qed}{%
  \ifmmode 
  \else \leavevmode\unskip\penalty9999 \hbox{}\nobreak\hfill
  \fi
  \quad\hbox{\qedsymbol}}
\newcommand{\openbox}{\leavevmode
  \hbox to.77778em{%
  \hfil\vrule
  \vbox to.675em{\hrule width.6em\vfil\hrule}%
  \vrule\hfil}}
\makeatletter
\newcommand{\printfnsymbol}[1]{%
  \textsuperscript{\@fnsymbol{#1}}%
}

\newcommand*\xbar[1]{%
   \hbox{%
     \vbox{%
       \hrule height 0.5pt 
       \kern0.5ex
       \hbox{%
         \kern-0.1em
         \ensuremath{#1}%
         \kern-0.1em
       }%
     }%
   }%
}
\newcount\colveccount
\newcommand*\colvec[1]{
        \global\colveccount#1
        \begin{pmatrix}
        \colvecnext
}
\def\colvecnext#1{
        #1
        \global\advance\colveccount-1
        \ifnum\colveccount>0
                \\
                \expandafter\colvecnext
        \else
                \end{pmatrix}
        \fi
}

\icmltitlerunning{Exploration Through Reward Biasing: Reward-Biased Maximum Likelihood Estimation for Stochastic Multi-Armed Bandits}

\begin{document}
\twocolumn[
\icmltitle{Exploration Through Reward Biasing: Reward-Biased Maximum Likelihood Estimation for Stochastic Multi-Armed Bandits}
\icmlsetsymbol{equal}{*}
\begin{icmlauthorlist}
\icmlauthor{Xi Liu}{equal,to}
\icmlauthor{Ping-Chun Hsieh}{equal,nctu}
\icmlauthor{Yu-Heng Hung}{nctu}
\icmlauthor{Anirban Bhattacharya}{goo}
\icmlauthor{P. R. Kumar}{to}
\end{icmlauthorlist}
\icmlaffiliation{to}{Department of Electrical and Computer Engineering, Texas A\&M University, College Station, Texas, USA}
\icmlaffiliation{goo}{Department of Statistics, Texas A\&M University, College Station, Texas, USA}
\icmlaffiliation{nctu}{Department of Computer Science, National Chiao Tung University, Hsinchu, Taiwan}
\icmlcorrespondingauthor{Ping-Chun Hsieh}{pinghsieh@nctu.edu.tw}
\icmlcorrespondingauthor{Xi Liu}{xiliu.tamu@gmail.com}
\icmlkeywords{Bandit Learning, Reinforcement Learning, Online Learning, Maximum Likelihood Estimation, Regularization}

\vskip 0.3in
]
\printAffiliationsAndNotice{\icmlEqualContribution}
\begin{abstract}
Inspired by the Reward-Biased Maximum Likelihood Estimate method of adaptive control, we propose RBMLE -- a novel family of learning algorithms for stochastic multi-armed bandits (SMABs). 
For a broad range of SMABs including both the \emph{parametric} Exponential Family as well as the \emph{non-parametric} sub-Gaussian/Exponential family, we show that RBMLE yields an index policy.
To choose the bias-growth rate $\alpha(t)$ in RBMLE, we reveal the nontrivial interplay between $\alpha(t)$ and the regret bound that generally applies in both the Exponential Family as well as the sub-Gaussian/Exponential family bandits. To quantify the finite-time performance, we prove that RBMLE attains \emph{order-optimality} by adaptively estimating the unknown constants in the expression of $\alpha(t)$ for Gaussian and sub-Gaussian bandits.
Extensive experiments demonstrate that the proposed RBMLE achieves empirical regret performance competitive with the state-of-the-art methods, while being more computationally efficient and scalable in comparison to the best-performing ones among them.
\end{abstract}

\section{Introduction}
\label{section:intro}

Controlling an unknown system to maximize long-term average reward is a well studied adaptive control problem \cite{kumar1985survey}. For unknown Markov Decision Processes (MDPs), \citet{mandl1974estimation} proposed the certainty equivalent (CE) method that at each stage makes a maximum likelihood estimate (MLE) of the unknown system parameters and then applies the action optimal for that estimate. Specifically, consider an MDP with state space $\mathcal{S}$, action space $\mathcal{U}$, and controlled transition probabilities $p(i,j,u;\bm{\eta})$ denoting the probability of transition to a next state $s(t+1)=j$ when the current state $s(t)=i$ and action $u(t)=u$ is applied at time $t$, indexed by a parameter $\bm{\eta}$ in a set $\Xi$. The true parameter is $\bm{\eta}^0 \in \Xi$, but is unknown. A reward $r(i,j,u)$ is obtained when the system transitions from $i$ to $j$ under $u$. Consider the goal of maximizing the long-term average reward $\mbox{lim}_{T \to \infty} \frac{1}{T} \sum_{t=0}^{T-1} r(s(t),s(t+1),u(t))$.
Let $J(\phi,{\bm{\eta}})$ denote the long-term average reward accrued by the stationary control law $\phi: \mathcal{S} \rightarrow \mathcal{U}$
which chooses the action $u(t) = \phi(s(t))$. Let $J_{opt}({\bm{\eta}}) := \max_{\phi}J(\phi,{\bm{\eta}}) = J(\phi_{\bm{\eta}},{\bm{\eta}})   $ be the optimal long-term reward, and $\phi_{\bm{\eta}}$  an optimal control law, if the true parameter is ${\bm{\eta}}$. Denote by $\widehat{{\bm{\eta}}}_t \in \argmax_{\bm{\eta}\in\Xi}\prod_{\tau=0}^{t-1}p(s(\tau),s(\tau+1),u(\tau);{\bm{\eta}})$ a MLE of the true parameter ${\bm{\eta}}^0$. Under the CE approach, the action taken at time $t$ is $u(t)=\phi_{\widehat{{\bm{\eta}}}_t}(s(t))$. This approach was shown to be sub-optimal by \citet{borkar1979adaptive} since it suffers from the ``closed-loop identifiability problem'': the parameter estimates $\widehat{{\bm{\eta}}}_t$ converge to a ${\bm{\eta}}^*$ for which it can only be guaranteed that:
\begin{equation}
p(i,j,\phi_{{\bm{\eta}}^*}(i);{\bm{\eta}}^*)=p(i,j,\phi_{{\bm{\eta}}^*}(i);{\bm{\eta}}^0) \mbox{ for all }i,j, \label{closed-loop identifiability}
\end{equation}
and, in general, $\phi_{{\bm{\eta}}^*}$ need not be optimal for ${\bm{\eta}}^0$.

To solve this fundamental problem, \citet{kumar1982new} noticed that due to (\ref{closed-loop identifiability}),
$J(\phi_{{\bm{\eta}}^*}, {\bm{\eta}}^*)=J(\phi_{{\bm{\eta}}^*}, {\bm{\eta}}^0)$. Since $J(\phi_{{\bm{\eta}}^*},\bm{\eta}^0) \leq J_{opt}({\bm{\eta}^0})$, but $J(\phi_{{\bm{\eta}}^*}, {\bm{\eta}}^*)=J_{opt}({\bm{\eta}}^*)$ due to
$\phi_{\bm{\eta}}^*$ being optimal for $\bm{\eta}^*$,  it follows that
$J_{opt}({\bm{\eta}}^*) \leq J_{opt}({\bm{\eta}}^0)$, i.e., the parameter estimates are biased in favor of parameters with smaller optimal reward. To undo this bias, with $f$ denoting any strictly monotone increasing fucntion, they suggested employing the RBMLE estimate:
\begin{align}
    \widehat{\bm{\eta}}^{\text{RBMLE}}_t = &\argmax_{\bm{\eta} \in \Xi} \nonumber\\ 
    &f(J_{opt}(\bm{\eta}))^{\alpha(t)}\prod_{\tau=0}^{t-1}p(s(\tau),s(\tau+1),u(\tau),{\bm{\eta}})
    ,\label{equation:general_bmle}
\end{align}
in the CE scheme, with $u(t)=\phi_{\widehat{\bm{\eta}}^{\text{RBMLE}}_t} (s(t))$. In (\ref{equation:general_bmle}), $\alpha(t):[1,\infty)\rightarrow \mathbb{R}_{+}$ is allowed to be any function that satisfies $\lim_{t\rightarrow \infty} \alpha(t)=\infty$ and $\lim_{t\rightarrow \infty}$ $\alpha(t)/t=0$. This method was shown to yield the optimal long-term average reward in a variety of settings
\cite{kumar1982optimal,kumar1983simultaneous,kumar1983optimal,borkar1990kumar,borkar1991self,stettner1993nearly,duncan1994almost,campi1998adaptive,prandini2000adaptive}. For the case of Bernoulli bandits, it was shown in \citet{BeckerKumar81b} that the RBMLE approach provides an index policy where each arm has a simple index, and the policy is to just play the arm with the largest index.  

The structure of (\ref{equation:general_bmle}) has a few critical properties that contribute the success of RBMLE. First, the bias term $J_{opt}(\bm{\eta})^{\alpha(t)}$ multiplying the likelihood encourages active exploration by favoring $\bm{\eta}$s with potentially higher maximal long-term rewards. Second, the effect of the bias term gradually diminishes as $\alpha(t)$ grows like $o(t)$, which makes the exploitation dominate the estimation at later stages. 

Several critical questions need to be answered to tailor the RBMLE to the stochastic multi-armed bandits (SMABs).
\begin{enumerate}[wide]
\item
The average reward optimality proved in prior RBMLE studies is a gross measure implying only that regret (defined below) is $o(T)$, while in SMABs a much stronger $O(\log(T))$ \emph{finite-time} order-optimal regret is desired. 
What is the regret bound of the RBMLE algorithms in SMABs?
\item
The above options for $\alpha(t)$ are very broad, and not all of them lead to order-optimality of regret.
How should one choose the function $\alpha(t)$ to attain order-optimality?  
\item
What are the advantages of RBMLE algorithms compared to the existing methods? Recall that the Upper Confidence Bounds (UCB) approach pioneered by \cite{lai1985asymptotically}, and streamlined by \cite{auer2002finite}, is conductive to establish the regret bound, but suffers from much higher empirical regret than its counterparts, while the Information Directed Sampling (IDS) \cite{russo2014learning, russo2017learning} approach appears to achieve the smallest empirical regret in various bandits, but suffers from high computational complexity with resulting poor scalability due to the calculation or estimation of high dimensional integrals. 
\end{enumerate}
The major contributions of this paper are:
\begin{enumerate}[wide]
    \item We show that RBMLE yields an ``index policy" for Exponential Family SMABs, and explicitly determine the indices. (An index policy is one where each arm has an index that depends only on its own past performance history, and one simply plays the policy with the highest index).  We also propose novel RBMLE learning algorithms for SMABs from sub-Gaussian/Exponential non-parametric families.
    \item We reveal the general interplay between the choice of $\alpha(t)$ and the regret bound. When a lower bound on the ``minimum gap,'' the difference between the means of the best and second best arms, and an upper bound on the maximal mean reward are known, simple closed-form indices as well as $O(\log (T))$ order-optimal regret are achieved for reward distributions for both parametric Exponential families as well as sub-Gaussian/Exponential non-parametric families. 
    \item When the two bounds are unknown, the proposed RBMLE algorithms still attain order-optimality in the Gaussian and sub-Gaussian cases by adaptively estimating them in the index expressions on the fly.
    \item We evaluate the empirical performance of RBMLE algorithms in extensive experiments. They demonstrate competitive performance in regret as well as scalability against the current best policies.
\end{enumerate}
\section{Problem Setup} \label{section:formulation}
Consider an $N$-armed bandit, where each arm $i$ is characterized by its reward distribution $\mathcal{D}_i$ with mean {$\theta_i\in\Theta$}, where $\Theta$ denotes the set of possible values for the mean rewards.
Without loss of generality, let $\theta_1>\theta_2\geq \cdots \geq \theta_N\geq 0$.
For each arm $i$, let $\Delta_i:=\theta_1 - \theta_i$ be the gap between its mean reward and that of the optimal arm, and
 $\Delta:=\Delta_2$ the ``minimum gap.''
Let $\bm{\theta}$ denote the vector $(\theta_1,\cdots, \theta_N)$. 
At each time $t=1,\cdots, T$, the decision maker chooses an arm $\pi_t\in [N] :=\{1,\cdots, N\}$ and obtains the corresponding reward $X_t$, which is independently drawn from the distribution $\mathcal{D}_{\pi_t}$.
Let $N_i(t)$ and $S_i(t)$ be the total number of trials of arm $i$ and the total reward collected from pulls of arm $i$ up to time $t$, respectively.
Define $p_i(t):=S_i(t)/N_i(t)$ as the empirical mean reward up to $t$.
Denote by $\mathcal{H}_t=(\pi_1,X_1,\pi_2,X_2,\cdots,\pi_t, X_t)$ the history of all the choices of the decision maker and the reward observations up to time $t$.
Let $L(\mathcal{H}_t; \{\mathcal{D}_i\})$ denote the likelihood of the history $\mathcal{H}_t$ under the reward distributions $\{\mathcal{D}_i\}$. 
The objective is to minimize the \emph{regret} defined as $\mathcal{R}(T):=T\theta_1 - \E[\sum_{t=1}^{T}X_t]$, where the expectation is taken with respect to the randomness of the rewards and the employed policy.
\section{The RBMLE Policy for Exponential Families and its Indexability} \label{section:exponential} \label{section:alg}


Let the probability density function of the reward obtained from
arm $i$ be a one-parameter Exponential Family distribution:
\begin{equation}
   p(x;\eta_i)=A(x)\exp\big(\eta_i x -F(\eta_i)\big), \hspace{3pt}
   \label{equation:exponential family}
\end{equation}
where $\eta_i \in \mathcal{N}$ is the canonical parameter, $\mathcal{N}$ is the parameter space, $A(\cdot)$ is a real-valued function, and $F(\cdot)$ is a real-valued twice-differentiable function.
Then (see \cite{JordanExponentialFamily2010}), $\theta_i =\dot{F}(\eta_i)$
(``dot'' denoting derivative) is the mean of the reward distribution, and  its
variance is $\ddot{F}(\eta_i)$.  
Also,
(i) $F(\cdot)$ is strictly \emph{convex}, and hence (ii) the function $\dot{F}(\eta)$ is strictly \emph{increasing}, hence invertible. 
A critical property that will be used to derive the RBMLE
is that $\eta_i = \dot{F}\inv(\theta_i)$ is strictly monotone increasing in the mean reward $\theta_i$.

We consider the case where the reward distribution of each arm $i$ has the density function $p(\cdot;\eta_i)$, with $F(\cdot)$ and $A(\cdot)$ being identical across all the arms $1 \leq i \leq N$.
Let $\eta := (\eta_1, \eta_2, \ldots , \eta_N)$ denote the parameter vector
that collectively describes the set of all arms. Based on (\ref{equation:exponential family}), if
$X_\tau$ is the reward obtained at time $\tau$ by pulling arm $\pi_\tau$,
then
the likelihood of $\mathcal{H}_t$ at time $t$ under the parameter vector $\bm{\eta}$ is
\begin{equation}
    L(\mathcal{H}_t;\bm{\eta})=\prod_{\tau=1}^{t}A(X_\tau)\exp\big(\eta_{\pi_\tau}X_\tau-F(\eta_{\pi_\tau})\big).\label{equation:likelihood exp}
\end{equation}
Now let us consider the reward bias term $f(J_{opt}(\bm{\eta}))^{\alpha(t)}$ in (\ref{equation:general_bmle}).
The optimal reward obtainable from $\bm{\eta}$ is the maximum of the mean rewards of the arms, i.e.,
 $J_{opt}(\bm{\eta}) = {\max}_{1 \leq i \leq N} \theta_i$,
where $\theta_i =\dot{F}(\eta_i)$ is the mean reward from arm $i$.
By choosing the strictly monotone increasing function $f(\cdot)=\exp(\dot{F}\inv(\cdot))$,
the reward-bias term reduces to 
\begin{align}
    f(J_{opt}(\bm{\eta}))^{\alpha(t)}:=\max_{i\in [N]}\big\{\exp(\eta_i\alpha(t)) \big\}.\label{reward-bias_term}
\end{align}
Multiplying the reward-bias term and the likelihood term, the RBMLE estimator is
\begin{align}
    &\widehat{\bm{\eta}}_t^{\text{RBMLE}}:=\argmax_{\bm{\eta}:\eta_j\in\mathcal{N},\forall j} \big\{L(\mathcal{H}_t;\bm{\eta}) \max_{i \in [N]}\exp(\eta_i \alpha (t))\big\}.\label{equation:BMLE for exp} 
\end{align}
The corresponding arm chosen by RBMLE at time $t$ is
\begin{align}
    \pi^{\text{RBMLE}}_t=\argmax_{i\in [N]}\big\{\widehat{\eta}^{\text{RBMLE}}_{t,i}\big\}.\label{equation:exchange theta and eta}
\end{align}
By combining (\ref{equation:exchange theta and eta}) with (\ref{equation:BMLE for exp}), we have the following index strategy equivalent to (\ref{equation:exchange theta and eta}):
\begin{align}
    \pi^{\text{RBMLE}}_t 
    &=\argmax_{i\in [N]}\Big\{\max_{\bm{\eta}:{\eta}_j\in\mathcal{N},\forall j}\big\{L(\mathcal{H}_t;\bm{\eta}) \exp(\eta_i\alpha (t))\big\}\Big\}.\label{equation:BMLE for exp 2-1}
\end{align}
The proof of the above result is provided in Appendix \ref{appendix:indexability}.

\subsection{The RBMLE Indices for Exponential Families} \label{section:indices}
An index policy is one where each arm $i$ has an index $I_i(t)$
that is only a function of the history of that arm $i$ up to time
$t$,
and the policy is to simply play the arm with the largest index $I_i(t)$
at time $t$ \cite{whittle1980multi}.
For the RBMLE policy as written in (\ref{equation:BMLE for exp 2-1})
it is not quite obvious that it is an index policy since the term
$L(\mathcal{H}_t;\bm{\eta}) \exp(\eta_i\alpha (t))$ for arm
$i$ depends on the combined history $\mathcal{H}_t$ of all arms,
including arms other than $i$.
It was recognized in \cite{BeckerKumar81b}
that RBMLE yields an index policy for the case of Bernoulli bandits. The following proposition shows that RBMLE is indeed an index policy for the Exponential Family described above, 
i.e., RBMLE is ``indexable," and
explicitly identifies what the index of an arm is.\footnote{The proofs of all Lemmas, Corollaries and Propositions are provided in the Appendices.}
\begin{prop}
\label{prop:BMLE index for exponential families}
The arm selected at time $t$ by the RBMLE algorithm for Exponential Family rewards is
\begin{align}
    &\pi^{\text{RBMLE}}_t=\argmax_{i\in [N]}I(p_{i}(t), N_i(t), \alpha (t)), \mbox{   where} \\
    &I(\nu, n, \alpha (t))= \big(n\nu+{\alpha(t)}\big)\dot{F}\inv\big(\big[\nu+\frac{\alpha(t)}{n}\big]_{\Theta}\big)\label{equation:RBMLE index for exp 1}\\
    &\hspace{-0pt}-n\nu\dot{F}\inv(\nu)-nF\big(\dot{F}\inv\big(\big[\nu+\frac{\alpha(t)}{n}\big]_{\Theta}\big)\big)+nF\big(\dot{F}\inv(\nu)\big), \label{equation:RBMLE index for exp 2}
\end{align}
with $[\cdot]_{\Theta}$ denoting the clipped value within the set $\Theta$.
\end{prop}

\begin{remark}
\normalfont {The clipping 
ensures that the input of $\dot{F}\inv$ is within $[0,1]$, since, 
for example, under Bernoulli reward distributions, $p_i(t)+\alpha(t)/N_i(t)$ could be larger than $1$.}
\end{remark}
The indices for three commonly-studied distributions are provided below.

\label{section:alg:Bernoulli}
\begin{corollary}
\label{corollary:Bernoulli index}
For Bernoulli distributions, with $F(\eta)=\log(1+e^{\eta})$, $\dot{F}(\eta)=\frac{e^{\eta}}{1+e^{\eta}}$, $\dot{F}\inv(\theta)=\log(\frac{\theta}{1-\theta})$, $F(\dot{F}\inv(\theta))=\log(\frac{1}{1-\theta})$, and with $\tilde{p}_i(t):=\min\{p_i(t)+\alpha(t)/N_i(t), 1\}$,
the RBMLE index is
\begin{align}
    &I(p_i(t),N_i(t),\alpha(t))= \\
    &N_i(t)\big\{\tilde{p}_i(t)\log\tilde{p}_i(t)+(1-\tilde{p}_i(t))\log(1-\tilde{p}_i(t))\label{equation:Bernoulli index 3}\\
    &-p_i(t)\log(p_i(t))-(1-p_i(t))\log(1-p_i(t))\big\}.\label{equation:Bernoulli index 4}
\end{align}
\end{corollary}

\begin{corollary}
\label{corollary:bmle gaussian}
For Gaussian reward distributions with the same variance $\sigma^2$ across arms, $F(\eta_i)=\sigma^2\eta_i^2/2$, $\dot{F}(\eta_i)=\sigma^2\eta_i$, $\dot{F}\inv(\theta_i)=\theta_i/\sigma^2$, and $F(\dot{F}\inv(\theta_i))=\theta_i^2/2\sigma^2$, for each arm $i$,
the RBMLE index is
\begin{equation}
I(p_i(t),N_i(t),\alpha(t)) = p_i(t)+\frac{\alpha (t)}{2 N_i(t)}.\label{equation:BMLE Gaussian in corollary}
\end{equation}
\end{corollary}

\begin{corollary}
\label{corollary:bmle exponential}
For Exponential distributions, 
the index is
\begin{equation}
I(p_i(t),N_i(t),\alpha(t)) = N_i(t)\log\big(\frac{N_i(t)p_i(t)}{N_i(t)p_i(t)+\alpha(t)}\big).
\end{equation}
\end{corollary}

\begin{remark}
\label{remark:applied to non-parametric}
\normalfont The RBMLE indices derived for parametric distributions can also be applied to \textit{non-parametric} distributions. 
As shown in Propositions~\ref{prop:BMLE regret for sub-Gaussian} and~\ref{prop:BMLE regret for sub-exponential}, they still achieve $O(\log(T))$ regret.
\end{remark}

\begin{table}[!htbp]
\caption{Comparison of indices produced by RBMLE with other approaches. Below, $H(p)$ is the binary entropy, $\overline{V}_{t}(i)$ is the upper bound on the variance, and the other quantities are defined in Sections \ref{section:formulation} and \ref{section:exponential}.}
\label{tab:my-table}
\begin{tabular}{|c|l|}
\hline
Algorithm             & \multicolumn{1}{c|}{Index}                                                      \\ \hline
\multirow{3}{*}{BMLE} & (Bernoulli) $N_i(t)\big( H(p_i(t))-H(\tilde{p}_i(t)\big)$                       \\ \cline{2-2} 
                      & (Gaussian) $p_i(t)+\alpha(t)/(2N_i(t))$                                         \\ \cline{2-2} 
                      & (Exponential) $N_i(t)\log\big(\frac{N_i(t)p_i(t)}{N_i(t)p_i(t)+\alpha(t)}\big)$ \\ \hline
UCB                   & $p_i(t)+\sqrt{{2\log t}/{N_i(t)}}$                                          \\ \hline
UCB-Tuned             & $p_i(t) + \sqrt{\min\{\frac{1}{4}, \overline{V}_{t}(i)\}\log(t)/N_i(t)\}}$      \\ \hline
MOSS                  & $p_i(t)+\sqrt{\max(\log(\frac{T}{N_i(t)\cdot N}),0)/N_i(t)}$                    \\ \hline
\end{tabular}
\end{table}

Table \ref{tab:my-table} compares the RBMLE indices with other policies. That these new indices have performance
competitive with state-of-the-art (Section \ref{section:simulation}), is of potential interest.

\subsection{Properties of the RBMLE Indices}\label{section:properties}
We introduce several useful properties of the index $I(\nu, n,\alpha(t))$ in (\ref{equation:RBMLE index for exp 1})-(\ref{equation:RBMLE index for exp 2}) to better understand the behavior of the derived RBMLE indices and prepare for regret analysis in subsequent sections.
To begin with, we discuss the dependence of $I(\nu,n,\alpha(t))$ on $\nu$ and $n$.

\begin{lemma}
\label{lemma:index changes with n}
(i) For a fixed $\nu\in \Theta$ and $\alpha(t)>0$, $I(\nu,n,\alpha(t))$ is strictly decreasing with $n$, for all $n>0$.

(ii) For a fixed $n>0$ and $\alpha(t)>0$, $I(\nu,n,\alpha(t))$ is strictly increasing with $\nu$, for all $\nu\in \Theta$.
\end{lemma}

Since the RBMLE index is $I(p_i(t), N_i(t),\alpha(t))$,  Lemma \ref{lemma:index changes with n}.(ii) suggests that the index of an arm increases with its empirical mean reward $p_i(t)$.

{To prepare for the following lemmas, we first define a function $\xi(k;\nu):\mathbb{R}_{++}\rightarrow \mathbb{R}$ as}
\begin{align}
    \xi(k;\nu)=&k\Big[\big(\nu+\frac{1}{k}\big)\dot{F}\inv(\nu+\frac{1}{k})-\nu\dot{F}\inv(\nu)\Big]\\
    &-k\Big[F\Big(\dot{F}\inv\big(\nu+\frac{1}{k}\big)\Big)-F(\dot{F}\inv(\nu))\Big].
\end{align}
It is easy to verify that $I(\nu, k\alpha(t),\alpha(t))=\alpha(t)\xi(k;\nu)$.
By Lemma \ref{lemma:index changes with n}.(i), we know $\xi(k;\nu)$ is strictly decreasing with $k$.
{Moreover, define a function $K^{*}(\theta',\theta'')$ as}
\begin{equation}
    K^{*}(\theta',\theta'')=\inf\{k:\dot{F}\inv(\theta')>\xi(k;\theta'')\}.\label{equation:K star definition}
\end{equation}
\begin{lemma}
\label{lemma:I(mu1,s1,alpha(t))>I(mu2,s2,alpha(t))}
\pch{Given any pair of real numbers $\mu_1,\mu_2\in \Theta$ with $\mu_1>\mu_2$, for any real numbers 
$n_1,n_2$ that satisfy $n_1>0$ and $n_2> K^{*}(\mu_1,\mu_2)\alpha(t)$ (with $K^{*}(\mu_1,\mu_2)$ being finite), we have $I(\mu_1, n_1, \alpha(t))> I(\mu_2, n_2, \alpha(t))$.}
\end{lemma}

\begin{lemma}
\label{lemma:I(0,s1,alpha(t))>I(mu2,s2,alpha(t))}
Given any real numbers $\mu_0, \mu_1,\mu_2\in \Theta$ with $\mu_0>\mu_1$ and $\mu_0>\mu_2$, for any real numbers $n_1,n_2$ that satisfy $n_1\leq K^*(\mu_0,\mu_1) \alpha (t)$ and $n_2> {K^*(\mu_0,\mu_2)\alpha (t)}$, we have $I(\mu_1, n_1, \alpha(t))> I(\mu_2, n_2, \alpha(t))$.
\end{lemma}

\begin{remark}
Lemmas \ref{lemma:I(mu1,s1,alpha(t))>I(mu2,s2,alpha(t))} and \ref{lemma:I(0,s1,alpha(t))>I(mu2,s2,alpha(t))} show how RBMLE naturally engages in the exploration vs. exploitation tradeoff. Lemma \ref{lemma:I(mu1,s1,alpha(t))>I(mu2,s2,alpha(t))} shows that RBMLE indeed tends to avoid an arm with a smaller empirical mean reward after sufficient exploration, as quantified in terms of $\alpha(t)$ by $n_2>K^*(\mu_1,\mu_2)\alpha(t)$.
On the other hand, Lemma \ref{lemma:I(0,s1,alpha(t))>I(mu2,s2,alpha(t))} suggests that RBMLE is designed to continue exploration even if the empirical mean reward is initially fairly low (which is reflected by the fact that there is no restriction on the ordering between $\mu_1$ and $\mu_2$ in Lemma \ref{lemma:I(0,s1,alpha(t))>I(mu2,s2,alpha(t))}), when there has been insufficient exploration, as quantified by $n_1\leq K^*(\mu_0,\mu_1)\alpha(t)$. These properties emerge naturally out of the reward biasing.
\end{remark}
\section{Regret Analysis of the RBMLE Algorithm}
\label{section:theory}

We now analyze the finite-time performance of the proposed RBMLE algorithm.
The KL divergence $\text{KL}(\eta'\,||\,\eta'')$ between two distributions can be expressed as
\begin{equation}
    \text{KL}(\eta'\,||\,\eta'')=F(\eta'')-[F(\eta')+\dot{F}(\eta')(\eta''-\eta')].
\end{equation}
Define $D(\theta',\theta''):\Theta\times\Theta\rightarrow \mathbb{R}_+$ by
\begin{equation}
    D(\theta',\theta''):=\text{KL}(\dot{F}\inv(\theta')\,||\,\dot{F}\inv(\theta'')).\label{equation:KL divergence using mean}
\end{equation}

\subsection{Interplay of Bias-Growth Rate $\alpha(t)$ and Regret} \label{first_question}
We determine the regret bounds for several classes of distributions, both parametric as well as non-parametric.
\subsubsection{Lower-Bounded Exponential Family}
\label{section:theory:exponential}
We consider the regret performance of RBMLE for Exponential Families with a known lower bound $\underline{\theta}$ on the mean. 
{E.g., $\underline{\theta}=0$ for Bernoulli distributions.}
Such a collection includes commonly-studied Exponential Families that are defined on the positive half real line, such as Exponential, Binomial, Poisson, and Gamma (with a fixed shape parameter).


\begin{prop}
\label{prop:regret}
For any Exponential Family with a lower bound $\underline{\theta}$ on the mean, for any $\varepsilon\in(0,1)$, the regret of RBMLE using (\ref{equation:RBMLE index for exp 1})-(\ref{equation:RBMLE index for exp 2}) with $\alpha(t)=C_{\alpha}\log t$ and $C_\alpha\geq {4}/({D(\theta_1-\frac{\varepsilon\Delta}{2},\theta_1) K^{*}(\theta_1-\frac{\varepsilon \Delta}{2},\underline{\theta})})$ satisfies
\begin{align}
    &\mathcal{R}(T)\leq \sum_{a=2}^{N}\Delta_a\big[\max\big\{\frac{4}{D(\theta_a+\frac{\varepsilon\Delta_a}{2},\theta_a)},\\
    &C_{\alpha}K^{*}(\theta_1-\frac{\varepsilon \Delta_a}{2},\theta_a+\frac{\varepsilon\Delta_a}{2})\big\}\log T + 1 + \frac{\pi^2}{3}\big].
\end{align}
\end{prop}

\subsubsection{Gaussian Distributions}
\label{section:theory:guassian}
\begin{prop}
\label{prop:regret Gaussian}
For Gaussian reward distributions with variance bounded by $\sigma^2$ for all arms, the regret of RBMLE using (\ref{equation:BMLE Gaussian in corollary}) with $\alpha(t)=C_{\alpha}\log t$ and $C_\alpha\geq \frac{256\sigma^2}{\Delta}$ satisfies
\begin{equation}
    \mathcal{R}(T)\leq\sum_{a=2}^{N}\Delta_a\big[\frac{2}{\Delta_a}C_{\alpha}\log T + \frac{2\pi^2}{3}\big].
\end{equation}
\end{prop}


\subsubsection{Beyond Parametric Distributions}
\label{section:theory:beyond parametric}
RBMLE indices derived for Exponential Families can be readily applied to other non-parametric distributions.
Moreover, the regret proofs in Propositions \ref{prop:regret}-\ref{prop:regret Gaussian} can be readily extended if the non-parametric rewards also satisfy proper concentration inequalities. 
We consider two classes of reward distributions, namely sub-Gaussian and sub-Exponential~\cite{wainwright2019high}:
\begin{definition}
A random variable $X$ with mean $\mu=\E[X]$ is $\sigma$-sub-Gaussian if there exists $\sigma>0$ such that
\begin{equation}
    \E[e^{\lambda(X-\mu)}]\leq e^{\frac{\sigma^2\lambda^2}{2}}, \hspace{3pt}\forall\lambda\in\mathbb{R}.
\end{equation}
\end{definition}
\begin{definition}
A random variable $X$ with mean $\mu=\E[X]$ is $(\rho,\kappa)$-sub-Exponential if there exist $\rho,\kappa\geq 0$ such that
\begin{equation}
    \E[e^{\lambda(X-\mu)}]\leq e^{\frac{\rho^2\lambda^2}{2}}, \hspace{3pt}\forall\lvert\lambda\rvert<\frac{1}{\kappa}.
\end{equation}
\end{definition}
\begin{prop}
\label{prop:BMLE regret for sub-Gaussian}
\normalfont For any $\sigma$-sub-Gaussian reward distributions, RBMLE using (\ref{equation:BMLE Gaussian in corollary}) with $\alpha(t)=C_{\alpha}\log t$ and $C_\alpha\geq \frac{256\sigma^2}{\Delta}$ yields $\mathcal{R}(T)\leq\sum_{a=2}^{N}\Delta_a\big[\frac{2}{\Delta_a}C_{\alpha}\log T + \frac{2\pi^2}{3}\big]$.
\end{prop}
\begin{myproof}
\normalfont The proof of Proposition \ref{prop:regret Gaussian} still holds for Proposition \ref{prop:BMLE regret for sub-Gaussian} without any change as Hoeffding's inequality directly works for sub-Gaussian distributions.
\end{myproof}

\begin{prop}
\label{prop:BMLE regret for sub-exponential}
\normalfont \pch{For any $(\rho,\kappa)$-sub-Exponential reward distribution defined on the positive half line with a lower bound $\underline{\theta}$ on the mean, RBMLE using (\ref{equation:RBMLE index for exp 1})}-(\ref{equation:RBMLE index for exp 2}) with $\alpha(t)=C_{\alpha}\log t$ and $C_{\alpha}\geq {16(\kappa\varepsilon\Delta+2\rho^2)}/((\varepsilon\Delta)^2 K^{*}(\theta_1-\frac{\varepsilon \Delta}{2},\underline{\theta}))$ achieves a regret bound
\begin{align}
    \mathcal{R}(T)\leq &\sum_{a=2}^{N}\Delta_a\big[1 + \frac{\pi^2}{3}+\max\big\{\frac{16(\kappa\varepsilon\Delta+2\rho^2)}{(\varepsilon\Delta_a)^2},\\
    &C_{\alpha}K^{*}(\theta_1-\frac{\varepsilon \Delta_a}{2},\theta_a+\frac{\varepsilon\Delta_a}{2})\big\}\log T\big].
\end{align}
\end{prop}


\begin{remark}
\normalfont
We highlight that the Propositions \ref{prop:regret}-\ref{prop:BMLE regret for sub-exponential} aim to provide the relationship between the upper bound on regret and the gap $\Delta$. We find that to establish the $O(\log(T))$ regret bound, the pre-constant $C_\alpha$ has to be large enough. One of our major technical contributions is to quantify the non-trivial relationship between $C_\alpha$ and $\Delta$ as well as the dependency between the bound and $\Delta$. A similar dependency also exists in other algorithms such as UCB, KL-UCB, and IDS, etc, due to the sharp characterization of the pre-constant by \cite{lai1985asymptotically}.
\pch{Moreover, we consider adaptive estimation of the gap as illustrated in Algorithms \ref{alg:pseudo code Gaussian}-\ref{alg:pseudo code Exponential}. 
In practice, we show that such adaptive scheme is sufficient to achieve excellent performance (see Section~\ref{section:simulation}).}
\end{remark}

\subsection{{\color{black}RBMLE with Adaptive Estimation of $C_{\alpha}$}}\label{second_question}
In this section, we provide the pseudo code of the experiments in Section \ref{section:simulation}. 
As discussed above, RBMLE achieves logarithmic regret by estimating $C_{\alpha}$ in $\alpha(t)=C_{\alpha}\log t$, where the estimation of $C_{\alpha}$ involves the minimum gap $\Delta$ and the largest mean $\theta_1$.
We consider the following adaptive scheme that gradually learns $\Delta$ and $\theta_1$.
\begin{algorithm}[!htbp]
   \caption{Adaptive Scheme with Estimation of $C_{\alpha}$ in Gaussian Bandits}
   \label{alg:pseudo code Gaussian}
\begin{algorithmic}[1]
   \STATE {\bfseries Input:} $N$, $\sigma$, and $\beta(t)$
    \FOR{$t=1,2,\cdots$}
      \FOR{$i=1$ {\bfseries to} $N$}
      \STATE $U_i(t)=p_i(t)+\sqrt{2\sigma^2(N+2)\log t/N_i(t)}$ $\slash\slash$ upper confidence bound of the empirical mean
      \STATE $L_i(t)=p_i(t)-\sqrt{2\sigma^2(N+2)\log t/N_i(t)}$ $\slash\slash$ lower confidence bound of the empirical mean
      \ENDFOR
       \STATE $\hat{\Delta}_t=\max_{i}\Big\{\max\big(0,L_i(t)-\max_{j\neq i}U_j(t)\big)\Big\}$
        \STATE Calculate $\hat{C}_{\alpha}(t)=\frac{256\sigma^2}{\hat{\Delta}_t}$
        \STATE $\alpha(t)=\min\{\hat{C}_{\alpha}(t),\beta(t)\}\log t$
    \ENDFOR
\end{algorithmic}
\end{algorithm}

\begin{itemize}[leftmargin=*]
    \item \textbf{Estimate $\Delta$ and $\theta_1$}: Note that $\Delta$ can be expressed as $\max_{1\leq i\leq N}\{\theta_i-\max_{j\neq i}\theta_j\}$. For each arm $i$, construct $U_i(t)$ and $L_i(t)$ as the upper and lower confidence bounds of $p_i(t)$ based on proper concentration inequalities. Then, construct an estimator of $\Delta$ as $\hat{\Delta}_t:=\max_{1\leq i\leq N}\big\{\max\big(0, L_i(t)-\max_{j\neq i}U_j(t)\big)\big\}$.
    Meanwhile, we use $U_{\max}(t):=\max_{1\leq i\leq N}U_i(t)$ as an estimate of $\theta_1$.
    Based on the confidence bounds, we know $\hat{\Delta}_t\leq \Delta$ and $U_{\max}(t)\geq \theta_1$, with high probability.
    \item \textbf{Construct the bias using estimators}: We construct $\alpha(t)=\min\{\widehat{C}_{\alpha}(t),\beta(t)\}\log t $, where $\widehat{C}_{\alpha}(t)$ estimates $C_{\alpha}(t)$ by replacing $\Delta$ with $\hat{\Delta}_t$ and $\theta_1$ with $U_{\max}(t)$, with $\beta(t)$ a non-negative strictly increasing function satisfying $\lim_{t\rightarrow\infty}\beta(t)=\infty$ (e.g. $\beta(t)=\sqrt{\log t}$ in the experiments in Section \ref{section:simulation}). 
    With high probability, $\widehat{C}_{\alpha}(t)$ gradually approaches the target value $C_{\alpha}(t)$ from above as time evolves. 
    On the other hand, $\beta(t)$ guarantees smooth exploration initially and will ultimately exceed $\widehat{C}_{\alpha}(t)$. 
\end{itemize}

\subsubsection{(sub) Gaussian Distributions}\label{section:adaptive}
To further illustrate the overall procedure, we first use the $C_\alpha$ in Propositions \ref{prop:regret Gaussian} and \ref{prop:BMLE regret for sub-Gaussian} as an example, with the pseudo code provided in Algorithm~\ref{alg:pseudo code Gaussian}.
The main idea is to learn the appropriate $C_{\alpha}$ considered in the regret analysis by gradually increasing $C_{\alpha}$ until it is sufficiently large.
This is accomplished by setting $\alpha(t)=\min\{\hat{C}_{\alpha}(t),\beta(t)\}\log t$ (Line 9 in Algorithm \ref{alg:pseudo code Gaussian})
where $\hat{C}_{\alpha}(t)$ serves as an over-estimate of the minimum required $C_{\alpha}$ based on the estimators $\hat{\Delta}_t$ for $\Delta$ (Lines 3-8 in Algorithm \ref{alg:pseudo code Gaussian}). 
$\hat{\Delta}_t$ is a conservative estimate of $\Delta$ in the sense that $\hat{\Delta}_t\leq \Delta$, conditioned on the high probability events $\theta_i\in[L_i(t),U_i(t)]$, for all $i$.
Here the confidence bounds $L_i(t)$ and $U_i(t)$ are constructed with the help of Hoeffding's inequality.
For small $t$, it is expected that $\hat{\Delta}_t$ is very close to zero and hence $\hat{C}_{\alpha}(t)$ is large.
Therefore, initially $\beta(t)$ serves to gradually increase $C_{\alpha}$ and guarantees enough exploration after $\beta(t)$ exceeds the minimum required $C_{\alpha}$.
Given sufficient exploration enabled by $\beta(t)$, the estimate $\hat{\Delta}_t$ gets accurate (i.e. $\hat{\Delta}_t\approx \Delta$), and subsequently $\hat{C}_{\alpha}(t)$ is clamped at some value slightly larger than the minimum required $C_{\alpha}$.

Next, we quantify the regret performance of RBMLE in Algorithm \ref{alg:pseudo code Gaussian} in Proposition \ref{prop:regret adaptive Gaussian} as follows.

\begin{prop}
\label{prop:regret adaptive Gaussian}
For any $\sigma$-sub-Gaussian reward distributions, the regret of RBMLE given by Algorithm \ref{alg:pseudo code Gaussian} satisfies
\begin{equation}
    \mathcal{R}(T)\leq\sum_{a=2}^{N}\Delta_a\Big[\max\Big\{\frac{1024\sigma^2(N+2)}{\Delta^2}\log T, T_0 \Big\}+ \frac{N\pi^2}{3}\Big],
\end{equation}
where $T_0:=\min\{t\in\mathbb{N}:\beta(t)\geq \frac{256\sigma^2 (N+2)}{\Delta}\}<\infty$.
\end{prop}


\subsubsection{Bernoulli Distributions}\label{section:adaptive_bernoulli}

As above, Algorithm \ref{alg:pseudo code Bernoulli} shows the pseudo code for estimating the $C_\alpha$ of $\alpha(t)$ in Bernoulli bandits.
Similar to the Gaussian case, $C_\alpha$ is estimated based on $\hat{\Delta}_t$ and $U_{\max}(t)$ with the help of Hoeffding's inequality.
In addition, as the calculation of $\hat{C}_{\alpha}(t)$ involves the subroutine of searching for the value $K^*(U_{\max}(t)-\frac{\varepsilon\hat{\Delta}_t}{2},0)$, we can accelerate the adaptive scheme by first checking if it is possible to have $\hat{C}_{\alpha}(t)\geq \beta(t)$.
Equivalently, this can be done by quickly verifying whether $\xi(\frac{N+2}{2(\varepsilon\hat{\Delta}_t)^2\beta(t)},0) <\dot{F}\inv(U_{\max}(t)-\frac{\varepsilon\hat{\Delta}_t}{2})$ (Line 9 in Algorithm \ref{alg:pseudo code Bernoulli}).


\begin{algorithm}[!tb]
   \caption{Adaptive Scheme with Estimation of $C_\alpha$ in Bernoulli Bandits}
   \label{alg:pseudo code Bernoulli}
\begin{algorithmic}[1]
   \STATE {\bfseries Input:} $N$, $\varepsilon\in(0,\frac{1}{2})$, and $\beta(t)$
    \FOR{$t=1,2,\cdots$}
      \FOR{$i=1$ {\bfseries to} $N$}
      \STATE $U_i(t)=\min\Big(p_i(t)+\sqrt{(N+2)\log t/N_i(t)},1\Big)$ $\slash\slash$ upper confidence bound of the empirical mean \label{lst:Bernoulli Ui}
      \STATE $L_i(t)=\max\Big(p_i(t)-\sqrt{(N+2)\log t/N_i(t)},0\Big)$ $\slash\slash$ lower confidence bound of the empirical mean \label{lst:Bernoulli Li}
      \ENDFOR 
       \STATE $U_{\max}(t)=\max_{i=1,\cdots,N}U_i(t)$ \label{lst:Bernoulli Umax}
       \STATE $\hat{\Delta}_t=\max_{i}\Big\{\max\big(0,L_i(t)-\max_{j\neq i}U_j(t)\big)\Big\}$ \label{lst:Bernoulli Delta hat}
       \IF{$\xi\big(\frac{N+2}{2(\varepsilon\hat{\Delta}_t)^2\beta(t)},0\big)< \dot{F}\inv(U_{\max}(t)-\frac{\varepsilon\hat{\Delta}_t}{2})$}\label{lst:Bernoulli if}
           \STATE $\alpha(t)=\beta(t)\log t$ $\slash\slash$ In this case, we know $\hat{C}_{\alpha}(t)>\beta(t)$ \label{lst:Bernoulli if action}
        \ELSE 
           \STATE Find $\hat{C}_{\alpha}(t)=\frac{N+2}{2(\varepsilon\hat{\Delta}_t)^2 K^*(U_{\max}(t)-\frac{\varepsilon\hat{\Delta}_t}{2},0)}$ by solving the minimization problem of (\ref{equation:K star definition}) for $K^*(U_{\max}(t)-\frac{\varepsilon\hat{\Delta}_t}{2},0)$. \label{lst:Bernoulli else action 1}
           \STATE $\alpha(t)=\min\{\hat{C}_{\alpha}(t),\beta(t)\}\log t$ \label{lst:Bernoulli else action 2}
        \ENDIF
    \ENDFOR
\end{algorithmic}
\end{algorithm}

\subsubsection{Exponential Distributions}\label{section:adaptive_exponential}
Algorithm \ref{alg:pseudo code Exponential} demonstrates the pseudo code for selecting $\alpha(t)$ in exponential bandits.
Compared to the Bernoulli case, 
the main difference in the exponential case lies in the construction of the confidence bounds (Lines 4-5 in Algorithm \ref{alg:pseudo code Exponential}), which leverage the sub-exponential tail bounds instead of Hoeffding's inequality.

\begin{algorithm}[!tb]
   \caption{Adaptive Scheme with Estimation $\alpha(t)$ in Exponential Bandits}
  \label{alg:pseudo code Exponential}
\begin{algorithmic}[1]
   \STATE {\bfseries Input:} $N$, $\varepsilon\in(0,\frac{1}{2})$, and $\beta(t)$
    \FOR{$t=1,2,\cdots$}
      \FOR{$i=1$ {\bfseries to} $N$}
      \STATE $U_i(t)=p_i(t)+\frac{\kappa(N+2)\log t+\sqrt{\kappa^2(N+2)^2(\log t)^2}+2\rho^2(N+2)\log t}{N_i(t)}$ $\slash\slash$ upper confidence bound
      \STATE $L_i(t)=\max\Big(p_i(t)-\frac{\kappa(N+2)\log t+\sqrt{\kappa^2(N+2)^2(\log t)^2}+2\rho^2(N+2)\log t}{N_i(t)},0\Big)$ $\slash\slash$ lower confidence bound 
      \ENDFOR
       \STATE $U_{\max}(t)=\max_{i=1,\cdots,N}U_i(t)$
       \STATE $\hat{\Delta}_t=\max_{i}\Big\{\max\big(0,L_i(t)-\max_{j\neq i}U_j(t)\big)\Big\}$
       \IF{$\xi\big(\frac{16(\kappa\varepsilon\hat{\Delta}_t+2\rho^2)}{(\varepsilon\hat{\Delta}_t)^2\beta(t)},0\big)< \dot{F}\inv(U_{\max}(t)-\frac{\varepsilon\hat{\Delta}_t}{2})$}
            \STATE $\alpha(t)=\beta(t)\log t$ $\slash\slash$ In this case, we know $\hat{C}_{\alpha}(t)>\beta(t)$
        \ELSE
           \STATE Find $\hat{C}_{\alpha}(t)=\frac{16(\kappa\varepsilon \hat{\Delta}_t+2\rho^2)}{(\varepsilon\hat{\Delta}_t)^2 K^*(U_{\max}(t)-\frac{\varepsilon\hat{\Delta}_t}{2},0)}$ by solving the minimization problem of (\ref{equation:K star definition}) for $K^*(U_{\max}(t)-\frac{\varepsilon\hat{\Delta}_t}{2},0)$. 
           $\alpha(t)=\min\{\hat{C}_{\alpha}(t),\beta(t)\}\log t$
        \ENDIF
    \ENDFOR
\end{algorithmic}
\end{algorithm}

\begin{figure*}[!tb]
\centering
$\begin{array}{c c c}
    \multicolumn{1}{l}{\mbox{\bf }} & \multicolumn{1}{l}{\mbox{\bf }} & \multicolumn{1}{l}{\mbox{\bf }} \\ 
    \hspace{-1mm} \scalebox{0.33}{\includegraphics[width=\textwidth]{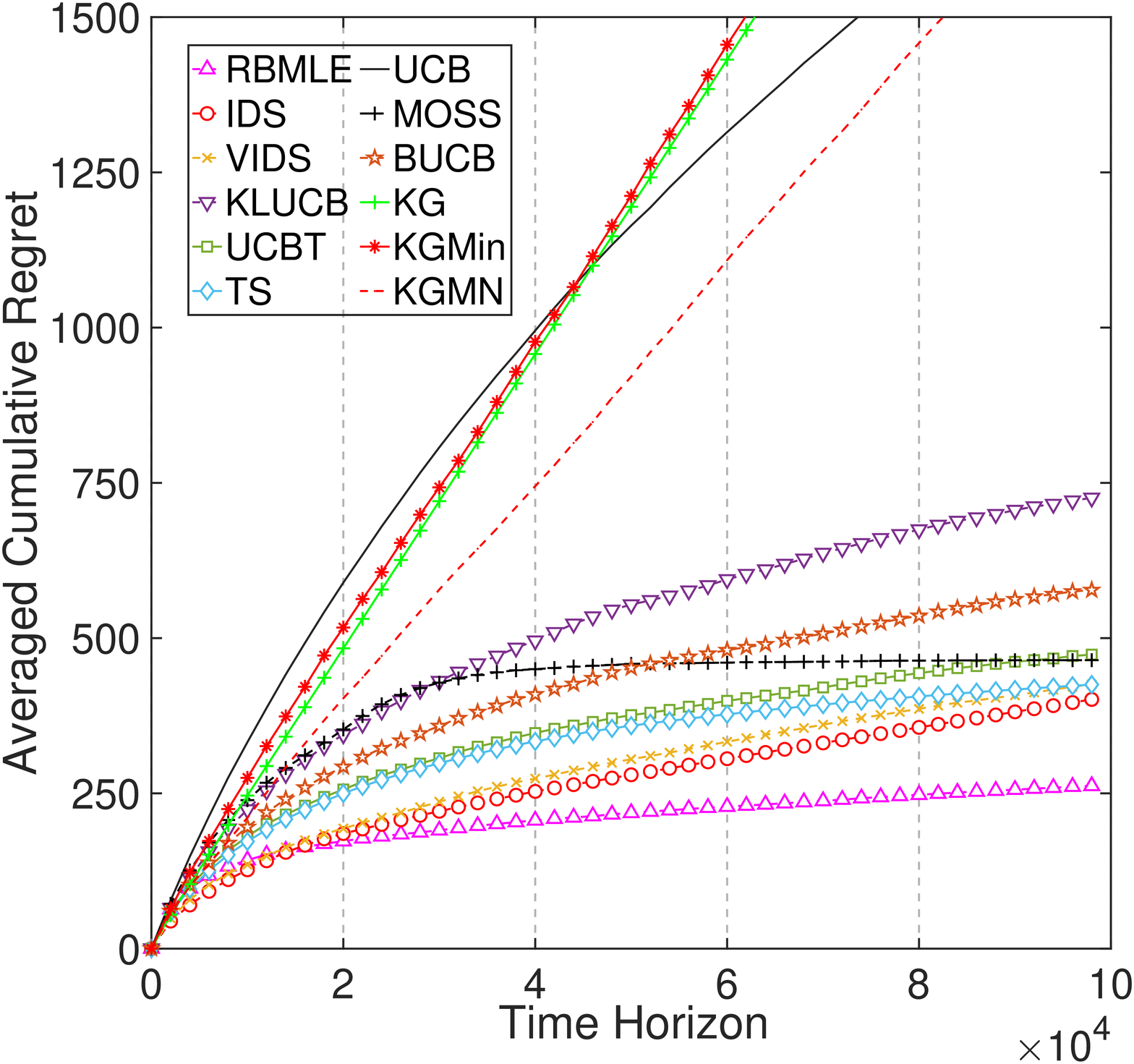}}  \label{fig:Bernoulli_ID=2_Regret_T=1e5} & 
    \hspace{-3mm} \scalebox{0.33}{\includegraphics[width=\textwidth]{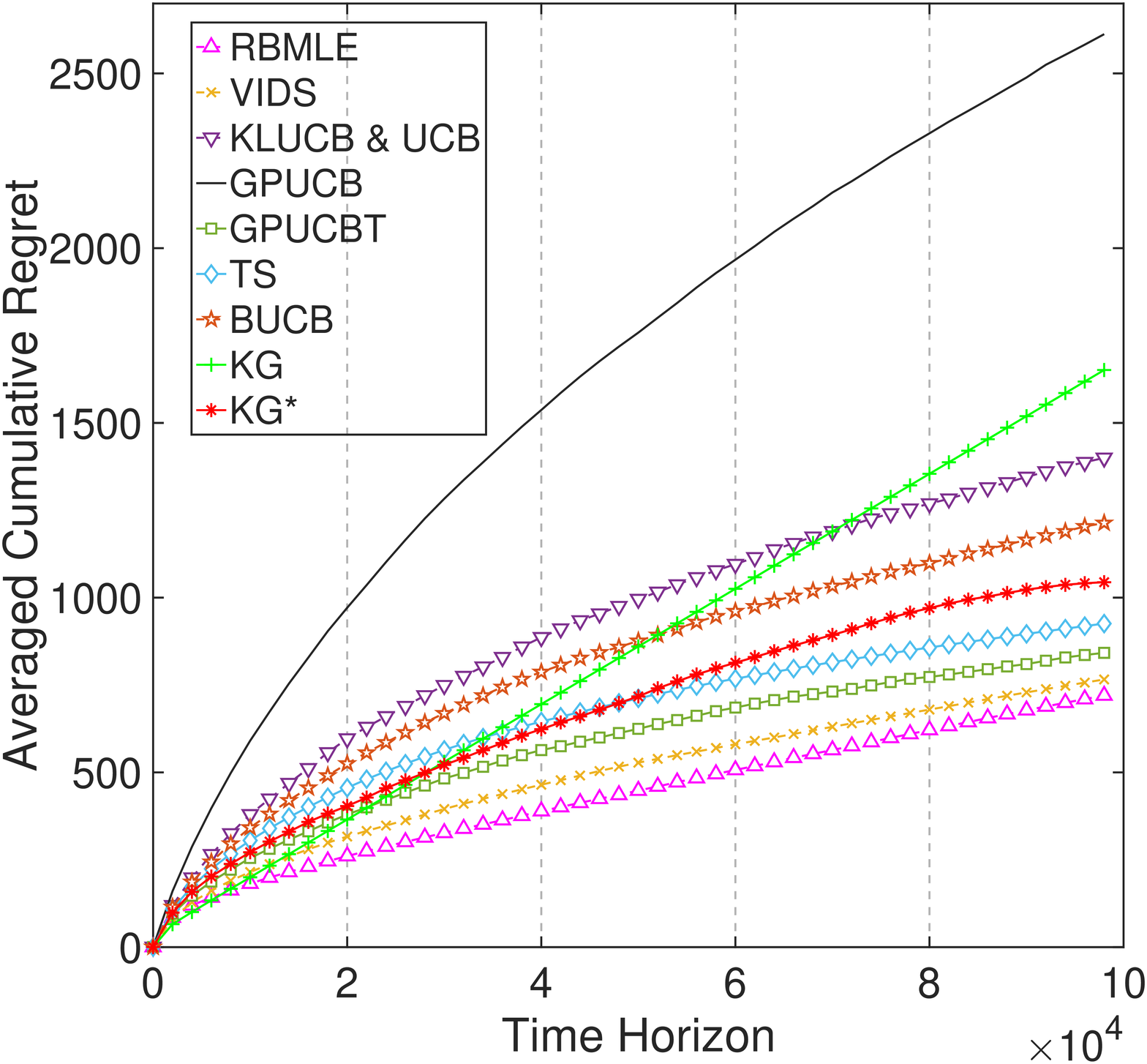}} \label{fig:Gaussian_ID=16_Regret_T=1e5} & \hspace{-3mm}
    \scalebox{0.33}{\includegraphics[width=\textwidth]{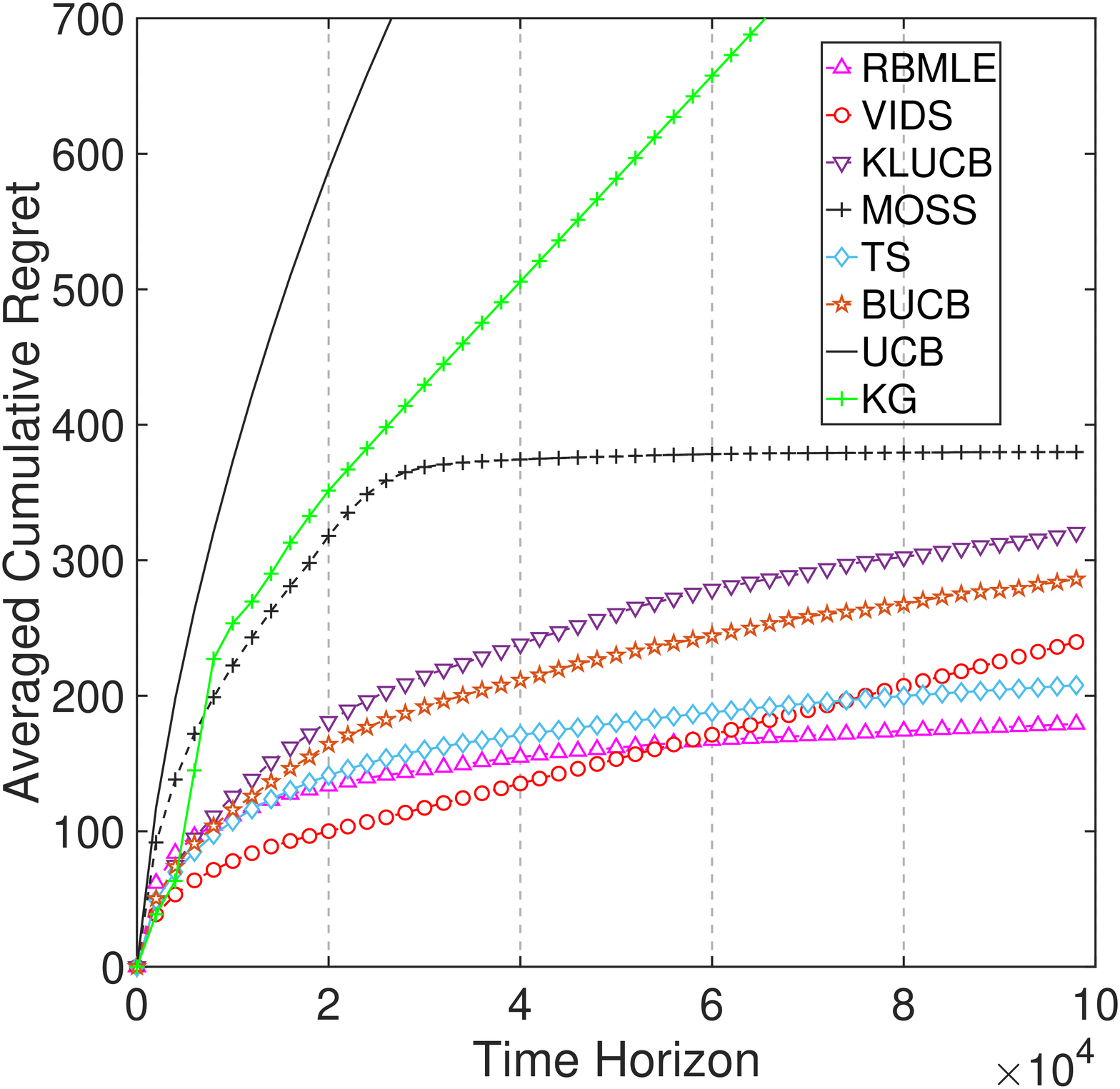}} \label{fig:Exp_ID=21_Regret_T=1e5}\\ [0.0cm]
    \mbox{\small (a)} & \hspace{-2mm} \mbox{\small (b)} & \hspace{-2mm} \mbox{\small (c)} \\[-0.2cm]
\end{array}$
\caption{Averaged cumulative regret: (a) Bernoulli bandits with $(\theta_i)_{i=1}^{10}=$ (0.66, 0.67, 0.68, 0.69, 0.7, 0.61, 0.62, 0.63, 0.64, 0.65) \& $\Delta=0.01$; (b) Gaussian bandits with $(\theta_i)_{i=1}^{10}=$ (0.41, 0.52, 0.66, 0.43, 0.58, 0.65, 0.48, 0.67, 0.59, 0.63) \& $\Delta=0.01$; (c) Exponential bandits with $(\theta_i)_{i=1}^{10}=$ (0.31, 0.1, 0.2, 0.32, 0.33, 0.29, 0.2, 0.3, 0.15, 0.08) \& $\Delta=0.01$.}
\label{fig:regret all}
\end{figure*}

\section{Simulation Experiments}
\label{section:simulation}
To evaluate the performance of the proposed RBMLE algorithms, we conduct a comprehensive empirical comparison with other state-of-the-art methods vis-a-vis three aspects: effectiveness (cumulative regret), efficiency (computation time per decision vs. cumulative regret), and scalability (in number of arms). We paid particular attention to the fairness of comparisons and reproducibility of results. To ensure sample-path sameness for all methods, we considered each method over a pre-prepared dataset containing the context of each arm and the outcomes of pulling each arm over all rounds. Hence, the outcome of pulling an arm is obtained by querying the pre-prepared data instead of calling the random generator and changing its state. A few benchmarks such as Thompson Sampling (TS) and Variance-based Information Directed Sampling (VIDS) that rely on outcomes of random sampling in each round of decision-making are separately evaluated with the same prepared data and with the same seed. To ensure reproducibility of experimental results, we set up the seeds for the random number generators at the beginning of each experiment. 

The benchmark methods compared include UCB \cite{auer2002finite}, UCB-Tuned (UCBT) \cite{auer2002finite}, KLUCB \cite{cappe2013kullback}, MOSS~\cite{audibert2009minimax}, Bayes-UCB (BUCB) \cite{kaufmann2012bayesian}, GPUCB \cite{srinivas2012information}, GPUCB-Tuned (GPUCBT) \cite{srinivas2012information}, TS \cite{agrawal2012analysis}, Knowledge Gradient (KG) \cite{ryzhov2012knowledge},  KG*~\cite{ryzhov2010robustness}, KGMin \cite{kaminski2015refined} KGMN \cite{kaminski2015refined}, IDS \cite{russo2018learning}, and VIDS, \cite{russo2018learning}. A detailed review of these methods is presented in Section \ref{section:related}. The values of their hyper-parameters are as follows. In searching for a solution of KLUCB and $\hat{C}_{\alpha}(t)$ in RBMLE, the maximum number of iterations is set to be 100. Following the suggestion in the original papers, we take $c=0$ in KLUCB and BUCB. We take $\delta=10^{-5}$ in GPUCB. We tune the parameter $c$ in GPUCBT for each experiment and choose $c=0.9$ that achieves the best performance. In the comparison with IDS and VIDS, we uniformly sampled $100$ points over the interval $[0,1]$ for Bernoulli and Exponential Bandits and sampled $1000$ points for Gaussian bandits (the $q$ in Algorithm 4 in~\cite{russo2018learning}) and take $M=10^4$ in sampling (Algorithm 3 in~\cite{russo2018learning}). The conjugate priors for Bayesian-family methods are Beta distribution $\mathcal{B}(1,1)$ for Bernoulli bandits, $\mathcal{N}(0, 1)$ for Gaussian bandits with $\sigma=1$, and Gamma distribution $\Gamma(1,1)$ for Exponential bandits with $\rho=10$ and $\kappa=10$. The average is taken over 100 trials. The time horizon in the experiments for effectiveness and efficiency is $10^5$, and for scalability is $10^4$.

\textbf{Effectiveness.} 
Figures \ref{fig:regret all}, \ref{fig:regret all 2}--\ref{fig:regret all 3} and Tables \ref{table:Bernoulli_ID=2}-\ref{table:Exponential_ID=17} (some are in Appendix \ref{appendix:more empirical results on effectiveness}) illustrate the effectiveness of RBMLE with respect to the cumulative regret as well as quantiles. Note that in the (b) sub-figures of these figures, since KLUCB shares the same closed-form index as UCB in Gaussian bandits, their curves coincide. We observe that for 
all three types of bandits, Bernoulli, Gaussian and Exponential, RBMLE achieves competitive performance, often slightly better than the best performing benchmark method. IDS or VIDS are often the closest competitors to RBMLE. However, the computational complexity of RBMLE is much lower compared to IDS and VIDS, which need to compute high dimensional integrals or estimate them through sampling. One other advantage of RBMLE over some benchmark methods is that it is ``time horizon agnostic'', i.e., the computation of RBMLE index does not need the knowledge of time horizon $T$. In contrast, BUCB, MOSS, GPUCBT, and KG-family algorithms (KG, KG*) need to know $T$. It is worth mentioning that in Bernoulli bandits, KG, KGMin, and KGMN perform poorly as they explore insufficiently. This is not surprising as several papers have pointed out the limitations of KG-family methods when observations are discrete~\cite{kaminski2015refined}.

\begin{figure*}[!tbp]
\centering
$\begin{array}{c c c}
    \multicolumn{1}{l}{\mbox{\bf }} & \multicolumn{1}{l}{\mbox{\bf }} & \multicolumn{1}{l}{\mbox{\bf }} \\ 
    \hspace{-1mm} \scalebox{0.33}{\includegraphics[width=\textwidth]{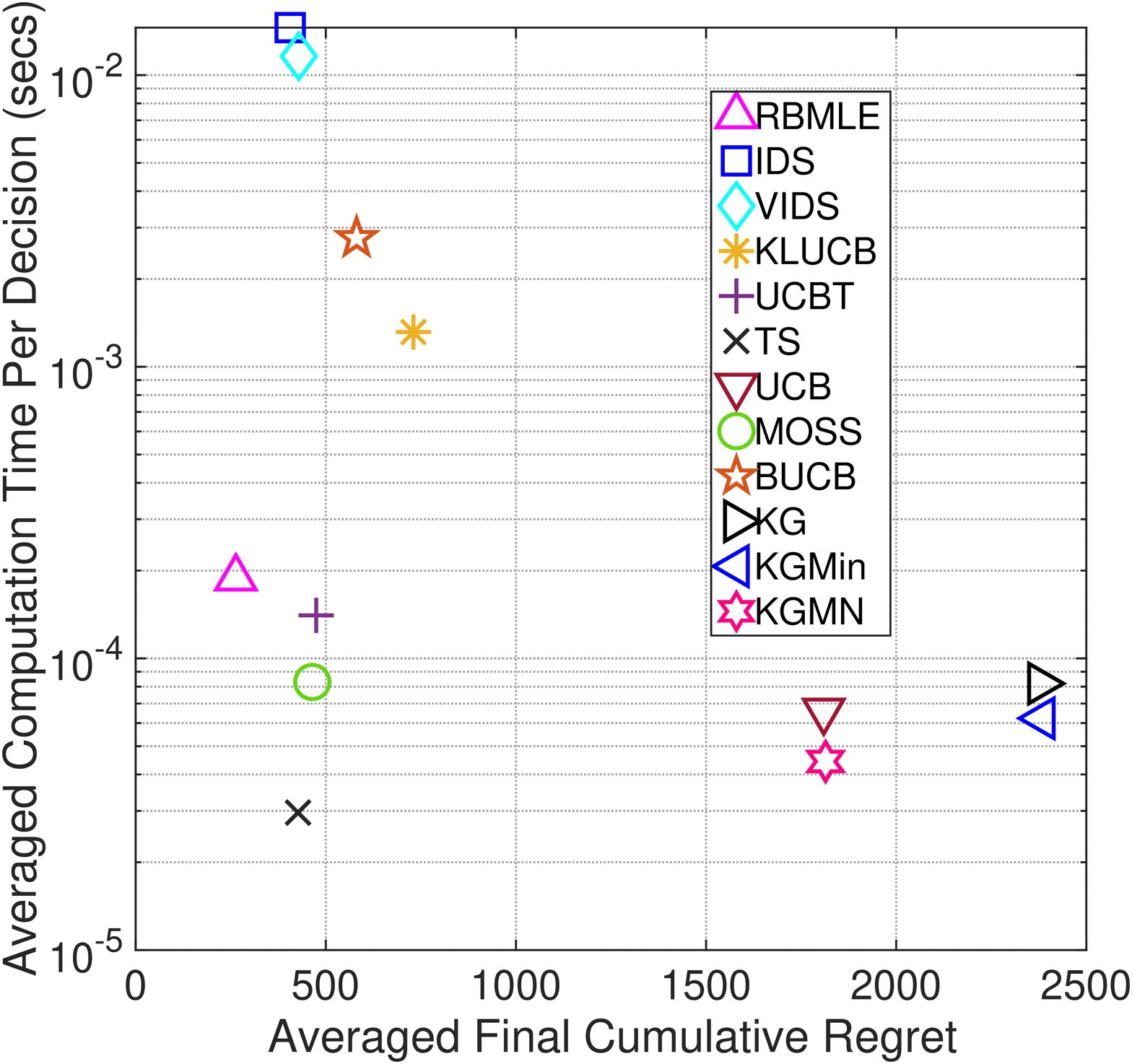}}  \label{fig:Bernoulli_ID=2_Scatter_T=1e5} & 
    \hspace{-3mm} \scalebox{0.33}{\includegraphics[width=\textwidth]{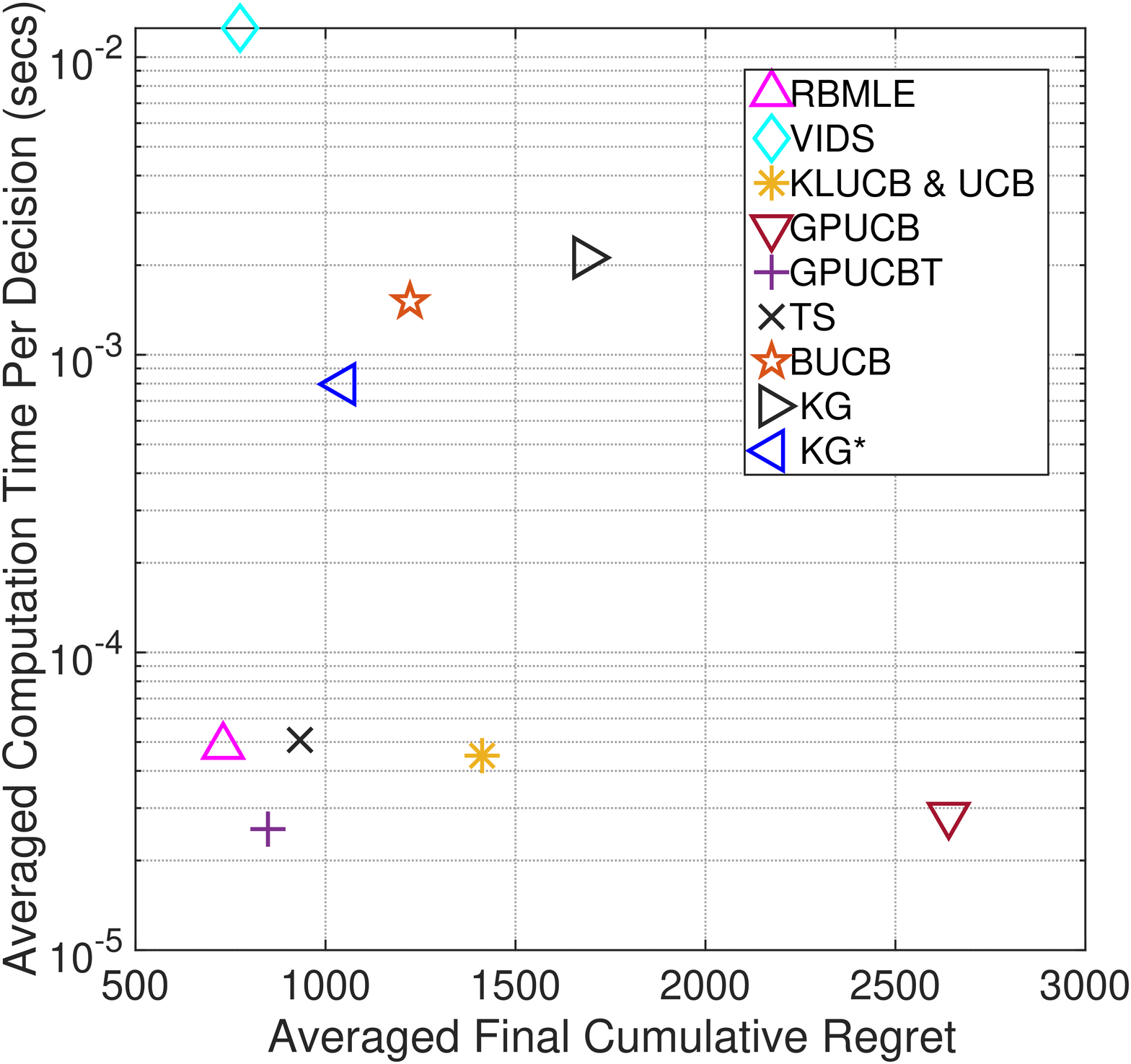}} \label{fig:Gaussian_ID=16_Scatter_T=1e5} & \hspace{-3mm}
    \scalebox{0.33}{\includegraphics[width=\textwidth]{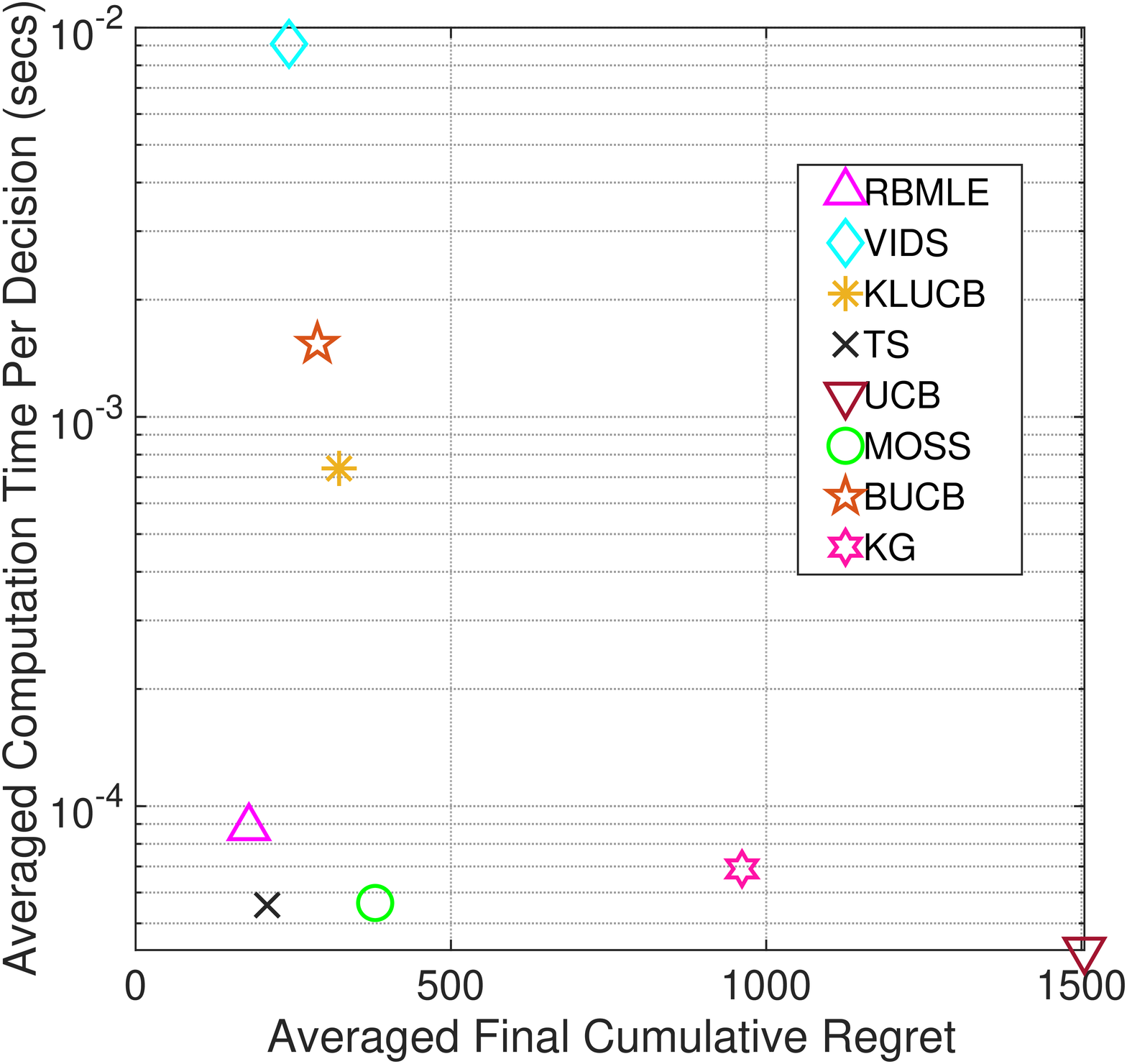}} \label{fig:Exp_ID=21_Scatter_T=1e5}\\ [0.0cm]
    \mbox{\small (a)} & \hspace{-2mm} \mbox{\small (b)} & \hspace{-2mm} \mbox{\small (c)} \\[-0.2cm]
\end{array}$
\caption{Averaged computation time per decision vs. averaged final cumulative regret: (a) Figure \ref{fig:regret all}(a); (b) Figure \ref{fig:regret all}(b);  (c) Figure \ref{fig:regret all}(c).}
\label{fig:time all}
\end{figure*}

\textbf{Efficiency}
Figures \ref{fig:time all} and \ref{fig:time all 2}--\ref{fig:time all 3} (some are in Appendix \ref{appendix:more empirical results on efficiency}) present the efficiency of RBMLE in terms of averaged computation time per decision vs. averaged final cumulative regret. The former averaged the total time spent in all trials over the number of trials and number of rounds. The latter averaged the total cumulative regret in all trials over the number of trials. RBMLE is seen to provide competitive performance compared to other benchmark methods. 
It achieves slightly better regret, {\color{black}and does so with orders of magnitude less computation time}, than IDS, VIDS and KLUCB. This is largely because IDS and VIDS need to estimate several integrals in each round, and KLUCB often relies on Newton's method or bisection search to find the index for an arm, except in Gaussian bandtis, where KLUCB has a simple closed-form solution. It is also observed that the computation time of RBMLE is larger than some benchmark methods such as UCB, GPUCB, and KG, which enjoy a simple closed-form index. However, their regret performance is far worse than RBMLE's. In the comparison of efficiency, the closest competitors to RBMLE are TS, MOSS, UCBT, and GPUCBT. Compared to them, RBMLE still enjoys salient advantages in different aspects. Compared to TS, RBMLE follows the frequentist formulation and thus its performance does not deteriorate when a bad prior is chosen. Compared to MOSS, RBMLE does not rely on the knowledge of $T$ to compute its index. Compared to UCBT as well as GPUCBT, RBMLE has a order-optimal regret bound, as proved in the earlier sections.

\textbf{Scalability}
Tables~\ref{table:Bernoulli_computation_time}-\ref{table:Exponential_computation_time} show the scalability of RBMLE as the number of arms is increased. This is illustrated through comparing different methods' averaged computation time per decision averaged computation time per decision under varying numbers of arms. We observe that RBMLE scales well for various reward distributions as the number of arms increases, often demonstrating performance comparable to  the most scalable ones among the benchmark methods. The averaged computation time per decision stays at a few $10^{-4}$ seconds even when the number of arms reaches 70. In contrast, it can be as high as thousands of $10^{-4}$ seconds for IDS and VIDS, and it is often tens of times higher for KLUCB.

\section{Related Work} \label{section:related}
The RBMLE approach proposed in \cite{kumar1982new} has been examined in a variety of settings, including MDPs and Linear-Quadratic-Gaussian systems, in \cite{kumar1982optimal,kumar1983simultaneous,kumar1983optimal,borkar1990kumar,borkar1991self,stettner1993nearly,duncan1994almost,campi1998adaptive,prandini2000adaptive}. The simple index for the case of Bernoulli bandits was derived in \cite{BeckerKumar81b}.

However, prior studies have been limited to focusing on long-term average reward optimality, which corresponds to a loose $o(T)$ bound on regret. This paper aims to tailor RBMLE to more general SMAB problems, and to prove its finite-time regret performance. 

Learning algorithms for SMAB problems have been extensively studied. Most prior studies can be categorized into frequentist approaches or Bayesian approaches.
In the frequentist settings, the family of UCB algorithms, including UCB~\cite{auer2002finite}, UCBT~\cite{auer2002finite}, and MOSS~\cite{audibert2009minimax,degenne2016anytime}, are among the most popular ones given their simplicity in implementation and the established regret bounds.
An upper confidence bound can be directly derived from concentration inequalities or constructed with the help of other information measures, such as the Kullback--Leibler divergence used by KLUCB \cite{filippi2010optimism,garivier2011kl,cappe2013kullback}.
The concept of upper confidence bound has later been extended to various types of models, such as contextual linear bandits \cite{chu2011contextual,abbasi2011improved,rusmevichientong2010linearly}, Gaussian process bandit optimization (GPUCB and GPUCBT) \cite{srinivas2012information}, and model-based reinforcement learning \cite{jaksch2010near}.
The above list is by no means exhaustive but is mainly meant to illustrate the wide applicability of the
UCB approach in different settings.
While being a simple and generic index-type algorithm, UCB-based methods sometimes suffer from much higher regret than their counterparts \cite{russo2014learning,chapelle2011empirical}.
Different from the UCB solutions, the proposed RBMLE algorithm addresses the exploration and exploitation tradeoff by directly operating with the likelihood function to navigate the exploration, and therefore it makes better use of the information of the parametric distributions.

On the other hand, the Bayesian approach studies the setting where the unknown reward parameters are drawn from an underlying prior distribution.
As one of the most popular Bayesian bandit algorithms, TS \cite{scott2010modern,chapelle2011empirical,agrawal2012analysis,korda2013thompson,kaufmann2012thompson} follows the principle of probability matching by continuously updating the posterior distribution based on a prior.
In addition to strong theoretical guarantees \cite{agrawal2012analysis,kaufmann2012thompson}, TS has been reported to achieve superior empirical performance to its counterparts \cite{chapelle2011empirical,scott2010modern}.
While being a powerful bandit algorithm, TS can be sensitive to the choice of the prior \cite{korda2013thompson,liu2016prior}.
Another popular Bayesian algorithm is BUCB \cite{kaufmann2012bayesian}, which 
combines the Bayesian interpretation of bandit problems and the simple closed-form expression of UCB-type algorithms.
In contrast, RBMLE does not rely on a prior and hence completely obviates the potential issues arising from an inappropriate prior choice. 
Another line of investigation takes advantage of the Bayesian update in information-related measures. KG~\cite{ryzhov2012knowledge} and its variant KG*~\cite{ryzhov2010robustness}, KGMin, and KGMN~\cite{kaminski2015refined,ryzhov2012knowledge,ryzhov2010robustness} proceed by making a greedy one-step look-ahead measurement for exploration, as suggested by their names. 
While KG has been shown to empirically perform well for Gaussian distributions \cite{ryzhov2010robustness,wang2016knowledge}, its performance is not readily quantifiable, and it does not always converge to optimality \cite{russo2014learning}. 
Another competitive solution is IDS and its variant,VIDS, proposed by \citet{russo2018learning}.
Different from the KG algorithm, IDS blends in the concept of information gain by looking at the ratio between the square of expected immediate regret and the expected reduction in the entropy of the target.
Moreover, it has been reported in \cite{russo2014learning,russo2017learning} that IDS achieves state-of-the-art results in various bandit models.
However, IDS and VIDS suffer from high computational complexity and poor scalability due to the excessive sampling required for estimating high dimensional integrals.
Compared to these regret-competitive solutions, the proposed RBMLE algorithms can achieve comparable performance both theoretically and empirically, but at the same time it retains computational efficiency.

\vspace{-2mm}
\section{Concluding Remarks}
\label{section:discussion}
The RBMLE method, developed in the general study of adaptive control four decades ago, provides a scheme for optimal control of general Markovian systems. 
It exploits the observation that the MLE has a one-sided bias favoring parameters with smaller optimal rewards, and so delicately steers the scheme to an optimal estimate by biasing the MLE in the reverse way. Just like the later Upper Confidence Bound policy, it also can be regarded as exemplifying the philosophy of being ``optimistic in the face of uncertainty," but does it in a very different way. The resulting indices (Table \ref{tab:my-table}) are very different. 
Over the four decades since its introduction, the RBMLE method has not been further analyzed or empirically evaluated for its regret performance.

This paper takes an initial step in this direction. It shows how indices can be derived naturally for the Exponential Family of reward distributions, and how these indices can even be applied to other non-parametric distributions. 
It studies the interplay between the choice of the growth rate of the reward-bias and the resulting regret. It exposes the important role played by the knowledge of the minimum gap in the choice of the reward-bias growth rate. When this minimum gap is not known, it shows how it can be adaptively estimated. It empirically shows that
RBMLE attains excellent regret performance compared with other state-of-art methods, while requiring low computation time per decision compared to other methods with comparable regret performance, and scales well as the number of arms is increased.

Being a general purpose approach for optimal decision making under certainty, RBMLE holds potential for a number of such problems, including contextual bandits, adversarial bandits, Bayesian optimization and more beyond.  

\vspace{-2mm}
\section*{Acknowledgments}

\vspace{-2mm}
{We sincerely thank all anonymous reviewers and meta reviewers for their insightful suggestions during rebuttal! This work is partially supported by the Ministry of
Science and Technology of Taiwan under Contract No. MOST 108-2636-E-009-014. This material is also based upon work partially supported by the U.S.~Army Research Office under Contract No. W911NF-18-10331, U.S.~ONR under Contract No. N00014-18-1-2048, and NSF Science \& Technology Center Grant CCF-0939370. }

{
\small
\bibliographystyle{icml2020}
\bibliography{reference}

\begin{thebibliography}{43}
\providecommand{\natexlab}[1]{#1}
\providecommand{\url}[1]{\texttt{#1}}
\expandafter\ifx\csname urlstyle\endcsname\relax
  \providecommand{\doi}[1]{doi: #1}\else
  \providecommand{\doi}{doi: \begingroup \urlstyle{rm}\Url}\fi

\bibitem[Abbasi-Yadkori et~al.(2011)Abbasi-Yadkori, P{\'a}l, and
  Szepesv{\'a}ri]{abbasi2011improved}
Abbasi-Yadkori, Y., P{\'a}l, D., and Szepesv{\'a}ri, C.
\newblock Improved algorithms for linear stochastic bandits.
\newblock In \emph{Advances in Neural Information Processing Systems}, pp.\
  2312--2320, 2011.

\bibitem[Agrawal \& Goyal(2012)Agrawal and Goyal]{agrawal2012analysis}
Agrawal, S. and Goyal, N.
\newblock Analysis of {T}hompson sampling for the multi-armed bandit problem.
\newblock In \emph{Conference on Learning Theory (COLT)}, pp.\  39--1, 2012.

\bibitem[Audibert \& Bubeck(2009)Audibert and Bubeck]{audibert2009minimax}
Audibert, J.-Y. and Bubeck, S.
\newblock Minimax policies for adversarial and stochastic bandits.
\newblock In \emph{COLT}, pp.\  217--226, 2009.

\bibitem[Auer et~al.(2002)Auer, Cesa-Bianchi, and Fischer]{auer2002finite}
Auer, P., Cesa-Bianchi, N., and Fischer, P.
\newblock Finite-time analysis of the multiarmed bandit problem.
\newblock \emph{Machine learning}, 47\penalty0 (2-3):\penalty0 235--256, 2002.

\bibitem[Becker \& Kumar(1981)Becker and Kumar]{BeckerKumar81b}
Becker, A. and Kumar, P.~R.
\newblock {Optimal strategies for the N-armed bandit problem}.
\newblock Technical report, Mathematics Research Report No. 81-1, Department of
  Mathematics, University of Maryland Baltimore County, Jan 1981.

\bibitem[Borkar \& Varaiya(1979)Borkar and Varaiya]{borkar1979adaptive}
Borkar, V. and Varaiya, P.
\newblock Adaptive control of {M}arkov chains, {I}\: {F}inite parameter set.
\newblock \emph{IEEE Transactions on Automatic Control}, 24\penalty0
  (6):\penalty0 953--957, 1979.

\bibitem[Borkar(1990)]{borkar1990kumar}
Borkar, V.~S.
\newblock The {K}umar-{B}ecker-{L}in scheme revisited.
\newblock \emph{Journal of Optimization Theory and Applications}, 66\penalty0
  (2):\penalty0 289--309, 1990.

\bibitem[Borkar(1991)]{borkar1991self}
Borkar, V.~S.
\newblock Self-tuning control of diffusions without the identifiability
  condition.
\newblock \emph{Journal of optimization theory and applications}, 68\penalty0
  (1):\penalty0 117--138, 1991.

\bibitem[Campi \& Kumar(1998)Campi and Kumar]{campi1998adaptive}
Campi, M.~C. and Kumar, P.~R.
\newblock Adaptive linear quadratic {G}aussian control: the cost-biased
  approach revisited.
\newblock \emph{SIAM Journal on Control and Optimization}, 36\penalty0
  (6):\penalty0 1890--1907, 1998.

\bibitem[Capp{\'e} et~al.(2013)Capp{\'e}, Garivier, Maillard, Munos, Stoltz,
  et~al.]{cappe2013kullback}
Capp{\'e}, O., Garivier, A., Maillard, O.-A., Munos, R., Stoltz, G., et~al.
\newblock Kullback-leibler upper confidence bounds for optimal sequential
  allocation.
\newblock \emph{The Annals of Statistics}, 41\penalty0 (3):\penalty0
  1516--1541, 2013.

\bibitem[Chapelle \& Li(2011)Chapelle and Li]{chapelle2011empirical}
Chapelle, O. and Li, L.
\newblock An empirical evaluation of {T}hompson sampling.
\newblock In \emph{Advances in neural information processing systems}, pp.\
  2249--2257, 2011.

\bibitem[Chu et~al.(2011)Chu, Li, Reyzin, and Schapire]{chu2011contextual}
Chu, W., Li, L., Reyzin, L., and Schapire, R.
\newblock Contextual bandits with linear payoff functions.
\newblock In \emph{Proceedings of the Fourteenth International Conference on
  Artificial Intelligence and Statistics (AISTATS)}, pp.\  208--214, 2011.

\bibitem[Degenne \& Perchet(2016)Degenne and Perchet]{degenne2016anytime}
Degenne, R. and Perchet, V.
\newblock Anytime optimal algorithms in stochastic multi-armed bandits.
\newblock In \emph{International Conference on Machine Learning}, pp.\
  1587--1595, 2016.

\bibitem[Duncan et~al.(1994)Duncan, Pasik-Duncan, and
  Stettner]{duncan1994almost}
Duncan, T.~E., Pasik-Duncan, B., and Stettner, L.
\newblock Almost self-optimizing strategies for the adaptive control of
  diffusion processes.
\newblock \emph{Journal of optimization theory and applications}, 81\penalty0
  (3):\penalty0 479--507, 1994.

\bibitem[Filippi et~al.(2010)Filippi, Capp{\'e}, and
  Garivier]{filippi2010optimism}
Filippi, S., Capp{\'e}, O., and Garivier, A.
\newblock Optimism in reinforcement learning and {K}ullback-{L}eibler
  divergence.
\newblock In \emph{2010 48th Annual Allerton Conference on Communication,
  Control, and Computing (Allerton)}, pp.\  115--122, 2010.

\bibitem[Garivier \& Capp{\'e}(2011)Garivier and Capp{\'e}]{garivier2011kl}
Garivier, A. and Capp{\'e}, O.
\newblock The {KL-UCB} algorithm for bounded stochastic bandits and beyond.
\newblock In \emph{Proceedings of the 24th annual Conference On Learning Theory
  (COLT)}, pp.\  359--376, 2011.

\bibitem[Jaksch et~al.(2010)Jaksch, Ortner, and Auer]{jaksch2010near}
Jaksch, T., Ortner, R., and Auer, P.
\newblock Near-optimal regret bounds for reinforcement learning.
\newblock \emph{Journal of Machine Learning Research}, 11\penalty0
  (Apr):\penalty0 1563--1600, 2010.

\bibitem[Jordan(2010)]{JordanExponentialFamily2010}
Jordan, M.
\newblock Chapter 8: {T}he {E}xponential family: {B}asics, 2010.
\newblock URL
  \url{https://people.eecs.berkeley.edu/~jordan/courses/260-spring10/other-readings/chapter8.pdf}.

\bibitem[Kami{\'n}ski(2015)]{kaminski2015refined}
Kami{\'n}ski, B.
\newblock Refined knowledge-gradient policy for learning probabilities.
\newblock \emph{Operations Research Letters}, 43\penalty0 (2):\penalty0
  143--147, 2015.

\bibitem[Kaufmann et~al.(2012{\natexlab{a}})Kaufmann, Capp{\'e}, and
  Garivier]{kaufmann2012bayesian}
Kaufmann, E., Capp{\'e}, O., and Garivier, A.
\newblock On {B}ayesian upper confidence bounds for bandit problems.
\newblock In \emph{Artificial intelligence and statistics (AISTATS)}, pp.\
  592--600, 2012{\natexlab{a}}.

\bibitem[Kaufmann et~al.(2012{\natexlab{b}})Kaufmann, Korda, and
  Munos]{kaufmann2012thompson}
Kaufmann, E., Korda, N., and Munos, R.
\newblock Thompson sampling: an asymptotically optimal finite-time analysis.
\newblock In \emph{Proceedings of the 23rd international conference on
  Algorithmic Learning Theory}, pp.\  199--213. Springer-Verlag,
  2012{\natexlab{b}}.

\bibitem[Korda et~al.(2013)Korda, Kaufmann, and Munos]{korda2013thompson}
Korda, N., Kaufmann, E., and Munos, R.
\newblock Thompson sampling for 1-dimensional exponential family bandits.
\newblock In \emph{Advances in Neural Information Processing Systems}, pp.\
  1448--1456, 2013.

\bibitem[Kumar(1983{\natexlab{a}})]{kumar1983optimal}
Kumar, P.~R.
\newblock Optimal adaptive control of linear-quadratic-{G}aussian systems.
\newblock \emph{SIAM Journal on Control and Optimization}, 21\penalty0
  (2):\penalty0 163--178, 1983{\natexlab{a}}.

\bibitem[Kumar(1983{\natexlab{b}})]{kumar1983simultaneous}
Kumar, P.~R.
\newblock Simultaneous identification and adaptive control of unknown systems
  over finite parameter sets.
\newblock \emph{IEEE Transactions on Automatic Control}, 28\penalty0
  (1):\penalty0 68--76, 1983{\natexlab{b}}.

\bibitem[Kumar(1985)]{kumar1985survey}
Kumar, P.~R.
\newblock A survey of some results in stochastic adaptive control.
\newblock \emph{SIAM Journal on Control and Optimization}, 23\penalty0
  (3):\penalty0 329--380, 1985.

\bibitem[Kumar \& Becker(1982)Kumar and Becker]{kumar1982new}
Kumar, P.~R. and Becker, A.
\newblock A new family of optimal adaptive controllers for {M}arkov chains.
\newblock \emph{IEEE Transactions on Automatic Control}, 27\penalty0
  (1):\penalty0 137--146, 1982.

\bibitem[Kumar \& Lin(1982)Kumar and Lin]{kumar1982optimal}
Kumar, P.~R. and Lin, W.
\newblock Optimal adaptive controllers for unknown {M}arkov chains.
\newblock \emph{IEEE Transactions on Automatic Control}, 27\penalty0
  (4):\penalty0 765--774, 1982.

\bibitem[Lai \& Robbins(1985)Lai and Robbins]{lai1985asymptotically}
Lai, T.~L. and Robbins, H.
\newblock Asymptotically efficient adaptive allocation rules.
\newblock \emph{Advances in applied mathematics}, 6\penalty0 (1):\penalty0
  4--22, 1985.

\bibitem[Liu \& Li(2016)Liu and Li]{liu2016prior}
Liu, C.-Y. and Li, L.
\newblock On the prior sensitivity of {T}hompson sampling.
\newblock In \emph{International Conference on Algorithmic Learning Theory
  (ALT)}, pp.\  321--336, 2016.

\bibitem[Mandl(1974)]{mandl1974estimation}
Mandl, P.
\newblock Estimation and control in markov chains.
\newblock \emph{Advances in Applied Probability}, pp.\  40--60, 1974.

\bibitem[Prandini \& Campi(2000)Prandini and Campi]{prandini2000adaptive}
Prandini, M. and Campi, M.~C.
\newblock Adaptive {LQG} control of input-output systems---{A} cost-biased
  approach.
\newblock \emph{SIAM Journal on Control and Optimization}, 39\penalty0
  (5):\penalty0 1499--1519, 2000.

\bibitem[Rusmevichientong \& Tsitsiklis(2010)Rusmevichientong and
  Tsitsiklis]{rusmevichientong2010linearly}
Rusmevichientong, P. and Tsitsiklis, J.~N.
\newblock Linearly parameterized bandits.
\newblock \emph{Mathematics of Operations Research}, 35\penalty0 (2):\penalty0
  395--411, 2010.

\bibitem[Russo \& Van~Roy(2014)Russo and Van~Roy]{russo2014learning}
Russo, D. and Van~Roy, B.
\newblock Learning to optimize via information-directed sampling.
\newblock In \emph{Advances in Neural Information Processing Systems}, pp.\
  1583--1591, 2014.

\bibitem[Russo \& Van~Roy(2018{\natexlab{a}})Russo and
  Van~Roy]{russo2017learning}
Russo, D. and Van~Roy, B.
\newblock Learning to optimize via information-directed sampling.
\newblock \emph{Operations Research}, 66\penalty0 (1):\penalty0 230--252,
  2018{\natexlab{a}}.

\bibitem[Russo \& Van~Roy(2018{\natexlab{b}})Russo and
  Van~Roy]{russo2018learning}
Russo, D. and Van~Roy, B.
\newblock Learning to optimize via information-directed sampling.
\newblock \emph{Operations Research}, 66\penalty0 (1):\penalty0 230--252,
  2018{\natexlab{b}}.

\bibitem[Ryzhov et~al.(2010)Ryzhov, Frazier, and Powell]{ryzhov2010robustness}
Ryzhov, I.~O., Frazier, P.~I., and Powell, W.~B.
\newblock On the robustness of a one-period look-ahead policy in multi-armed
  bandit problems.
\newblock \emph{Procedia Computer Science}, 1\penalty0 (1):\penalty0
  1635--1644, 2010.

\bibitem[Ryzhov et~al.(2012)Ryzhov, Powell, and Frazier]{ryzhov2012knowledge}
Ryzhov, I.~O., Powell, W.~B., and Frazier, P.~I.
\newblock The knowledge gradient algorithm for a general class of online
  learning problems.
\newblock \emph{Operations Research}, 60\penalty0 (1):\penalty0 180--195, 2012.

\bibitem[Scott(2010)]{scott2010modern}
Scott, S.~L.
\newblock A modern {B}ayesian look at the multi-armed bandit.
\newblock \emph{Applied Stochastic Models in Business and Industry},
  26\penalty0 (6):\penalty0 639--658, 2010.

\bibitem[Srinivas et~al.(2012)Srinivas, Krause, Kakade, and
  Seeger]{srinivas2012information}
Srinivas, N., Krause, A., Kakade, S.~M., and Seeger, M.~W.
\newblock Information-theoretic regret bounds for {G}aussian process
  optimization in the bandit setting.
\newblock \emph{IEEE Transactions on Information Theory}, 58\penalty0
  (5):\penalty0 3250--3265, 2012.

\bibitem[Stettner(1993)]{stettner1993nearly}
Stettner, {\L}.
\newblock On nearly self-optimizing strategies for a discrete-time uniformly
  ergodic adaptive model.
\newblock \emph{Applied Mathematics and Optimization}, 27\penalty0
  (2):\penalty0 161--177, 1993.

\bibitem[Wainwright(2019)]{wainwright2019high}
Wainwright, M.~J.
\newblock \emph{{High-dimensional statistics: A non-asymptotic viewpoint}},
  volume~48.
\newblock Cambridge University Press, 2019.

\bibitem[Wang et~al.(2016)Wang, Wang, and Powell]{wang2016knowledge}
Wang, Y., Wang, C., and Powell, W.
\newblock The knowledge gradient for sequential decision making with stochastic
  binary feedbacks.
\newblock In \emph{International Conference on Machine Learning (ICML)}, pp.\
  1138--1147, 2016.

\bibitem[Whittle(1980)]{whittle1980multi}
Whittle, P.
\newblock Multi-armed bandits and the {G}ittins index.
\newblock \emph{Journal of the Royal Statistical Society: Series B
  (Methodological)}, 42\penalty0 (2):\penalty0 143--149, 1980.

\end{thebibliography}
}
\appendix
\newpage
\onecolumn
\section*{Appendix}

\makeatletter
\newcommand*{\diff}%
  {\@ifnextchar^{\DIfF}{\DIfF^{}}}
\makeatother
\def\DIfF^#1{%
  \mathop{\mathrm{\mathstrut d}}%
  \nolimits^{#1}\gobblespace}
\def\gobblespace{%
  \futurelet\diffarg\opspace}
\def\opspace{%
  \let\DiffSpace\!%
  \ifx\diffarg(%
    \let\DiffSpace\relax
  \else
    \ifx\diffarg[%
      \let\DiffSpace\relax
    \else
      \ifx\diffarg\{%
        \let\DiffSpace\relax
      \fi%
    \fi%
  \fi%
  \DiffSpace}
\makeatother
\newcommand*{\deriv}[3][]{%
  \frac{\diff^{#1}#2}{\diff #3^{#1}}}
\newcommand*{\pderiv}[3][]{%
  \frac{\partial^{#1}#2}%
  {\partial #3^{#1}}}

\newcommand{\equad}{\mathrel{\phantom{=}}}

\appendix

\section{Proof of the Index Strategy in (\ref{equation:BMLE for exp 2-1})}
\label{appendix:indexability}
Recall that $\widehat{\bm{\eta}}_t^{\text{RBMLE}}=(\widehat{\eta}^{\text{RBMLE}}_{t,1},\cdots,\widehat{\eta}^\text{RBMLE}_{t,N})$ is the reward-biased MLE for $\bm{\eta}$ and from (\ref{equation:BMLE for exp}) that $\widehat{\bm{\eta}}_t^{\text{RBMLE}}$ is a maximizer of the following problem:
\begin{align}
    &\max_{\bm{\eta}:\eta_j\in\mathcal{N},\forall j} \big\{L(\mathcal{H}_t;\bm{\eta}) \max_{i \in [N]}\exp(\eta_i \alpha (t))\big\}.\label{equation:BMLE for exp in appendix} 
\end{align}
Define an index set and a parameter set as
\begin{align}
    \mathcal{I}_t&:=\argmax_{i\in [N]}\{\widehat{\eta}^{\text{RBMLE}}_{t,i}\}\\
    H_{t,i}&:=\argmax_{\bm{\eta}:\eta_j\in\mathcal{N},\forall j} \big\{L(\mathcal{H}_t;\bm{\eta})\exp(\eta_i \alpha (t))\big\}\label{eq:RBMLE index set} 
\end{align}
Note that at each time $t$, RBMLE would select an arm from the index set $\mathcal{I}_t$, as shown in (\ref{equation:exchange theta and eta}).
For each arm $i$, consider an estimator $\bar{\bm{\eta}}_t^{(i)}=(\bar{{\eta}}_{t,1}^{(i)},\cdots,\bar{{\eta}}_{t,N}^{(i)})\in H_{t,i}$.
Accordingly, we further define an index set
\begin{equation}
    \mathcal{I}_t':=\argmax_{i\in [N]}\{L(\mathcal{H}_t;\bar{\bm{\eta}}_t^{(i)})\exp(\bar{{\eta}}_{t,i}^{(i)}\alpha (t))\}.\label{eq:proof of index strategy index set}
\end{equation}
Next, we show that the two index sets are identical, i.e. $\mathcal{I}_t=\mathcal{I}_t'$.
Since $L(\mathcal{H}_t;\widehat{\bm{\eta}}_t^{\text{RBMLE}})$ does not depend on $i$, we know
\begin{align}
  \argmax_{i\in [N]} \{\widehat{\eta}^{\text{RBMLE}}_{t,i}\}=\argmax_{i\in [N]} \big\{L(\mathcal{H}_t;\widehat{\bm{\eta}}_t^{\text{RBMLE}})\exp(\widehat{{\eta}}_{t,i}^{\text{RBMLE}}\alpha(t))\big\}. 
\end{align}
Moreover, we have
\begin{align}
    \max_{i\in [N]} \Big\{L(\mathcal{H}_t;\widehat{\bm{\eta}}_t^{\text{RBMLE}})\exp(\widehat{{\eta}}_{t,i}^{\text{RBMLE}}\alpha(t))\Big\}
    &=L(\mathcal{H}_t;\widehat{\bm{\eta}}_t^{\text{RBMLE}})\cdot \max_{i\in[N]} \exp(\widehat{{\eta}}_{t,i}^{\text{RBMLE}}\alpha(t))\label{eq:proof of index strategy 1}\\
    &=\max_{\bm{\eta}:\eta_j\in\mathcal{N},\forall j} \Big\{L(\mathcal{H}_t;\bm{\eta})\cdot \max_{i\in [N]}\exp(\eta_i \alpha(t)) \Big\}\label{eq:proof of index strategy 2} \\
    &=\max_{\bm{\eta}:\eta_j\in\mathcal{N},\forall j}\Big\{\max_{i\in [N]} \big\{L(\mathcal{H}_t;\bm{\eta})\cdot \exp(\eta_i \alpha(t)) \big\}\Big\}\label{eq:proof of index strategy 3}\\
    &=\max_{i\in [N]}\Big\{\max_{\bm{\eta}:\eta_j\in\mathcal{N},\forall j} \big\{L(\mathcal{H}_t;\bm{\eta})\cdot \exp(\eta_i \alpha(t)) \big\}\Big\}\label{eq:proof of index strategy 4}\\
    &=\max_{i\in [N]} \big\{L(\mathcal{H}_t;\bar{\bm{\eta}}_t^{(i)})\exp(\bar{{\eta}}_{t,i}^{(i)}\alpha (t))\big\},\label{eq:proof of index strategy 5}
\end{align}
where (\ref{eq:proof of index strategy 1}) follows from the fact that $L(\mathcal{H}_t;\widehat{\bm{\eta}}_t^{\text{RBMLE}})$ does not depend on $i$, (\ref{eq:proof of index strategy 2}) holds by the definition of $\widehat{\bm{\eta}}_t^{\text{RBMLE}}$, (\ref{eq:proof of index strategy 3})-(\ref{eq:proof of index strategy 4}) follow from that interchanging the order of the two $\max$ operations does not change the optimal value and the optimizers, and (\ref{eq:proof of index strategy 5}) follows from the definitions of $H_{t,i}$ and $\bar{\bm{\eta}}_t^{(i)}$.
By (\ref{eq:RBMLE index set}), (\ref{eq:proof of index strategy index set}), and (\ref{eq:proof of index strategy 1})-(\ref{eq:proof of index strategy 5}), we conclude that $\mathcal{I}_t=\mathcal{I}_t'$, and hence (\ref{equation:BMLE for exp 2-1}) indeed holds. 


\section{Proof of Proposition \ref{prop:BMLE index for exponential families}}
\label{appendix:prop:BMLE index for exponential families}
Recall from (\ref{equation:BMLE for exp 2-1}) that
\begin{equation}
    \pi^{\text{RBMLE}}_t 
    =\argmax_{i\in [N]}\Big\{\max_{\bm{\eta}:{\eta}_j\in\mathcal{N},\forall j}\big\{L(\mathcal{H}_t;\bm{\eta}) \exp(\eta_i\alpha (t))\big\}\Big\}\label{equation:proof of exponential family index 1}
\end{equation}
By plugging $L(\mathcal{H}_t;\bm{\eta})$ into (\ref{equation:proof of exponential family index 1}) using the density function of the Exponential Families and taking the logarithm of (\ref{equation:proof of exponential family index 1}),
\begin{align}
    \pi^{\text{BMLE}}_t=\argmax_{i\in\{1,\cdots,N\}} \bigg\{\max_{\bm{\eta}:\eta_j\in\mathcal{N},\forall j} \Big\{\underbrace{\sum_{\tau=1}^{t}\big(\eta_{\pi_\tau}X_\tau-F(\eta_{\pi_\tau})\big) + \eta_i \alpha (t)}_{=:{\ell}_i(\mathcal{H}_t;\bm{\eta})}\Big\}\bigg\}.\label{equation:proof of exponential family index 2}
\end{align}
Note that the inner maximization problem for ${\ell}_i(\mathcal{H}_t;\bm{\eta})$ over $\bm{\eta}$ is convex since $F(\cdot)$ is a convex function.
Recall that $N_i(t)$ and $S_i(t)$ denote the total number of trials of arm $i$ and the total reward collected from pulling arm $i$ up to time $t$, as defined in Section \ref{section:formulation}.
By taking the partial derivatives of ${\ell}_i(\mathcal{H}_t;\bm{\eta})$ with respect to each $\eta_i$, we know that ${\ell}_i(\mathcal{H}_t;\bm{\eta})$ is maximized when $\dot{F}(\eta_i)=\big[\frac{S_i(t)+\alpha(t)}{N_i(t)}\big]_{\Theta}$ and $\dot{F}(\eta_j)=\frac{S_j(t)}{N_j(t)}$, for $j\neq i$, where $[\cdot]_{\Theta}$ denotes the clipped value within the set $\Theta$.
For each $i=1,\cdots,N$, we then define
\begin{align}
    \eta_i^{*}&:=\dot{F}^{-1}\big(\frac{S_i(t)}{N_i(t)}\big),\\
    \eta_i^{**}&:=\dot{F}^{-1}\Big(\Big[\frac{S_i(t)+\alpha(t)}{N_i(t)}\Big]_{\Theta}\Big).\label{equation:proof of exponential family index 3}
\end{align}
By substituting $\{\eta_i^{*}\}$ and $\{\eta_i^{**}\}$ into (\ref{equation:proof of exponential family index 2}), we have
\begin{align}
    \pi^{\text{BMLE}}_t&=\argmax_{i\in\{1,\cdots,N\}}\Big\{{\ell}_i\big(\mathcal{H}_t;\eta_i^{**},\{\eta_j^{**}\}_{j\neq i}\big)\Big\}\label{equation:proof of exponential family index 4}\\
    &=\argmax_{i\in\{1,\cdots,N\}}\Big\{{\ell}_i\big(\mathcal{H}_t;\eta_i^{**},\{\eta_j^{*}\}_{j\neq i}) - {\ell}_i(\mathcal{H}_t;\{\eta_j^{*}\}_{j=1,\cdots,N}\big)\Big\}\label{equation:proof of exponential family index 5}\\
    &=\argmax_{i\in\{1,\cdots,N\}}\bigg\{\Big[\big((S_i(t)+\alpha(t))\big)\eta_i^{**}-N_i(t)F(\eta_i^{**}) \Big] - \Big[S_i(t)\eta_{i}^{*}-N_i(t)F(\eta_i^{*}) \Big]\bigg\}.\label{equation:proof of exponential family index 6}
\end{align}
By substituting $N_i(t)p_{i}(t)$ for $S_i(t)$ in (\ref{equation:proof of exponential family index 6}), we then arrive at the index as
\begin{align}
    I(p_i(t),N_i(t),\alpha(t))=\Big[\big((N_i(t)p_i(t)+\alpha(t))\big)\eta_i^{**}-N_i(t)F(\eta_i^{**}) \Big] - \Big[N_i(t)p_i(t)\eta_{i}^{*}-N_i(t)F(\eta_i^{*}) \Big].\label{equation:general BMLE index in appendix}
\end{align}
\QED

\section{Proof of Corollary \ref{corollary:Bernoulli index}}
\label{appendix:corollary:bmle Bernoulli}
Recall from (\ref{equation:general BMLE index in appendix}) that for the Exponential Family rewards, the BMLE index is
\begin{align}
    I(p_i(t),N_i(t),\alpha(t))=\Big[\big((N_i(t)p_i(t)+\alpha(t))\big)\eta_i^{**}-N_i(t)F(\eta_i^{**}) \Big] - \Big[N_i(t)p_i(t)\eta_{i}^{*}-N_i(t)F(\eta_i^{*}) \Big].\label{equation:general BMLE index in appendix duplicate}
\end{align}
For the Bernoulli case, we know $F(\eta)=\log(1+e^{\eta})$, $\dot{F}(\eta)=\frac{e^{\eta}}{1+e^{\eta}}$, $\dot{F}\inv(\theta)=\log(\frac{\theta}{1-\theta})$, and $F(\dot{F}\inv(\theta))=\log(\frac{1}{1-\theta})$.
Since $\Theta=[0,1]$ for Bernoulli rewards, we need to analyze the following two cases when substituting the above $\dot{F}\inv(\theta)$ and $F(\dot{F}\inv(\theta))$ into (\ref{equation:general BMLE index in appendix duplicate}):
\begin{itemize}[leftmargin=*]
    \item \textbf{Case 1:} $\alpha(t)<N_i(t)(1-p_i(t))$ (or equivalently $\tilde{p}_i(t)<1$)
    
We have
\begin{align}
    &I(p_i(t),N_i(t),\alpha(t))\label{equation:derive Bernoulli index 1}\\
    =&\big(N_i(t) p_i(t)+\alpha(t)\big)\log\Big(\frac{N_i(t)p_i(t)+\alpha(t)}{N_i(t)-(N_i(t)p_i(t)+\alpha(t))}\Big)-N_i(t)\log\Big(\frac{N_i(t)}{N_i(t)-(N_i(t)p_i(t)+\alpha(t))}\Big)\label{equation:derive Bernoulli index 2}\\
    &-N_i(t)p_i(t)\log\Big(\frac{N_i(t)p_i(t)}{N_i(t)-N_i(t)p_i(t)}\Big)+N_i(t)\log\Big(\frac{N_i(t)}{N_i(t)-N_i(t)p_i(t)}\Big)\label{equation:derive Bernoulli index 3}\\
    =&N_i(t)\bigg\{\big(p_i(t)+\frac{\alpha(t)}{N_i(t)}\big)\log\Big(p_i(t)+\frac{\alpha(t)}{N_i(t)}\Big)+\big(1-(p_i(t)+\frac{\alpha(t)}{N_i(t)})\big)\log\Big(1-(p_i(t)+\frac{\alpha(t)}{N_i(t)})\Big)\label{equation:derive Bernoulli index 4}\\
    &-p_i(t)\log(p_i(t))-(1-p_i(t))\log(1-p_i(t))\bigg\},\label{equation:derive Bernoulli index 5}
\end{align}
where (\ref{equation:derive Bernoulli index 4})-(\ref{equation:derive Bernoulli index 5}) are obtained by reorganizing the terms in (\ref{equation:derive Bernoulli index 2})-(\ref{equation:derive Bernoulli index 3}).
    \item \textbf{Case 2:} $\alpha(t)\geq N_i(t)(1-p_i(t))$ (or equivalently $\tilde{p}_i(t)=1$)
    
In this case, the index would be the same as the case where $p_i(t)+\alpha(t)/N_i(t)=1$. Therefore, we simply have
\begin{align}
    &I(p_i(t),N_i(t),\alpha(t))=N_i(t)\Big\{-p_i(t)\log(p_i(t))-(1-p_i(t))\log(1-p_i(t))\Big\}.\label{equation:derive Bernoulli index 6}
\end{align}
\end{itemize}
\QED


\section{Proof of Corollary \ref{corollary:bmle gaussian}}
\label{appendix:corollary:bmle gaussian}
Recall from (\ref{equation:general BMLE index in appendix}) that for the Exponential Family rewards, the BMLE index is
\begin{align}
    I(p_i(t),N_i(t),\alpha(t))=\Big[\big((N_i(t)p_i(t)+\alpha(t))\big)\eta_i^{**}-N_i(t)F(\eta_i^{**}) \Big] - \Big[N_i(t)p_i(t)\eta_{i}^{*}-N_i(t)F(\eta_i^{*}) \Big],\label{equation:general BMLE index in appendix duplicate in Gaussian}
\end{align}
where $\eta_i^{*}=\dot{F}^{-1}\big(\frac{S_i(t)}{N_i(t)}\big)$ and $\eta_i^{**}=\dot{F}^{-1}\big(\frac{S_i(t)+\alpha(t)}{N_i(t)}\big)$.
For Gaussian rewards with the same variance $\sigma^2$ for all arms, we have $F(\eta_i)=\sigma^2\eta_i^2/2$, $\dot{F}(\eta_i)=\sigma^2\eta_i$, $\dot{F}\inv(\theta_i)=\theta_i/\sigma^2$, and $F(\dot{F}\inv(\theta_i))=\theta_i^2/2\sigma^2$, for each arm $i$.
Therefore, the BMLE index becomes
\begin{align}
    &I(p_i(t),N_i(t),\alpha(t))\\
    &=\frac{S_i(t)+\alpha(t)}{\sigma^2 N_i(t)}(S_i(t)+\alpha(t))-N_i(t)\frac{\sigma^2}{2}\Big(\frac{S_i(t)+\alpha(t)}{\sigma^2 N_i(t)}\Big)^2-S_i(t)\frac{S_i(t)}{\sigma^2 N_i(t)}+N_i(t)\frac{\sigma^2}{2}\Big(\frac{S_i(t)}{\sigma^2 N_i(t)}\Big)^2\\
    &=\frac{2 S_i(t)\alpha(t)+\alpha(t)^2}{2\sigma^2 N_i(t)}.
\end{align}
Equivalently, for the Gaussian rewards, the selected arm at each time $t$ is 
\begin{align}
    \pi^{\text{BMLE}}_t=\argmax_{i\in\{1,\cdots,N\}}\Big\{p_i(t)+\frac{\alpha (t)}{2 N_i(t)}\Big\}.
\end{align}
\QED


\section{Proof of Corollary \ref{corollary:bmle exponential}}
\label{appendix:corollary:bmle exponential}
Recall from (\ref{equation:general BMLE index in appendix}) that for the Exponential Family distributions, the BMLE index is
\begin{align}
    I(p_i(t),N_i(t),\alpha(t))=\Big[\big((N_i(t)p_i(t)+\alpha(t))\big)\eta_i^{**}-N_i(t)F(\eta_i^{**}) \Big] - \Big[N_i(t)p_i(t)\eta_{i}^{*}-N_i(t)F(\eta_i^{*}) \Big],\label{equation:general BMLE index in appendix duplicate in exponential}
\end{align}
where $\eta_i^{*}=\dot{F}^{-1}\big(\frac{S_i(t)}{N_i(t)}\big)$ and $\eta_i^{**}=\dot{F}^{-1}\big(\frac{S_i(t)+\alpha(t)}{N_i(t)}\big)$.
For the exponential distribution, we have $F(\eta_i)=\log(\frac{-1}{\eta_i})$, $\dot{F}(\eta_i)=\frac{-1}{\eta_i}$, $\dot{F}\inv(\theta_i)=\frac{-1}{\theta_i}$, and $F(\dot{F}\inv(\theta_i))=\log\theta_i$, for each arm $i$.
Therefore, the BMLE index becomes
\begin{align}
    &I(p_i(t),N_i(t),\alpha(t))\\
    =&(N_i(t)p_i(t)+\alpha(t))\cdot\Big(-\frac{N_i(t)}{N_i(t)p_i(t)+\alpha(t)}\Big)-N_i(t)\log\Big(\frac{N_i(t)p_i(t)+\alpha(t)}{N_i(t)}\Big)\\
    &-\Big(N_i(t)p_i(t)\big(-\frac{1}{p_i(t)}\big)\Big)+N_i(t)\log p_i(t)\\
    =&N_i(t)\log\big(\frac{N_i(t)p_i(t)}{N_i(t)p_i(t)+\alpha(t)}\big).
\end{align}
\QED

\section{Proof of Lemma \ref{lemma:index changes with n}}
\label{appendix:lemma:index decreases with n}



(i) Recall that 
\begin{align*}
    I(\nu, n, \alpha (t))=& \big(n\nu+{\alpha(t)}\big)\dot{F}\inv\big(\nu+\frac{\alpha(t)}{n}\big)-n\nu\dot{F}\inv(\nu)-nF\Big(\dot{F}\inv\big(\nu+\frac{\alpha(t)}{n}\big)\Big)+nF\big(\dot{F}\inv(\nu)\big).
\end{align*}
By taking the partial derivative of $I(\nu, n, \alpha (t))$ with respect to $n$, we have
\begin{align}
    \pderiv{I}{n} &= \nu\dot{F}\inv\big(\nu+\frac{\alpha(t)}{n}\big)+\big(n\nu+{\alpha(t)}\big)\pderiv{\dot{F}\inv\big(\nu+\frac{\alpha(t)}{n}\big)}{n}-\nu\dot{F}\inv(\nu)\label{eq:lemma1 proof 1}\\
    &-F\Big(\dot{F}\inv\big(\nu+\frac{\alpha(t)}{n}\big)\Big)-n\dot{F}\Big(\dot{F}\inv\big(\nu+\frac{\alpha(t)}{n}\big)\Big)\pderiv{\dot{F}\inv\big(\nu+\frac{\alpha(t)}{n}\big)}{n}+F\big(\dot{F}\inv(\nu)\big)\label{eq:lemma1 proof 2}\\
    &=\nu\cdot\Big[\dot{F}\inv\big(\nu+\frac{\alpha(t)}{n}\big)-\dot{F}\inv(\nu)\Big]-\Big[F\Big(\dot{F}\inv\big(\nu+\frac{\alpha(t)}{n}\big)\Big)-F\big(\dot{F}\inv(\nu)\big)\Big].\label{eq:lemma1 proof 3}
\end{align}
Since $\dot{F}(\cdot)$ is strictly increasing for the Exponential Families, we know $\dot{F}\inv(\cdot)$ is also strictly increasing and $\dot{F}\inv(\nu+{\alpha(t)}/{n}) > \dot{F}\inv(\nu)$.
Moreover, by the strict convexity of $F(\cdot)$, we have 
\begin{align}
    F\Big(\dot{F}\inv\big(\nu+\frac{\alpha(t)}{n}\big)\Big)-F\big(\dot{F}\inv(\nu)\big)&> \Big(\dot{F}\inv\big(\nu+\frac{\alpha(t)}{n}\big)-\dot{F}\inv(\nu)\Big) \cdot \underbrace{\dot{F}\big(\dot{F}\inv(\nu)\big)}_{=\nu}.\label{eq:lemma1 proof 4}
\end{align}
Therefore, by (\ref{eq:lemma1 proof 1})-(\ref{eq:lemma1 proof 4}), we conclude that $\pderiv{I}{n}<0$ and hence $I(\nu, n, \alpha (t))$ is strictly decreasing with $n$.


(ii) Recall that 
\begin{align*}
    I(\nu, n, \alpha (t))=& \big(n\nu+{\alpha(t)}\big)\dot{F}\inv\big(\nu+\frac{\alpha(t)}{n}\big)-n\nu\dot{F}\inv(\nu)-nF\Big(\dot{F}\inv\big(\nu+\frac{\alpha(t)}{n}\big)\Big)+nF\big(\dot{F}\inv(\nu)\big).
\end{align*}
By taking the partial derivative of $I(\nu, n, \alpha (t))$ with respect to $\nu$, we have
\begin{align}
    \pderiv{I}{\nu} &=n\dot{F}\inv\big(\nu+\frac{\alpha(t)}{n}\big)+(n\nu +\alpha(t))\pderiv{\dot{F}\inv\big(\nu+\frac{\alpha(t)}{n}\big)}{\nu}-\Big(n\dot{F}\inv(\nu)+n\nu\pderiv{\dot{F}\inv(\nu)}{\nu}\Big)\\
    &-n\underbrace{\dot{F}\Big(\dot{F}\inv\big(\nu+\frac{\alpha(t)}{n}\big)\Big)}_{\leq\nu+\alpha(t)/n}\pderiv{\dot{F}\inv\big(\nu+\frac{\alpha(t)}{n}\big)}{\nu}+n\underbrace{\dot{F}\Big(\dot{F}\inv(\nu)\Big)}_{=\nu}\pderiv{\dot{F}\inv(\nu)}{\nu}\\
    &\geq n\cdot\Big[\dot{F}\inv\big(\nu+\frac{\alpha(t)}{n}\big)- \dot{F}\inv(\nu)\Big]\\
    &> 0,
\end{align}
where the last inequality follows from the fact that $\dot{F}\inv(\cdot)$ is strictly increasing for the Exponential Families.
Therefore, we can conclude that $I(\nu, n, \alpha (t))$ is strictly increasing with $\nu$, for all $\alpha(t)>0$ and for all $n>0$.

\section{Proof of Lemma \ref{lemma:I(mu1,s1,alpha(t))>I(mu2,s2,alpha(t))}}
\label{appendix:lemma:I(mu1,s1,alpha(t))>I(mu2,s2,alpha(t))}
Recall that we define 
\begin{align}
    \xi(k;\nu)=&k\Big[\big(\nu+\frac{1}{k}\big)\dot{F}\inv(\nu+\frac{1}{k})-\nu\dot{F}\inv(\nu)\Big]-k\Big[F\Big(\dot{F}\inv\big(\nu+\frac{1}{k}\big)\Big)-F(\dot{F}\inv(\nu))\Big],\label{equation:xi definition in appendix}\\
    K^{*}(\theta',\theta'')=&\inf\{k:\dot{F}\inv(\theta')>\xi(k;\theta'')\}.\label{equation:K* definition in appendix}
\end{align}

Moreover, we have $I(\mu_1,k\alpha(t),\alpha(t))=\alpha(t)\xi(k;\mu_1)$.
By Lemma \ref{lemma:index changes with n}.(i), we know that $I(\mu_1, k \alpha (t), \alpha (t))$ decreases with $k$, for all $k>0$.
Let $z=\frac{1}{k}$.
Under any fixed $\mu_1\in \Theta$ and $\alpha(t)>0$, we also know that
\begin{align}
    \lim_{k\rightarrow \infty}{\xi(k;\mu_1)}=&\lim_{z\downarrow 0}\frac{\big[\big(\mu_1+z\big)\dot{F}\inv(\mu_1+z)-\mu_1\dot{F}\inv(\mu_1)\big]-\big[F\big(\dot{F}\inv\big(\mu_1+z\big)\big)-F(\dot{F}\inv(\mu_1))\big]}{z}\label{equation:limit of I1 1}\\
    =&\lim_{z\downarrow 0}\dot{F}\inv(\mu_1+z)+(\mu_1+z)\pderiv{\dot{F}\inv(\mu_1+z)}{z}-{\dot{F}\big(\dot{F}\inv(\mu_1+z)\big)}\pderiv{\dot{F}\inv(\mu_1+z)}{z}\label{equation:limit of I1 2}\\
    =&\dot{F}\inv(\mu_1),\label{equation:limit of I1 3}
\end{align}
where (\ref{equation:limit of I1 1}) is obtained by replacing $1/k$ with $z$, and (\ref{equation:limit of I1 2}) follows from L'Hôpital's rule.
Therefore, we have
\begin{align}
    \lim_{k\rightarrow \infty}{I(\mu_1,k\alpha(t),\alpha(t))}=\alpha(t)\cdot\dot{F}\inv(\mu_1).\label{equation:limit of I1 4}
\end{align}
By Lemma \ref{lemma:index changes with n}.(i) and (\ref{equation:limit of I1 4}), we know 
\begin{equation}
    I(\mu_1, k \alpha (t), \alpha (t))\geq \alpha(t)\dot{F}\inv(\mu_1), \hspace{6pt}\text{for all } k> 0.\label{equation:I1 geq alpha(t)F dot inverse mu1}
\end{equation}




For any $n_2> K^*(\mu_1,\mu_2)\alpha(t)$, we have
\begin{align}
    I(\mu_1, n_1, \alpha (t))&\geq \alpha(t)\dot{F}\inv(\mu_1)\label{equation:lemma 3 proof 1}\\
    &\geq I(\mu_2, K^*(\mu_1,\mu_2) \alpha (t), \alpha (t))\label{equation:lemma 3 proof 2}\\
    &> I(\mu_2, n_2, \alpha (t)),\label{equation:lemma 3 proof 3}
\end{align}
where (\ref{equation:lemma 3 proof 1}) follows from (\ref{equation:I1 geq alpha(t)F dot inverse mu1}), (\ref{equation:lemma 3 proof 2}) holds from the definition of $K^*(\cdot,\cdot)$, and (\ref{equation:lemma 3 proof 3}) holds due to
Lemma \ref{lemma:index changes with n}.(i).
Finally, we show that $K^{*}(\mu_1,\mu_2)$ is finite given that $\mu_1>\mu_2$.
We consider the limit of $\xi(k;\mu_2)$ when $k$ approaches zero and again let $z=\frac{1}{k}$:
\begin{align}
    \lim_{k\downarrow 0}{\xi(k;\mu_2)}=&\lim_{z\rightarrow \infty}\frac{\big[\big(\mu_2+z\big)\dot{F}\inv(\mu_2+z)-\nu\dot{F}\inv(\mu_2)\big]-\big[F\big(\dot{F}\inv\big(\mu_2+z\big)\big)-F(\dot{F}\inv(\mu_2))\big]}{z}\label{equation:limit of I1 when k goes to infinity 1}\\
    =&\lim_{z\rightarrow \infty}\dot{F}\inv(\mu_2+z)+(\mu_2+z)\underbrace{\pderiv{\dot{F}\inv(\mu_2+z)}{z}}_{\geq 0}-\underbrace{\dot{F}\big(\dot{F}\inv(\mu_2+z)\big)}_{\leq \mu_2+z}\underbrace{\pderiv{\dot{F}\inv(\mu_2+z)}{z}}_{\geq 0}\label{equation:limit of I1 when k goes to infinity 2}\\
    \geq&\lim_{z\rightarrow \infty}\dot{F}\inv(\mu_2+z)\label{equation:limit of I1 when k goes to infinity 3}\\
    \geq &\dot{F}\inv(\mu_1),\label{equation:limit of I1 when k goes to infinity 4}
\end{align}
where (\ref{equation:limit of I1 when k goes to infinity 2}) follows from L'Hôpital's rule and (\ref{equation:limit of I1 when k goes to infinity 4}) holds due to the fact that $\dot{F}\inv$ is increasing. By (\ref{equation:limit of I1 when k goes to infinity 1})-(\ref{equation:limit of I1 when k goes to infinity 4}) and since ${\xi(k;\mu_2)}$ is continuous and strictly decreasing with $k$, we know there must exist a finite $k'\geq 0$ such that $\dot{F}\inv(\mu_1)={\xi(k';\mu_2)}$.
This implies that $K^{*}(\mu_1,\mu_2)$ is finite given that $\mu_1>\mu_2$.$\hfill$\QED

\section{Proof of Lemma \ref{lemma:I(0,s1,alpha(t))>I(mu2,s2,alpha(t))}}
\label{appendix:lemma:I(0,s1,alpha(t))>I(mu2,s2,alpha(t))}

Similar to the proof of Lemma~\ref{lemma:I(mu1,s1,alpha(t))>I(mu2,s2,alpha(t))}, we leverage the function $K^*(\cdot,\cdot)$ as defined in (\ref{equation:K* definition in appendix}).
By (\ref{equation:K* definition in appendix}), we know that for any $k> K^*(\mu_0,\mu_2)$, we have $\xi(k;\mu_2)<\dot{F}\inv(\mu_0)$. 
Therefore, if $n_2>K^*(\mu_0,\mu_2)\alpha(t)$,
\begin{align}
    I(\mu_2,n_2,\alpha(t))&< I(\mu_2,K^*(\mu_0,\mu_2),\alpha(t))\label{equation:lemma 4 proof 1}\\
    &=\alpha(t)\xi(K^*(\mu_0,\mu_2);\mu_2)\label{equation:lemma 4 proof 2}\\
    &=\alpha(t)\dot{F}\inv(\mu_0).\label{equation:lemma 4 proof 3}
\end{align}

Similarly, for any $k\leq K^*(\mu_0,\mu_1)$, we have $\xi(k;\mu_1)\geq\dot{F}\inv(\mu_0)$.
Then, if $n_1\leq K^*(\mu_0,\mu_1)\alpha(t)$, we know
\begin{align}
    I(\mu_1,n_1,\alpha(t))&\geq I(\mu_1,K^*(\mu_0,\mu_1),\alpha(t))\label{equation:lemma 4 proof 4}\\
    &=\alpha(t)\xi(K^*(\mu_0,\mu_1);\mu_1)\label{equation:lemma 4 proof 5}\\
    &=\alpha(t)\dot{F}\inv(\mu_0).\label{equation:lemma 4 proof 6}
\end{align}

Hence, by (\ref{equation:lemma 4 proof 1})-(\ref{equation:lemma 4 proof 6}), we conclude that $I(\mu_1,n_1, \alpha (t))>I(\mu_2,n_2, \alpha (t))$,  for all $n_1\leq K^*(\mu_0,\mu_1) \alpha (t)$ and $n_2> {K^*(\mu_0,\mu_2)\alpha (t)}$.$\hfill$\QED

\section{Proof of Proposition \ref{prop:regret}}
\label{appendix:prop:regret}
\noindent \textbf{Proof Sketch:}
Our target is to quantify the expected number of trials of each sub-optimal arm $a$ up to time $T$.
The regret bound proof starts with a similar demonstration as for UCB1~\cite{auer2002finite} by studying the probability of the event $\{I(p_{1}(t),N_1(t),\alpha(t))\leq I(p_{a}(t),N_a(t),\alpha(t))\}$, using the Chernoff bound for Exponential Families. 
However, it is significantly different from the original proof as the dependency between the level of exploration and the bias term $\alpha(t)$ is technically more complex, compared to the straightforward confidence interval used by the conventional UCB-type policies.
Specifically, the main challenge lies in characterizing the behavior of the RBMLE index for both regimes where $N_1(t)$ is small compared to $\alpha(t)$, as well as when it is large compared to $\alpha(t)$.
Such a challenge is handled by considering three cases separately: (i) Consider $N_1(t)>\frac{4}{D(\theta_1-\frac{\varepsilon}{2}\Delta,\theta_1)}\log t$ and apply Lemma \ref{lemma:I(mu1,s1,alpha(t))>I(mu2,s2,alpha(t))}; (ii) Consider $N_1(t)\leq \frac{4}{D(\theta_1-\frac{\varepsilon}{2}\Delta,\theta_1)}\log t$ and $N_1(t)\leq K^{*}(\theta_1-\frac{\varepsilon}{2}\Delta, \underline{\theta})\alpha(t)$ and apply Lemma \ref{lemma:I(0,s1,alpha(t))>I(mu2,s2,alpha(t))}; (iii) Use Lemma \ref{lemma:I(0,s1,alpha(t))>I(mu2,s2,alpha(t))} to show that $\{N_1(t)\leq \frac{4}{D(\theta_1-\frac{\varepsilon}{2}\Delta,\theta_1)}\log t\}$ and $\{N_1(t)> K^{*}(\theta_1-\frac{\varepsilon}{2}\Delta, \underline{\theta})\alpha(t)\}$ cannot occur simultaneously. 

To begin with, for each arm $i$, we define $p_{i,n}$ to be the empirical average reward collected in the first $n$ pulls of arm $i$.
For any Exponential Family reward distribution, the empirical mean of each arm $i$ satisfies the following concentration inequalities~\cite{korda2013thompson}: 
For any $\delta>0$,
\begin{align}
    \Prob(p_{i,n}-\theta_i\geq \delta)&\leq \exp(-nD(\theta_i+\delta,\theta_i)),\label{equation:Chernoff bound in appendix 1}\\
    \Prob(\theta_i-p_{i,n}\geq \delta)&\leq \exp(-nD(\theta_i-\delta,\theta_i)).\label{equation:Chernoff bound in appendix 2}
\end{align}
Next, for each arm $i$, we define the following confidence intervals for each pair of $n,t\in \mathbb{N}$:
\begin{align}
    \delta_{i}^{+}(n,t)&:=\inf\Big\{\delta: \exp(-n D(\theta_i+\delta,\theta_i))\leq \frac{1}{t^4}\Big\},\\
    \delta_{i}^{-}(n,t)&:=\inf\Big\{\delta: \exp(-n D(\theta_i-\delta,\theta_i))\leq \frac{1}{t^4}\Big\}.
\end{align}
Accordingly, for each arm $i$ and for each pair of $n,t\in \mathbb{N}$, we define the following events: 
\begin{align}
        G_{i}^{+}(n,t)&=\Big\{p_{i,n}-\theta_{i} \leq \delta_{i}^{+}(n,t)\Big\},\\
        G_{i}^{-}(n,t)&=\Big\{\theta_{i}-p_{i,n} \leq \delta_{i}^{-}(n,t)\Big\}.
\end{align}
By the concentration inequality considered in Section \ref{section:formulation}, we have
\begin{align}
        \Prob({G_{i}^{+}(n,t)}^\mathsf{c})&\leq  e^{-n D(\theta_i+\delta_{i}^{+}(n,t),\theta_i)}\leq \frac{1}{t^4},\\
        \Prob({G_{i}^{-}(n,t)}^\mathsf{c})&\leq  e^{-n D(\theta_i-\delta_{i}^{-}(n,t),\theta_i)}\leq \frac{1}{t^4}.
\end{align}
Consider the bias term $\alpha (t)=C_\alpha\log t$ with $C_\alpha\geq {4}/({D(\theta_1-\frac{\varepsilon}{2}\Delta,\theta_1)\cdot K^{*}(\theta_1-\frac{\varepsilon}{2}\Delta,\underline{\theta})})$ and $\varepsilon\in(0,1)$.
Recall that we assume arm 1 is the unique optimal arm.
Our target is to quantify the total number of trials of each sub-optimal arm.
Define
\begin{equation}
    Q_a(T):=\max\Big\{\frac{4}{D(\theta_a+\frac{\varepsilon}{2}\Delta_a,\theta_a)},C_{\alpha}K^{*}(\theta_1-\frac{\varepsilon}{2}\Delta_a,\theta_a+\frac{\varepsilon}{2}\Delta_a)\Big\}\log T + 1.\label{equation:proof of prop3:Qa(T)}
\end{equation}
We start by characterizing $\E[N_a(T)]$ for each $a=2,\cdots,N$:
\begin{align}
        &\E[N_a (T)]\label{equation:proof of prop3:E[Na(T)] 0}\\
        &\leq Q_a(T)+\E\bigg[\sum^T_{t=Q_a(T)+1}\mathbb{I}\big(I(p_a(t),N_a(t),\alpha (t)\geq I(p_1(t),N_1(t),\alpha (t),N_a(t)\geq Q_a(T)\big)\bigg]\label{equation:proof of prop3:E[Na(T)] 1}\\
        &=Q_a(T)+\sum^T_{t=Q_a(T)+1}\Prob\Big(I\big(p_a(t),N_a(t),\alpha (t)\big)\geq I\big(p_1(t),N_1(t),\alpha (t)\big),N_a(t)\geq Q_a(T)\Big)\label{equation:proof of prop3:E[Na(T)] 2}\\
        &\leq Q_a(T)+\sum^T_{t=Q_a(T)+1}\Prob\Big(\max_{Q_a(T)\leq n_a\leq t}I\big(p_{a,n_a},n_a,\alpha (t)\big)\geq \min_{1\leq n_1\leq t}I\big(p_{1,n_1},n_1,\alpha (t)\big)\Big)\label{equation:proof of prop3:E[Na(T)] 3}\\
        &\leq Q_a(T)+\sum^T_{t=Q_a(T)+1}\sum^t_{n_1=1}\sum^t_{n_a=Q_a(T)}\Prob\Big(I\big(p_{a,n_a},n_a,\alpha (t)\big)\geq I\big(p_{1,n_1},n_1,\alpha (t)\big)\Big)\label{equation:proof of prop3:E[Na(T)] 4}\\
        &\leq Q_a(T)+\sum^T_{t=Q_a(T)+1}\sum^t_{n_1=1}\sum^t_{n_a=Q_a(T)}\Big(\underbrace{\Prob\big(G_1^{-}(n_1,t)^\mathsf{c}\big)}_{\leq\frac{1}{t^4}}+\underbrace{\Prob\big(G_a^{+}(n_a,t)^\mathsf{c}\big)}_{\leq\frac{1}{t^4}}\Big)\label{equation:proof of prop3:E[Na(T)] 5}\\
        &+\sum^T_{t=Q_a(T)+1}\sum^t_{n_1=1}\sum^t_{n_a=Q_a(T)}\Prob\Big(I\big(p_{a,n_a},n_a,\alpha (t)\big)\geq I\big(p_{1,n_1},n_1,\alpha (t)\big),G_1^{-}(n_1,t),G_a^{+}(n_a,t)\Big)\label{equation:proof of prop3:E[Na(T)] 6}\\
        &\leq Q_a(T)+\frac{\pi^2}{3}+\sum^T_{t=Q_a(T)+1}\sum^t_{n_1=1}\sum^t_{n_a=Q_a(T)}\Prob\Big(I\big(p_{a,n_a},n_a,\alpha (t)\big)\geq I\big(p_{1,n_1},n_1,\alpha (t)\big),G_1^{-}(n_1,t),G_a^{+}(n_a,t)\Big),\label{equation:proof of prop3:E[Na(T)] 7}
\end{align}
where the last equation follows from the fact that $\sum^T_{t=Q_a(T)+1}(\frac{1}{t^2})\leq \pi^2/6$ and (\ref{equation:proof of prop3:E[Na(T)] 1}) can be obtained by taking the expectation on both sides of the first inequality of (6) in~\cite{auer2002finite} and using the fact that arm $i$ is chosen implies that $i$’s index is larger than the optimal arm’s.
Next, to provide an upper bound for (\ref{equation:proof of prop3:E[Na(T)] 7}), we need to consider the following three cases separately.
As suggested by (\ref{equation:proof of prop3:E[Na(T)] 7}), we can focus on the case where $n_a\geq Q_a(T)$.

\begin{itemize}[leftmargin=*]
    \item \textbf{Case 1:} $n_1>\frac{4}{D(\theta_1-\frac{\varepsilon}{2}\Delta,\theta_1)}\log t$
    
    Since $n_1>\frac{4}{D(\theta_1-\frac{\varepsilon}{2}\Delta,\theta_1)}\log t$, we have $p_{1,n_1}\geq\theta_1-\frac{\varepsilon}{2}\Delta$ on the event $G_1^{-}(n_1,t)$. 
    Similarly, as $n_a\geq Q_a(T)> \frac{4}{D(\theta_a+\frac{\varepsilon}{2}\Delta_a,\theta_a)}\log t$, we have $p_{a,n_a}\leq\theta_a+\frac{\varepsilon}{2}\Delta_a$ on the event $G_a^{+}(n_a,t)$. Therefore, we know
    \begin{equation}
        p_{1, n_1}-p_{a, n_a}> (1-\varepsilon){\Delta}.
    \end{equation}
    Then, we have
    \begin{align}
        I(p_{1, n_1},n_1,\alpha(t))&> I(\theta_1-\frac{\varepsilon}{2}\Delta,n_1,\alpha(t))\label{equation:proof of prop3:case1 1}\\
        &\geq I(\theta_a-\frac{\varepsilon}{2}\Delta,K^{*}(\theta_1-\frac{\varepsilon}{2}\Delta,\theta_a+\frac{\varepsilon}{2}\Delta)\alpha(t),\alpha(t))\label{equation:proof of prop3:case1 2}\\
        &\geq I(\theta_a-\frac{\varepsilon}{2}\Delta_a,K^{*}(\theta_1-\frac{\varepsilon}{2}\Delta,\theta_a+\frac{\varepsilon}{2}\Delta)\alpha(t),\alpha(t))\label{equation:proof of prop3:case1 3}\\
        &\geq I(p_{a,n_a},K^{*}(\theta_1-\frac{\varepsilon}{2}\Delta,\theta_a+\frac{\varepsilon}{2}\Delta)\alpha(t),\alpha(t))\label{equation:proof of prop3:case1 4}\\
        &\geq I(p_{a,n_a},Q_a(T),\alpha(t))\label{equation:proof of prop3:case1 5}\\
        &\geq I(p_{a,n_a},n_a,\alpha(t)),\label{equation:proof of prop3:case1 6}
    \end{align}
    where (\ref{equation:proof of prop3:case1 1}) and (\ref{equation:proof of prop3:case1 3})-(\ref{equation:proof of prop3:case1 4}) hold by Lemma \ref{lemma:index changes with n}.(i) (i.e., $K^*(\theta, \theta^{\prime})$ is strictly decreasing with respect to $\theta$ and strictly increasing with respect to $\theta^{\prime}$), (\ref{equation:proof of prop3:case1 2}) holds by Lemma \ref{lemma:I(mu1,s1,alpha(t))>I(mu2,s2,alpha(t))}, and (\ref{equation:proof of prop3:case1 5})-(\ref{equation:proof of prop3:case1 6}) follow from Lemma \ref{lemma:index changes with n}.(i).
    Hence, in Case 1, we always have $I\big(p_{1,n_1},n_1,\alpha (t)\big)> I\big(p_{a,n_a},n_a,\alpha (t)\big)$.

    \item \textbf{Case 2:} $n_1\leq \frac{4}{D(\theta_1-\frac{\varepsilon}{2}\Delta,\theta_1)}\log t$ and $n_1\leq K^{*}(\theta_1-\frac{\varepsilon}{2}\Delta, \underline{\theta})\alpha(t)$
        
    Similar to Case 1, since $n_a\geq Q_a(T)> \frac{4}{D(\theta_a,\theta_a+\frac{\varepsilon}{2}\Delta_a)}\log t$, we have $p_{a,n_a}\leq\theta_a+\frac{\varepsilon}{2}\Delta_a$ on the event $G_a^{+}(n_a,t)$.
    Moreover, as $n_1\leq K^{*}(\theta_1-\frac{\varepsilon}{2}\Delta, \underline{\theta})\alpha(t)$ and $n_a\geq Q_a(T)> K^{*}(\theta_1-\frac{\varepsilon}{2}\Delta,\theta_a+\frac{\varepsilon}{2}\Delta)\alpha(t)$, by Lemma \ref{lemma:I(0,s1,alpha(t))>I(mu2,s2,alpha(t))} we know 
    \begin{equation}
        I(\underline{\theta},n_1,\alpha(t))>I(\theta_a+\frac{\varepsilon}{2}\Delta,n_a,\alpha(t)).\label{equation:Case 2-0}
    \end{equation}
    Therefore, we obtain that
    \begin{align}
        I\big(p_{1,n_1},n_1,\alpha (t)\big)
        &>I\big(\underline{\theta},n_1,\alpha (t)\big)\label{equation:Case 2-1}\\
        &>I(\theta_a+\frac{\varepsilon}{2}\Delta,n_a,\alpha(t))\label{equation:Case 2-2}\\
        &>I\big(p_{a,n_a},n_a,\alpha (t)\big),\label{equation:Case 2-3}
    \end{align}
    where (\ref{equation:Case 2-1}) and (\ref{equation:Case 2-3}) follow from Lemma \ref{lemma:index changes with n}.(ii), and (\ref{equation:Case 2-2}) is a direct result of (\ref{equation:Case 2-0}).
    Hence, in Case 2, we still have $I\big(p_{1,n_1},n_1,\alpha (t)\big)> I\big(p_{a,n_a},n_a,\alpha (t)\big)$.
    
    \item \textbf{Case 3:} $n_1\leq \frac{4}{D(\theta_1-\frac{\varepsilon}{2}\Delta,\theta_1)}\log t$ and $n_1> K^{*}(\theta_1-\frac{\varepsilon}{2}\Delta, \underline{\theta})\alpha(t)$

    Recall that $\alpha (t)=C_\alpha\log t$ with $C_\alpha\geq {4}/({D(\theta_1-\frac{\varepsilon}{2}\Delta,\theta_1)\cdot K^{*}(\theta_1-\frac{\varepsilon}{2}\Delta,\underline{\theta})})$.
    Therefore, the two events $\{n_1\leq \frac{4}{D(\theta_1-\frac{\varepsilon}{2}\Delta,\theta_1)}\log t\}$ and $\{n_1> K^{*}(\theta_1-\frac{\varepsilon}{2}\Delta, \underline{\theta})\alpha(t)\}$ cannot happen at the same time.
\end{itemize}
To sum up, in all the above three cases, we have
\begin{equation}
     \Prob\Big(I\big(p_{a,n_a},n_a,\alpha (t)\big)\geq I\big(p_{1,n_1},n_1,\alpha (t)\big),G_1^{-}(n_1,t),G_a^{+}(n_a,t)\Big)=0.\label{equation:combine case 1-3}
\end{equation}
By (\ref{equation:proof of prop3:E[Na(T)] 7}) and (\ref{equation:combine case 1-3}), we conclude that $E[N_a(T)]\leq Q_a(T)+\frac{\pi^2}{3}$, for every $a\neq 1$.


Finally, the total regret can be upper bounded as
\begin{align}
      \mathcal{R}(T)&\leq \sum_{a=2}^{N}\Delta_a\cdot E[N_a(T)]\\
      &=\sum_{a=2}^{N}\Delta_a\bigg[\max\Big\{\frac{4}{D(\theta_a+\frac{\varepsilon}{2}\Delta_a,\theta_a)},C_{\alpha}K^{*}(\theta_1-\frac{\varepsilon}{2}\Delta_a,\theta_a+\frac{\varepsilon}{2}\Delta_a)\Big\}\log T + 1 + \frac{\pi^2}{3}\bigg]. 
\end{align}
\QED

\section{Proof of Proposition \ref{prop:regret Gaussian}}
\label{appendix:prop:regret Gaussian}
\noindent \textbf{Proof Sketch:}
We extend the proof procedure of Proposition \ref{prop:regret} for Gaussian rewards, with the help of Hoeffding's inequality.
We then prove an additional lemma, which shows that conditioned on the ``good'' events, the RBMLE index of the optimal arm (i.e., arm 1) is always larger than that of a sub-optimal arm $a$ if $N_a(t)\geq \frac{2}{\Delta_a}\alpha(t)$ and $\alpha(t)\geq \frac{256\sigma^2}{\Delta_a}$, regardless of $N_1(t)$.

We extend the proof of Proposition \ref{prop:regret} to the case of Gaussian rewards.
To begin with, we define the confidence intervals and the ``good'' events.
Recall that for each arm $i$, we define $p_{i,n}$ to be the empirical average reward collected in the first $n$ pulls of arm $i$.
For each arm $i$, for each pair of $n,t\in \mathbb{N}$, we define
\begin{align}
    \delta_{i}(n,t)&:=\inf\Big\{\delta: \max\big\{\exp(-n D(\theta_i+\delta,\theta_i)),\exp(-n D(\theta_i-\delta,\theta_i))\big\}\leq \frac{1}{t^{4}}\Big\}.
\end{align}
Accordingly, for each arm $i$ and for each pair of $n,t\in \mathbb{N}$, we define the following events: 
\begin{align}
        G_{i}(n,t)&=\Big\{\lvert p_{i,n}-\theta_{i}\rvert \leq \delta_{i}(n,t)\Big\},
\end{align}
For the Gaussian rewards, we can leverage Hoeffding's inequality for sub-Gaussian distributions as follows:
\begin{lemma}
\label{lemma:Hoeffding for Gaussian}
Under $\sigma$-sub-Gaussian rewards for all arms, for any $n\in\mathbb{N}$, we have
\begin{equation}
    \Prob(\lvert p_{i,n}-\theta_{i}\rvert\geq \delta)\leq 2\exp(-\frac{n}{2\sigma^2}\delta^2).
\end{equation}
\end{lemma}
\noindent \textbf{Proof of Lemma \ref{lemma:Hoeffding for Gaussian}}: This is a direct result of Proposition 2.5 in~\cite{wainwright2019high}.\QED

Based on Lemma \ref{lemma:Hoeffding for Gaussian}, we focus on the case $D(\theta',\theta'')=\frac{1}{2\sigma^2}(|\theta'-\theta''|)^2$ and $\delta_{i}(n,t)=\sqrt{({{8\sigma^2}\log t})/{n}}$.
For ease of notation, we use $\gamma_{*}$ to denote the constant $8\sigma^2$.

Before providing the regret analysis, we first introduce the following useful lemma.
\begin{lemma}
\label{lemma:revised lemma 4 for Gaussian}
Suppose $\gamma>0$ and $\mu_1,\mu_2\in\mathbb{R}$ with $\mu_1>\mu_2$. Given $\alpha(t)=c\log t$ with $c\geq \frac{32\gamma}{\mu_1-\mu_2}$, for any $n_2\geq \frac{2}{\mu_1-\mu_2}\alpha(t)$ and any $n_1>0$, we have $I(\mu_1-\sqrt{{(\gamma\log t)}/{n_1}},n_1,\alpha(t))>I(\mu_2+\sqrt{{(\gamma\log t)}/{n_2}},n_2,\alpha(t))$.
\end{lemma}
\noindent \textbf{Proof of Lemma \ref{lemma:revised lemma 4 for Gaussian}}: 
We start by considering $n_2\geq M\alpha(t)$, for some $M>0$.
Then, note that
\begin{align}
    I\Big(\mu_1-\sqrt{\frac{\gamma\log t}{n_1}},n_1,\alpha(t)\Big)&=\mu_1-\sqrt{\frac{\gamma\log t}{n_1}}+\frac{\alpha(t)}{2n_1},\\
    I\Big(\mu_2+\sqrt{\frac{\gamma\log t}{n_2}},n_2,\alpha(t)\Big)&=\mu_2+\sqrt{\frac{\gamma\log t}{n_2}}+\frac{\alpha(t)}{2n_2}.
\end{align}
For ease of notation, we use $x_1$ and $x_2$ to denote $\sqrt{{(\gamma\log t)}/{n_1}}$ and $\sqrt{{(\gamma\log t)}/{n_2}}$, respectively.
Then, we know
\begin{align}
    I\Big(\mu_1-\sqrt{\frac{\gamma\log t}{n_1}},n_1,\alpha(t)\Big)-I\Big(\mu_2+\sqrt{\frac{\gamma\log t}{n_1}},n_2,\alpha(t)\Big)
    &\geq (\mu_1-\mu_2)-(x_1+x_2)+\frac{c}{2\gamma}(x_1^2-x_2^2)\label{equation:revised lemma 4 for Gaussian 1}\\
    &\geq (\mu_1-\mu_2)-x_1-\sqrt{\frac{\gamma}{cM}}+\frac{c}{2\gamma}x_1^2-\frac{1}{2M}\label{equation:revised lemma 4 for Gaussian 2},
\end{align}
where (\ref{equation:revised lemma 4 for Gaussian 2}) follows from $n_2\geq M\alpha(t)$.
Define $w(x_1):=(\mu_1-\mu_2)-x_1-\sqrt{\frac{\gamma}{cM}}+\frac{c}{2\gamma}x_1^2-\frac{1}{2M}$. The quadratic polynomial $w(x_1)$ remains positive for all $x_1\in\mathbb{R}$ if the discriminant of $w(x_1)$, denoted by $\text{Disc}(w(x_1))$, is negative. Indeed, we have
\begin{align}
    \text{Disc}(w(x_1))=1-4\cdot\frac{c}{2\gamma}\cdot(-\sqrt{\frac{\gamma}{cM}}-\frac{1}{2M}+(\mu_1-\mu_2))\leq -39,
\end{align}
where the last inequality follows from $c\geq \frac{32\gamma}{\mu_1-\mu_2}$ and $M=\frac{2}{\mu_1-\mu_2}$.\QED

\vspace{2mm}

Now, we are ready to prove Proposition \ref{prop:regret Gaussian}:
Consider the bias term $\alpha(t)=C_{\alpha}\log t$ with $C_{\alpha}\geq \frac{32\gamma^*}{\Delta}$, where $\gamma^*=8\sigma^2$.
Recall that we assume arm 1 is the unique optimal arm.
Our target is to quantify the total number of trials of each sub-optimal arm.
Next, we characterize the expected total number of trials of each sub-optimal arm, i.e., $\E[N_a(T)]$. We define $Q_a^{*}(T)=\frac{2}{\Delta_a}C_{\alpha}\log T$.
By using a similar argument to (\ref{equation:proof of prop3:E[Na(T)] 0})-(\ref{equation:proof of prop3:E[Na(T)] 7}), we have
\begin{align}
        &\E[N_a (T)]\leq Q_a^*(T)+\sum^T_{t=Q_a^*(T)+1}\Prob\Big(I\big(p_a(t),N_a(t),\alpha (t)\big)\geq I\big(p_1(t),N_1(t),\alpha (t)\big),N_a(t)\geq Q_a^*(T)\Big)\label{equation:regret proof Gaussian:E[Na(T)] 1}\\
        &\leq Q_a^*(T)+\sum^T_{t=Q_a^*(T)+1}\sum^t_{n_1=1}\sum^t_{n_a=Q_a^*(T)}\Prob\Big(I\big(p_{a,n_a},n_a,\alpha (t)\big)\geq I\big(p_{1,n_1},n_1,\alpha (t)\big)\Big)\label{equation:regret proof Gaussian:E[Na(T)] 2}\\
        &\leq Q_a^*(T)+\sum^T_{t=Q_a^*(T)+1}\sum^t_{n_1=1}\sum^t_{n_a=Q_a^*(T)}\Big(\underbrace{\Prob\big(G_1(n_1,t)^\mathsf{c}\big)}_{\leq\frac{2}{t^4}}+\underbrace{\Prob\big(G_a(n_a,t)^\mathsf{c}\big)}_{\leq\frac{2}{t^4}}\Big)\label{equation:regret proof Gaussian:E[Na(T)] 3}\\
        &+\sum^T_{t=Q_a^*(T)+1}\sum^t_{n_1=1}\sum^t_{n_a=Q_a^*(T)}\Prob\Big(I\big(p_{a,n_a},n_a,\alpha (t)\big)\geq I\big(p_{1,n_1},n_1,\alpha (t)\big),G_1(n_1,t),G_a(n_a,t)\Big)\label{equation:regret proof Gaussian:E[Na(T)] 4}\\
        &\leq Q_a^*(T)+\frac{2\pi^2}{3}+\sum^T_{t=Q_a^*(T)+1}\sum^t_{n_1=1}\sum^t_{n_a=Q_a^*(T)}\Prob\Big(I\big(p_{a,n_a},n_a,\alpha (t)\big)\geq I\big(p_{1,n_1},n_1,\alpha (t)\big),G_1(n_1,t),G_a(n_a,t)\Big).\label{equation:regret proof Gaussian:E[Na(T)] 5}
\end{align}
Conditioned on the events $G_i(n_1,t)$ and $G_a(n_a,t)$, we obtain that
\begin{align}
    I\big(p_{1,n_1},n_1,\alpha (t)\big)
    &\geq I(\theta_1-\sqrt{{(\gamma_{*}\log t)}/{n_1}},n_1,\alpha(t))\label{equation:Gaussian regret proof Case 2-1}\\
    &>I(\theta_a+\sqrt{{(\gamma_{*}\log t)}/{n_a}},n_a,\alpha(t))\label{equation:Gaussian regret proof Case 2-2}\\
    &\geq I\big(p_{a,n_a},n_a,\alpha (t)\big),\label{equation:Gaussian regret proof Case 2-3}
\end{align}
where (\ref{equation:Gaussian regret proof Case 2-1}) and (\ref{equation:Gaussian regret proof Case 2-3}) follow from Lemma \ref{lemma:index changes with n}.(i), and (\ref{equation:Gaussian regret proof Case 2-2}) follows from Lemma \ref{lemma:revised lemma 4 for Gaussian}.
Hence, for $n_1>0$ and $n_a\geq Q_a^*(T)$,
\begin{equation}
     \Prob\Big(I\big(p_{a,n_a},n_a,\alpha (t)\big)\geq I\big(p_{1,n_1},n_1,\alpha (t)\big),G_1(n_1,t),G_a(n_a,t)\Big)=0.\label{equation:regret proof Gaussian combine case 1-3}
\end{equation}
By (\ref{equation:regret proof Gaussian:E[Na(T)] 5}) and (\ref{equation:regret proof Gaussian combine case 1-3}), we know $E[N_a(T)]\leq Q_a^*(T)+\frac{2\pi^2}{3}$, for every $a\neq 1$.
Hence, the total regret can be upper bounded as
\begin{align}
      \mathcal{R}(T)&\leq\sum_{a=2}^{N}\Delta_a\big[\frac{2}{\Delta_a}C_{\alpha}\log T + \frac{2\pi^2}{3}\big]. 
\end{align}
\QED

\section{Proof of Proposition \ref{prop:BMLE regret for sub-exponential}} \label{appendix:prop:regret sub-exponential}
The proof of Proposition \ref{prop:regret} can be easily extended to Proposition \ref{prop:BMLE regret for sub-exponential} by replacing the Chernoff bound with the sub-Exponential tail bound. For sub-exponential reward distributions, we consider the sub-exponential tail bound as follows:
\begin{lemma}
Under $(\rho,\kappa)$-sub-exponential rewards for all arms, for any $n\in\mathbb{N}$, we have
\begin{equation}
    \Prob(p_{i,n}-\theta_i\geq \delta)\leq \exp\Big(-\frac{n^2\delta^2}{2(n\kappa\delta+\rho^2)}\Big). \label{equation:Bernstein in appendix}
\end{equation}
\end{lemma}
Similar to the proof of Proposition \ref{prop:regret}, we consider the bias term $\alpha (t)=C_\alpha\log t$, but with $C_{\alpha}\geq {16(\kappa\varepsilon\Delta+2\rho^2)}/((\varepsilon\Delta)^2 K^{*}(\theta_1-\frac{\varepsilon \Delta}{2},\underline{\theta}))$. 
Note that here we simply replace ${D(\theta_1-\frac{\varepsilon\Delta}{2},\theta_1)}$ with $\frac{(\varepsilon\Delta)^2}{4(\kappa\varepsilon\Delta+2\rho^2)}$ by comparing (\ref{equation:Bernstein in appendix}) with (\ref{equation:Chernoff bound in appendix 1}).
Similarly, we define
\begin{equation}
    \tilde{Q}_a(T):=\max\Big\{\frac{16(\kappa\varepsilon\Delta+2\rho^2)}{(\varepsilon\Delta_a)^2},C_{\alpha}K^{*}(\theta_1-\frac{\varepsilon}{2}\Delta_a,\theta_a+\frac{\varepsilon}{2}\Delta_a)\Big\}\log T + 1.\label{equation:regret proof of sub-exponential:Qa(T)}
\end{equation}
Note that the proof of Proposition \ref{prop:regret} relies only on Lemmas \ref{lemma:index changes with n}-\ref{lemma:I(0,s1,alpha(t))>I(mu2,s2,alpha(t))}, and these lemmas are tied to the distributions for deriving the RBMLE index, not to the underlying true reward distributions.
Therefore, it is easy to verify that the same proof procedure still holds here by replacing $Q_a(T)$ with $\tilde{Q}_a(T)$.\QED

\section{Proof of Proposition \ref{prop:regret adaptive Gaussian}} \label{appendix:prop:regret adaptive Gaussian}
\noindent \textbf{Proof Sketch:}
An $O(\log T)$ regret bound can be obtained by considering the extensions as follows:
\vspace{-2mm}
\begin{itemize}[leftmargin=*]
    \item By extending Lemma \ref{lemma:revised lemma 4 for Gaussian}, we show that for any two arms $i$ and $j$, there exist constants $M_1>0$ and $M_2>0$ such that $I_i > I_j$ for any $n_i\leq M_1 \log T$ and $n_j \geq M_2 \log T$.
    \vspace{-1mm}
    \item We then extend (\ref{equation:regret proof Gaussian:E[Na(T)] 1})-(\ref{equation:regret proof Gaussian:E[Na(T)] 2}) by using the fact that arm $a$ is chosen implies that its index is larger than all the other arm’s.
    \vspace{-1mm}
    \item Extend (\ref{equation:regret proof Gaussian:E[Na(T)] 3})-(\ref{equation:regret proof Gaussian:E[Na(T)] 5}) by considering the good events across all arms (instead of just arm $a$ and the optimal arm).
    \vspace{-1mm}
    \item Finally, we extend (\ref{equation:Gaussian regret proof Case 2-1})-(\ref{equation:Gaussian regret proof Case 2-3}) by using Lemma \ref{lemma:revised lemma 4 for Gaussian} and the fact that under the good events, the estimated $\hat{\Delta}_t$ is between $\Delta/2$ and $\Delta$.
\end{itemize}

In Algorithm \ref{alg:pseudo code Gaussian}, since gradually learning the minimal gap $\Delta$ involves all the arms (see Line 7 of Algorithm \ref{alg:pseudo code Gaussian}), we need to extend Lemma \ref{lemma:revised lemma 4 for Gaussian} to remove the assumption $\mu_1>\mu_2$. The extension is conducted in Lemma \ref{lemma:extended lemma for adaptive Gaussian} below.

\begin{lemma}
\label{lemma:extended lemma for adaptive Gaussian}
Let $\gamma$ be a positive constant. For any two arms $i$ and $j$ with $\mu_i,\mu_j\in \mathbb{R}$ and $\delta:=\mu_j-\mu_i$, given $\alpha(t)=c\log t$ with $c\geq \frac{32\gamma (N+2)}{\Delta}$, then for any $n_i\leq \frac{1}{8}M \frac{\Delta}{\max (\delta, \Delta)}\log(t)$ and $n_j\geq M \log t$, we have $I(\mu_i-\sqrt{{(\gamma\log t)}/{n_i}},n_i,\alpha(t))>I(\mu_j+\sqrt{{(\gamma\log t)}/{n_j}},n_j,\alpha(t))$, where $M=\frac{32\gamma (N+2)}{\Delta^2}$.
\end{lemma}
\noindent \textbf{Proof of Lemma \ref{lemma:extended lemma for adaptive Gaussian}}:
For ease of notation, we use $I_i$ and $I_j$ to denote $I(\mu_i-\sqrt{{(\gamma\log t)}/{n_i}},n_i,\alpha(t))$ and $I(\mu_j+\sqrt{{(\gamma\log t)}/{n_j}},n_j,\alpha(t))$, respectively,
within this proof. Then, we have
\begin{align}
    I_i&=\mu_i-\sqrt{\frac{\gamma\log t}{n_i}}+\frac{\alpha(t)}{2n_i},\\
    I_j&=\mu_j+\sqrt{\frac{\gamma\log t}{n_j}}+\frac{\alpha(t)}{2n_j}.
\end{align}
Using $x$ to denote $\sqrt{{(\gamma\log t)}/{n_i}}$, we have
\begin{align}
    I_i-I_j
    &\geq (\mu_i-\mu_j)-x-\sqrt{\frac{\gamma}{M}}+\frac{c}{2\gamma}x^2-\frac{c}{2\gamma}\Big(\sqrt{\frac{\gamma}{M}}\Big)^2 \label{equation: extension for gaussian 1}\\
    &= -\delta-\sqrt{\frac{\gamma}{M}}-\frac{c}{2M}+\frac{c}{2\gamma}x^2-x\label{equation: extension for gaussian 2},
\end{align}
where (\ref{equation: extension for gaussian 1}) follows from $n_j\geq M\log(t)$. Since $n_i\leq \frac{1}{8}M \frac{\Delta}{\max (\delta, \Delta)}\log(t)$, we have
\begin{align}
    x=\sqrt{\frac{\gamma\log t}{n_i}}\geq \sqrt{\frac{8\gamma\log t}{\frac{32\gamma(N+2)}{\Delta^2}\frac{\Delta}{\max(\delta,\Delta)}}}=\sqrt{\frac{\Delta\cdot\max(\delta,\Delta)}{4(N+2)}}\label{equation: extension for gaussian 3}.
\end{align}
By (\ref{equation: extension for gaussian 3}) and the fact that the ${c x^2}/({2\gamma})-x$ has its minimum at $x=\gamma/c\leq \Delta/(32(N+2))$, we can construct a lower bound for ${c x^2}/({2\gamma})-x$ in (\ref{equation: extension for gaussian 2}):
\begin{align}
    \frac{c}{2\gamma}x^2-x\geq \frac{c}{2\gamma}\frac{8\gamma}{M}\frac{\max(\delta,\Delta)}{\Delta}-\sqrt{\frac{8\gamma}{M}\frac{\max(\delta,\Delta)}{\Delta}}.
\end{align}
Then we can obtain a lower bound of $I_i-I_j$:
\begin{align}
    I_i-I_j&\geq -\delta-\sqrt{\frac{\gamma}{M}}-\frac{c}{2M}+\frac{c}{2M}\cdot 8\frac{\max(\delta,\Delta)}{\Delta}-\sqrt{\frac{8\gamma}{M} \cdot \frac{\max(\delta,\Delta)}{\Delta}} \\
    &= -\delta-\Big(\sqrt{\frac{\gamma}{M}} + \sqrt{\frac{8\gamma}{M} \cdot \frac{\max(\delta,\Delta)}{\Delta}}\Big) - \Big(\frac{c}{2M}-\frac{c}{2M}\cdot 8\frac{\max(\delta,\Delta)}{\Delta}\Big) \\
    &\geq -\delta-(\sqrt{8}+1)\sqrt{\frac{\gamma}{M} \cdot \frac{\max(\delta,\Delta)}{\Delta}}+\frac{c}{2M}\cdot 7 \cdot \frac{\max(\delta,\Delta)}{\Delta}\label{equation: extension for gaussian 30}\\
    &\geq -\delta-(\sqrt{8}+1)\sqrt{\frac{\Delta\max(\delta,\Delta)}{32(N+2)}}+\frac{7}{2}\max(\delta,\Delta),\label{equation: extension for gaussian 4}\\
    &> 0,
\end{align}
where (\ref{equation: extension for gaussian 30})-(\ref{equation: extension for gaussian 4}) follow from that $c\geq \frac{32\gamma(N+2)}{\Delta}$ and $M=\frac{32\gamma(N+2)}{\Delta^2}$. \QED

\noindent \textbf{Proof of Proposition \ref{prop:regret adaptive Gaussian}}: First, we set $\gamma=8\sigma^2$. Recall that $T_0:=\min\{t\in\mathbb{N}:\beta(t)\geq \frac{32\gamma (N+2)}{\Delta}\}<\infty$.
Within this proof, we take $\delta_i(n,t)=\sqrt{(2\sigma^2 (N+2)\log t)/n}$ and define the good events as $G_i(n,t):=\{\lvert p_{i,n}-\theta_i\rvert \leq \delta_{i}(n,t)\}$.
By Lemma \ref{lemma:Hoeffding for Gaussian}, we know $\Prob(G_i(n_i,t)^c)\leq 2/t^{N+2}$, for all $t$ and $i$.
We also define $Q_a^{*}(T):=\max\{4\cdot\frac{32\gamma (N+2)}{\Delta^2}\log T, T_0\}$.
Denote by $\E[N_a(T)]$ the expected total number of trials of arm $a$. Then for each $a$, we can construct an upper bound of $\E[N_a(T)]$ by:
\begin{align}
        &\E[N_a (T)]\leq Q_a^*(T)+\sum^T_{t=Q_a^*(T)+1}\Prob\Big(I\big(p_a(t),N_a(t),\alpha (t)\big)\geq I\big(p_i(t),N_i(t),\alpha (t)\big),\forall i\neq a, N_a(t)\geq Q_a^*(T)\Big)\\
        &\leq Q_a^*(T)+\sum^T_{t=Q_a^*(T)+1}\sum^t_{\substack{n_i=1\\ i\neq a}}\sum^t_{n_a=Q_a^*(T)}\Prob\Big(I\big(p_{a,n_a},n_a,\alpha (t)\big)\geq I\big(p_{i,n_i},n_i,\alpha (t),\forall i\neq a\big)\Big)\\
        &\leq Q_a^*(T)+\sum^T_{t=Q_a^*(T)+1}\sum^t_{\substack{n_i=1\\ i\neq a}}\sum^t_{n_a=Q_a^*(T)}\Big(\underbrace{\sum_i^N\Prob\big( G_i(n_i,t)^c\big)}_{\leq 2N/t^{N+2}}\Big)\\
        &+\sum^T_{t=Q_a^*(T)+1}\sum^t_{\substack{n_i=1\\ i\neq a}}\sum^t_{n_a=Q_a^*(T)}\Prob\Big(I\big(p_{a,n_a},n_a,\alpha (t)\big)\geq I\big(p_{i,n_i},n_,\alpha (t)\big),\forall i\neq a,G_1(n_1,t),G_2(n_2,t),\cdots,G_N(n_N,t)\Big)\\
        &\leq Q_a^*(T)+\frac{N\pi^2}{3}\\
        &+\sum^T_{t=Q_a^*(T)+1}\sum^t_{n_1=1}\sum^t_{n_a=Q_a^*(T)}\Prob\Big(I\big(p_{a,n_a},n_a,\alpha (t)\big)\geq I\big(p_{i,n_i},n_i,\alpha (t)\big),\forall i\neq a,G_1(n_1,t),G_2(n_2,t),\cdots,G_N(n_N,t)\Big).
\end{align}

\begin{itemize}[leftmargin=*]
    \item \textbf{Case 1:} $n_i\geq \frac{1}{8}\frac{32\gamma(N+2)}{\Delta^2}\frac{\Delta}{\max(\theta_a-\theta_i,\Delta)}\log t$ for all $i\neq a$. Since $n_i$ is large enough, it is easy to check that the following inequality holds under the good events for all $i\neq a$:
    \begin{align}
        |p_{i,n_i}-\theta_i|\leq \sqrt{\frac{2\sigma^2 \log t}{\frac{1}{8}\frac{32\gamma(N+2)}{\Delta^2}\frac{\Delta}{\max(\theta_a-\theta_i)}\log t}}\leq \frac{\max(\theta_a-\theta_i,\Delta)}{4\sqrt{N+2}}\leq \frac{\max(\theta_a-\theta_i,\Delta)}{8}.
    \end{align}
    Similarly, we have $|p_{a,n_a}-\theta_a|\leq \frac{\Delta}{8\sqrt{2}\sqrt{N+2}}$. Moreover, by checking $L_i(t)$ and $U_i(t)$ under the good events, we also know $\widehat{\Delta}_t\geq\frac{\Delta}{2}$.
    This also implies that $\widehat{C}_{\alpha}(t)\leq \frac{32\gamma(N+2)}{\Delta/2}$ and hence $\alpha(t)\leq \frac{32\gamma(N+2)}{\Delta/2}\log t$. Therefore by Lemma \ref{lemma:revised lemma 4 for Gaussian}, we know that $I\big(p_{1,n_1},n_1,\alpha (t)\big)\geq I\big(p_{a,n_a},n_a,\alpha (t))$.

     \item \textbf{Case 2:} There exists some $i\neq a$ such that $n_i < \frac{1}{8}\frac{32\gamma(N+2)}{\Delta^2}\frac{\Delta}{\max(\theta_a-\theta_i,\Delta)}\log t$. By Lemma \ref{lemma:extended lemma for adaptive Gaussian}, this implies that $I\big(p_{i,n_i},n_i,\alpha (t)\big)> I\big(p_{a,n_a},n_a,\alpha (t))$.
\end{itemize}
Therefore, we have for every $a\neq 1$,
\begin{align}
    E[N_a(T)]&\leq Q_a^*(T)+\frac{N\pi^2}{3} \\
    &=\max\Big\{\frac{128\gamma(N+2)}{\Delta^2}\log T, T_0 \Big\}+ \frac{N\pi^2}{3}.
\end{align}
Hence, the total regret can be upper bounded as
\begin{align}
      \mathcal{R}(T)&\leq\sum_{a=2}^{N}\Delta_a\Big[\max\Big\{\frac{128\gamma(N+2)}{\Delta^2}\log T, T_0 \Big\} + \frac{N\pi^2}{3}\Big]. 
\end{align} \QED

\section{Additional Empirical Results}
\label{appendix:additional empirical}
In this subsection, we present additional empirical results for more examples to demonstrate the effectiveness, efficiency and scalability of the proposed RBMLE algorithm. 

\subsection{Effectiveness}
\label{appendix:more empirical results on effectiveness}
\begin{figure*}[h]
\centering
$\begin{array}{c c c}
    \multicolumn{1}{l}{\mbox{\bf }} & \multicolumn{1}{l}{\mbox{\bf }} & \multicolumn{1}{l}{\mbox{\bf }} \\ 
    \hspace{-1mm} \scalebox{0.33}{\includegraphics[width=\textwidth]{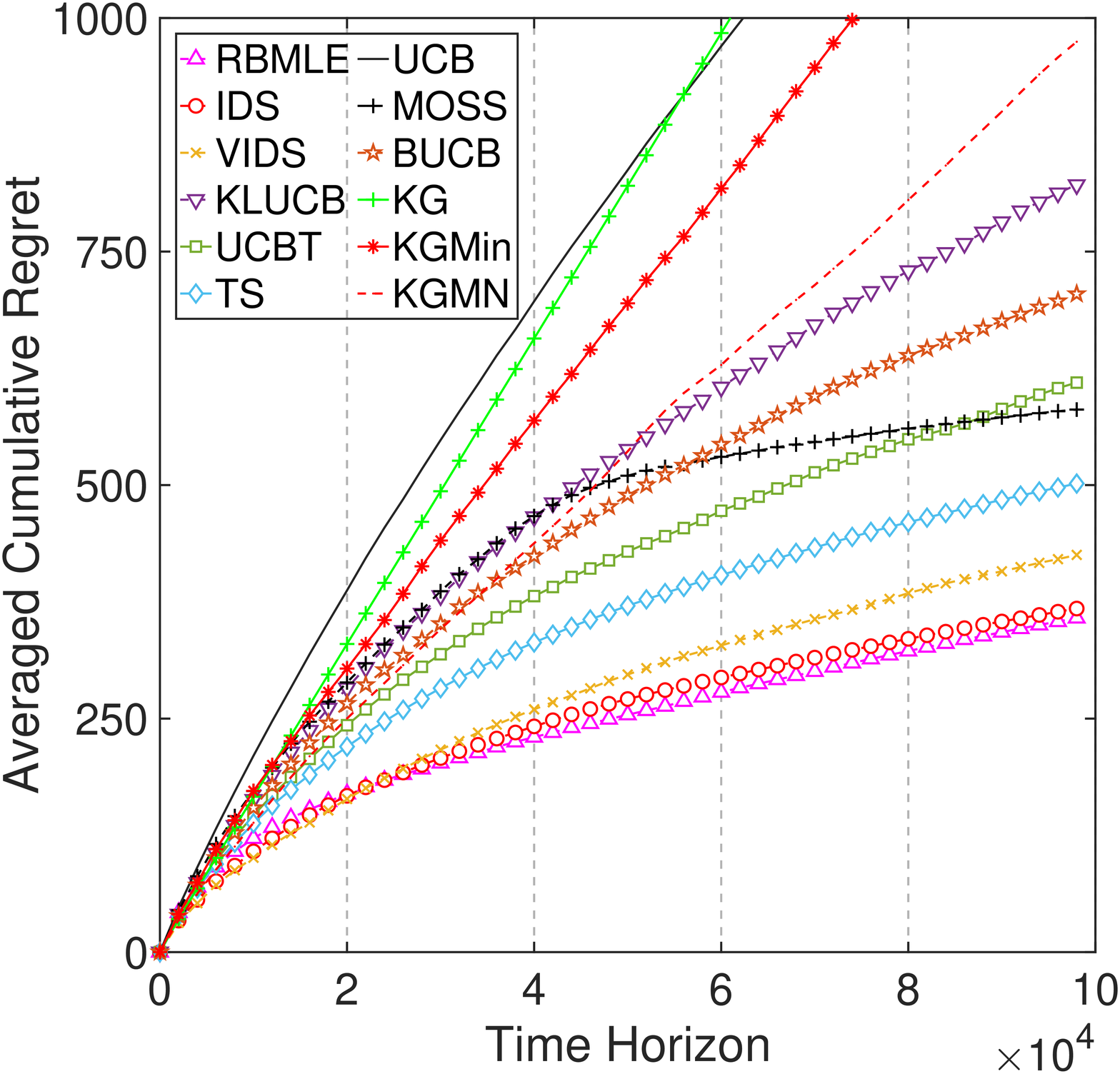}}  \label{fig:Bernoulli_ID=3_Regret_T=1e5} & 
    \hspace{-3mm} \scalebox{0.33}{\includegraphics[width=\textwidth]{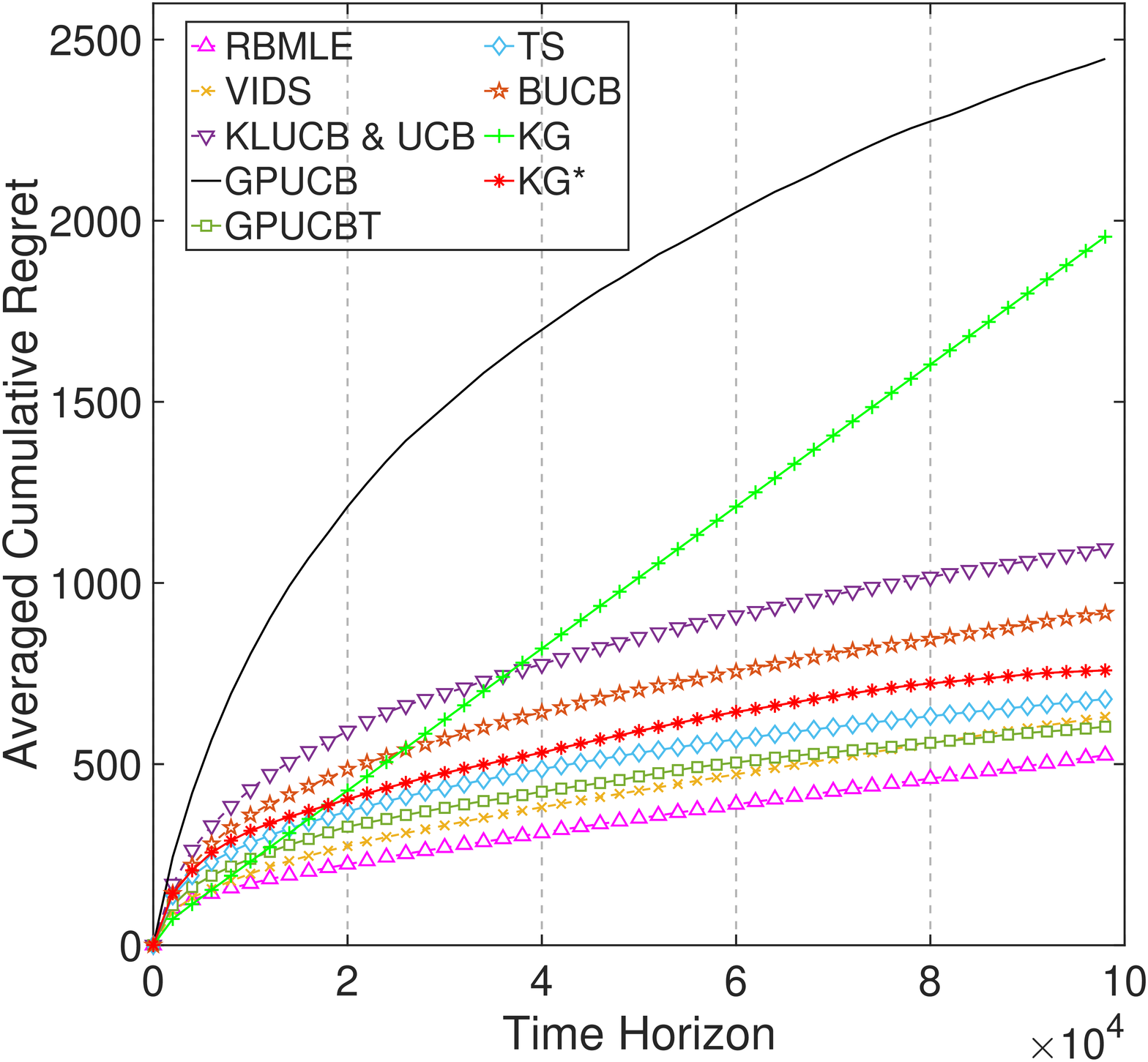}} \label{fig:Gaussian_ID=27_Regret_T=1e5} & \hspace{-3mm}
    \scalebox{0.33}{\includegraphics[width=\textwidth]{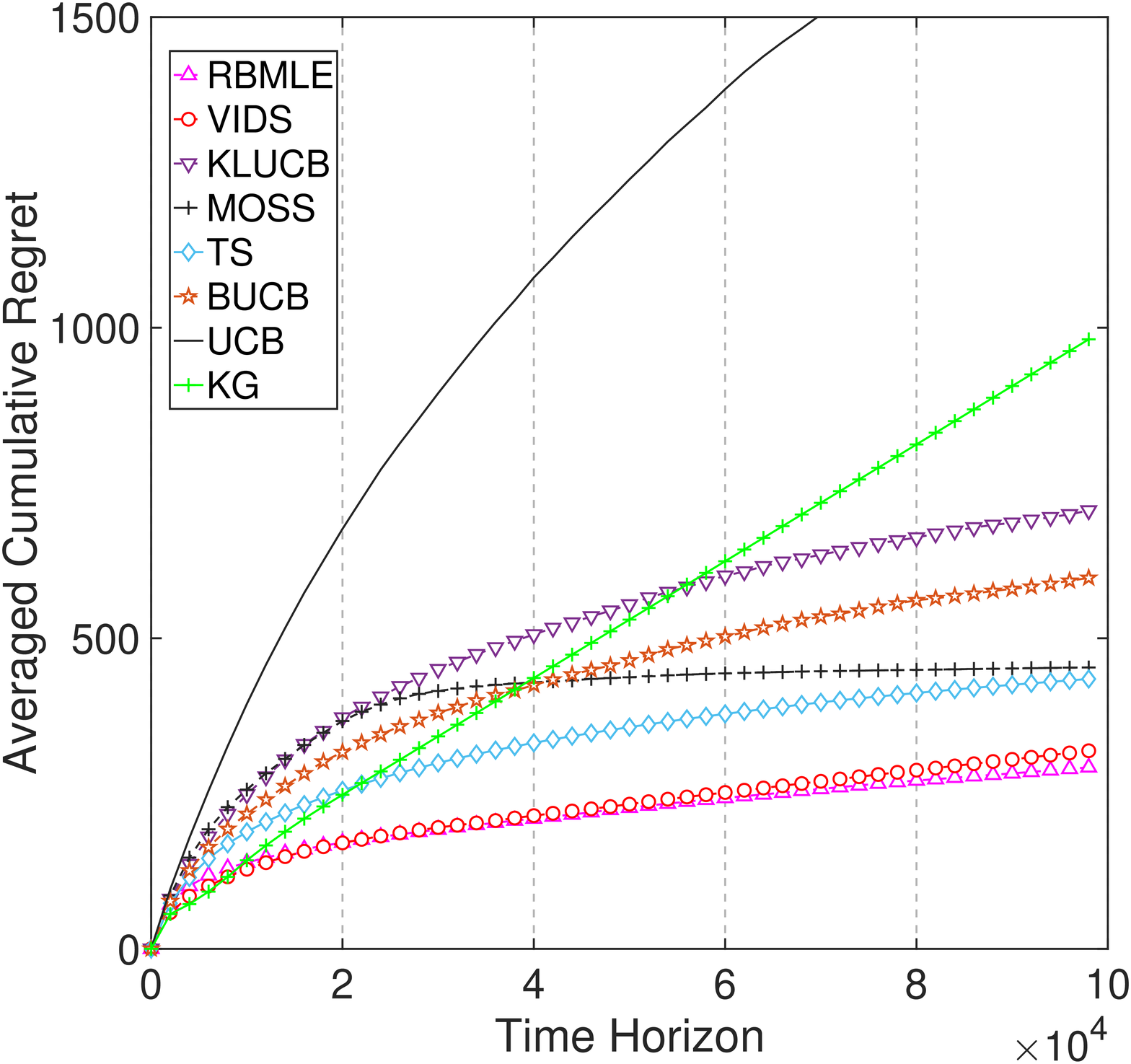}} \label{fig:Exp_ID=14_Regret_T=1e5}\\ [0.0cm]
    \mbox{\small (a)} & \hspace{-2mm} \mbox{\small (b)} & \hspace{-2mm} \mbox{\small (c)} \\[-0.2cm]
\end{array}$
\caption{Averaged cumulative regret: (a) Bernoulli bandits with $(\theta_i)_{i=1}^{10}=$ (0.655, 0.6, 0.665, 0.67, 0.675, 0.68, 0.685, 0.69, 0.695, 0.7) \& $\Delta=0.005$; (b) Gaussian bandits with $(\theta_i)_{i=1}^{10}=$ (0.5, 0.75, 0.4, 0.6, 0.55, 0.76, 0.68, 0.41, 0.52, 0.67) \& $\Delta=0.01$; (c) Exponential bandits with $(\theta_i)_{i=1}^{10}=$ (0.46, 0.45, 0.5, 0.48, 0.51, 0.4, 0.43, 0.42, 0.45, 0.44) \& $\Delta=0.01$.}
\label{fig:regret all 2}
\end{figure*}

\begin{figure*}[h]
\centering
$\begin{array}{c c c}
    \multicolumn{1}{l}{\mbox{\bf }} & \multicolumn{1}{l}{\mbox{\bf }} & \multicolumn{1}{l}{\mbox{\bf }} \\ 
    \hspace{-1mm} \scalebox{0.33}{\includegraphics[width=\textwidth]{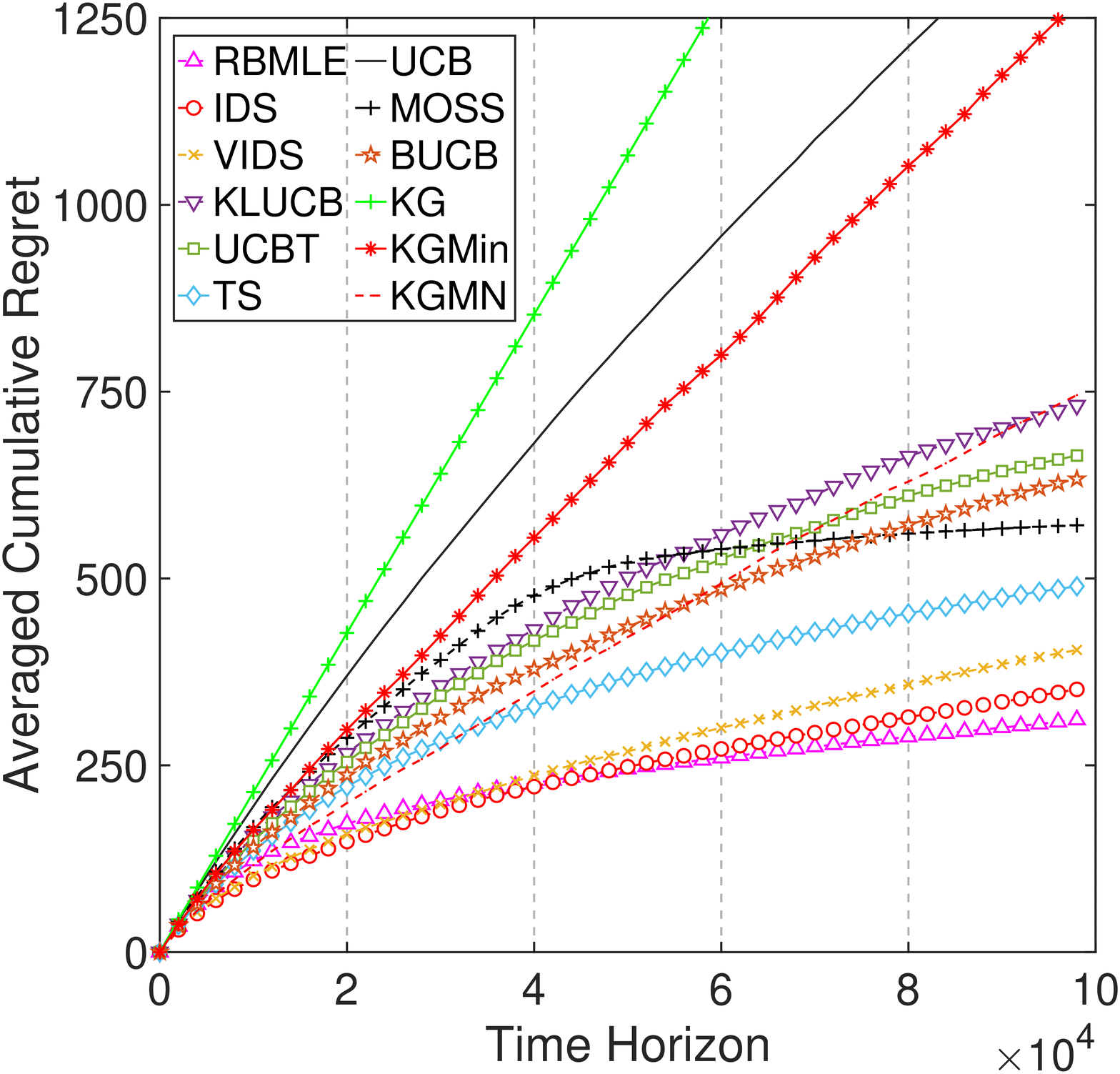}}  \label{fig:Bernoulli_ID=13_Regret_T=1e5} & 
    \hspace{-3mm} \scalebox{0.33}{\includegraphics[width=\textwidth]{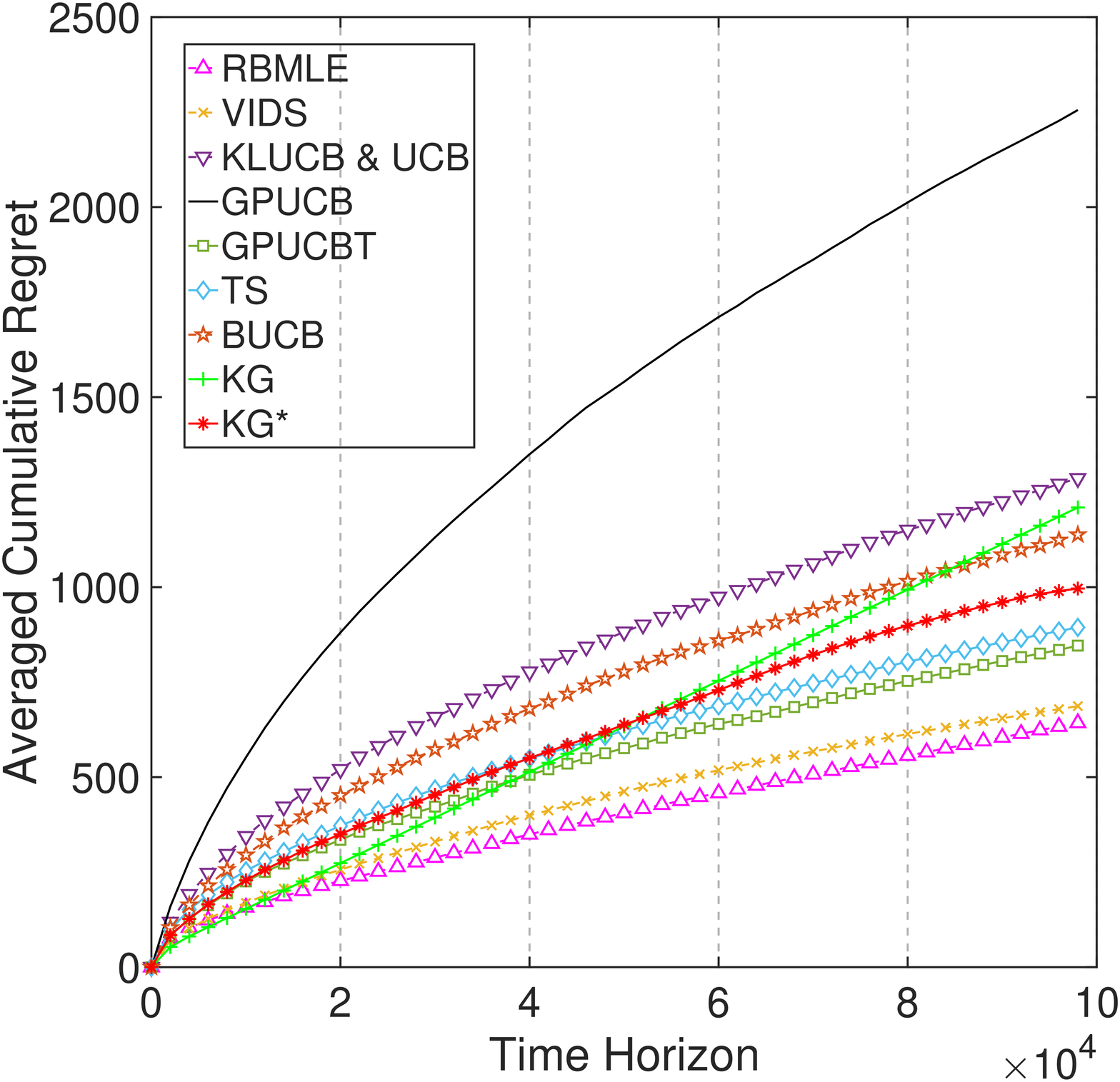}} \label{fig:Gaussian_ID=24_Regret_T=1e5} & \hspace{-3mm}
    \scalebox{0.33}{\includegraphics[width=\textwidth]{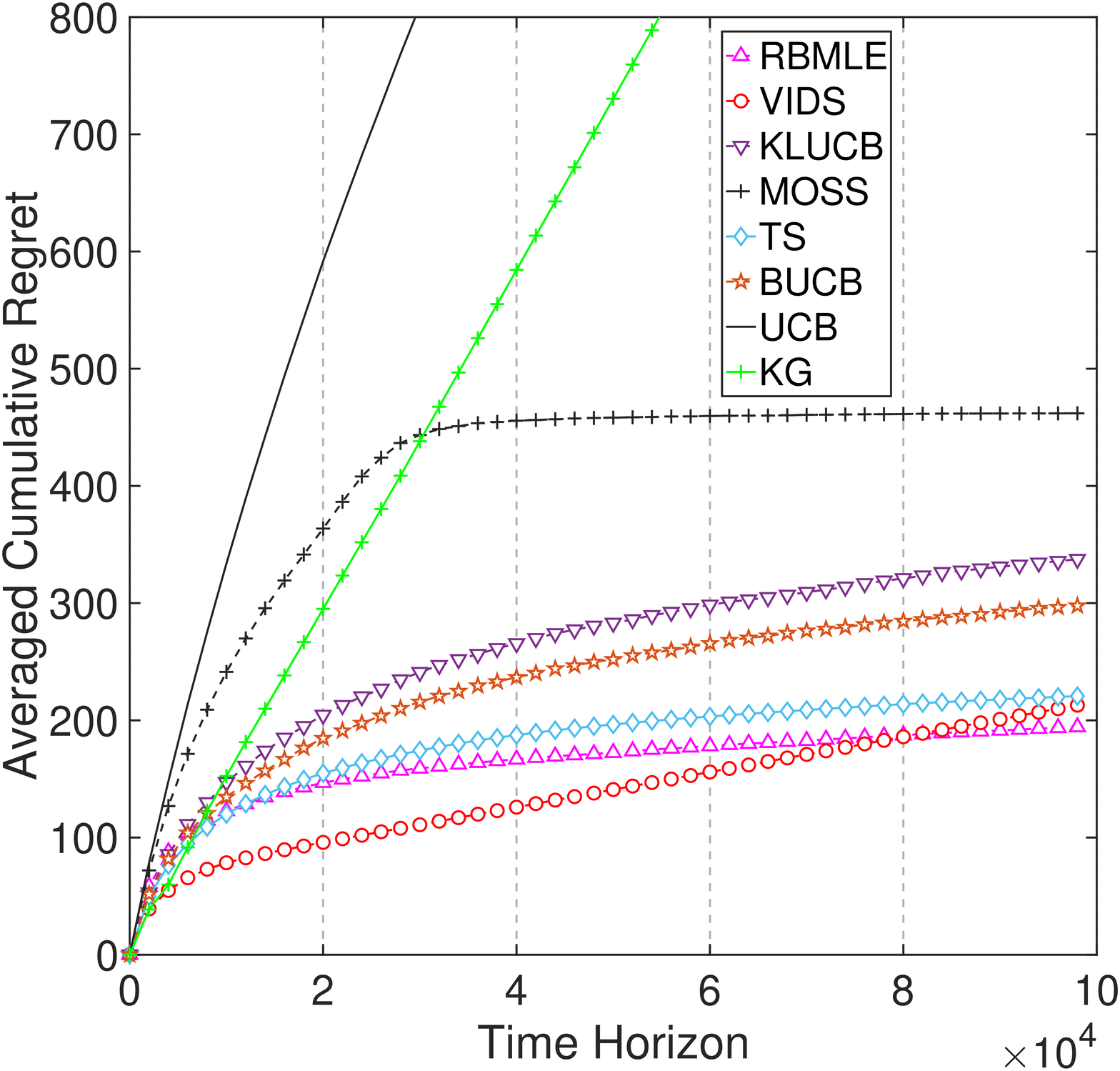}} \label{fig:Exp_ID=17_Regret_T=1e5}\\ [0.0cm]
    \mbox{\small (a)} & \hspace{-2mm} \mbox{\small (b)} & \hspace{-2mm} \mbox{\small (c)} \\[-0.2cm]
\end{array}$
\caption{Averaged cumulative regret: (a) Bernoulli bandits with $(\theta_i)_{i=1}^{10}=$ (0.755, 0.76, 0.765, 0.77, 0.775, 0.78, 0.785, 0.79, 0.795, 0.8) \& $\Delta=0.005$; (b) Gaussian bandits with $(\theta_i)_{i=1}^{10}=$ (0.65, 0.35, 0.66, 0.4, 0.65, 0.64, 0.55, 0.4, 0.57, 0.54) \& $\Delta=0.01$; (c) Exponential bandits with $(\theta_i)_{i=1}^{10}=$ (0.25, 0.28, 0.27, 0.3, 0.29, 0.22, 0.21, 0.24, 0.23, 0.26) \& $\Delta=0.01$.}
\label{fig:regret all 3}
\end{figure*}

Figures~\ref{fig:regret all 2}-\ref{fig:regret all 3} illustrate the effectiveness of RBMLE with respect to the cumulative regret,  under a different set of parameters, for the three types of bandits. Tables~\ref{table:Bernoulli_ID=2}-\ref{table:Exponential_ID=17} provide detailed statistics, including the mean as well as the standard deviation and quantiles of the final regrets, with the row-wise smallest values highlighted in boldface. From the Tables, we observe that RBMLE tends to have the smallest value of regret at medium to high quantiles, and comparable to the smallest values at other lower quantiles among those that have comparable mean values (e.g., IDS, VIDS, KLUCB). Along with the presented statistics of standard deviation, they suggest that RBMLE's performance enjoys comparable robustness as those baselines that achieve similar mean regret.

\subsection{Efficiency}
\label{appendix:more empirical results on efficiency}
\begin{figure*}[h]
\centering
$\begin{array}{c c c}
    \multicolumn{1}{l}{\mbox{\bf }} & \multicolumn{1}{l}{\mbox{\bf }} & \multicolumn{1}{l}{\mbox{\bf }} \\ 
    \hspace{-1mm} \scalebox{0.33}{\includegraphics[width=\textwidth]{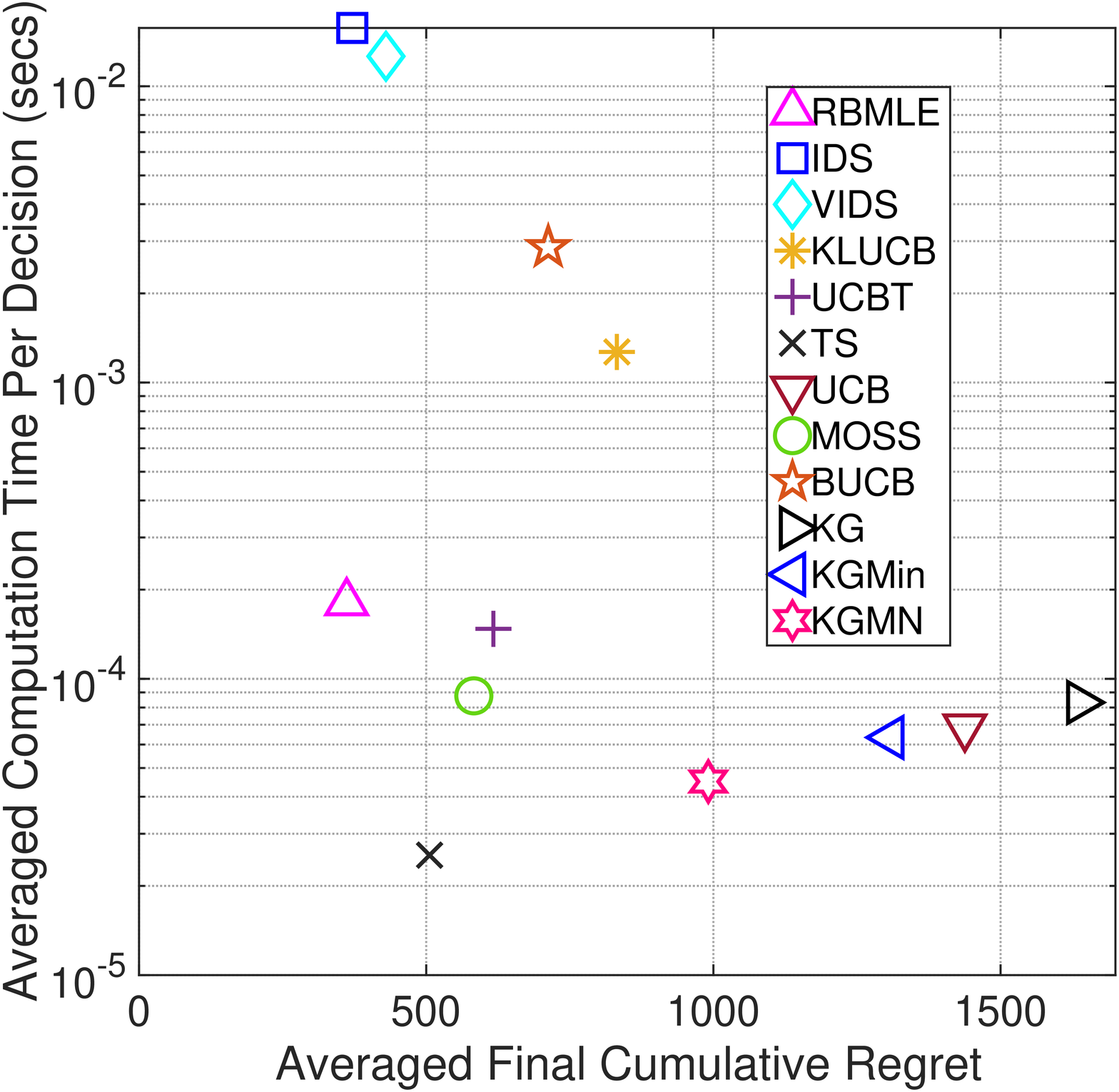}}  \label{fig:Bernoulli_ID=3_Scatter_T=1e5} & 
    \hspace{-3mm} \scalebox{0.33}{\includegraphics[width=\textwidth]{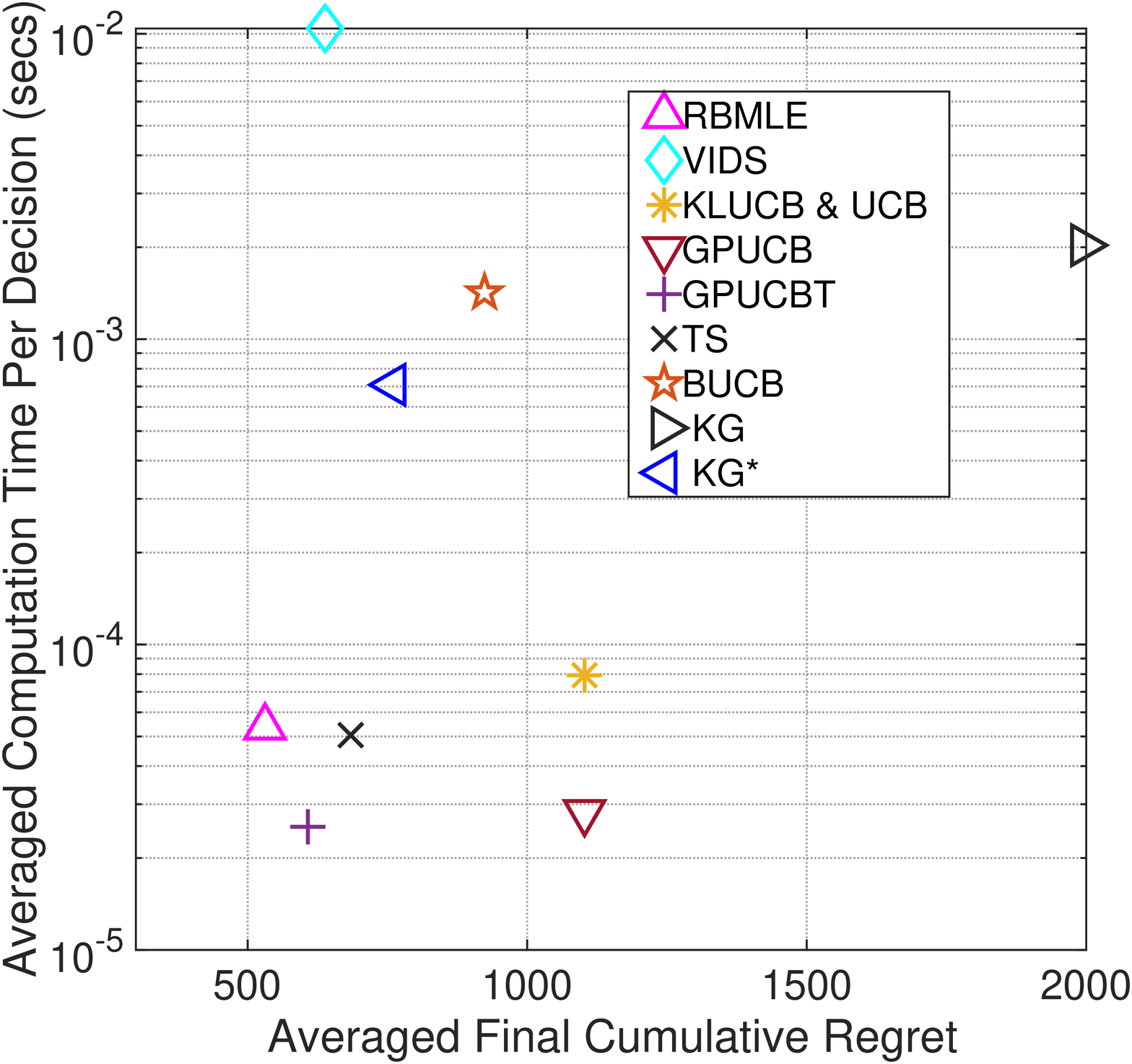}} \label{fig:Gaussian_ID=27_Scatter_T=1e5} & \hspace{-3mm}
    \scalebox{0.33}{\includegraphics[width=\textwidth]{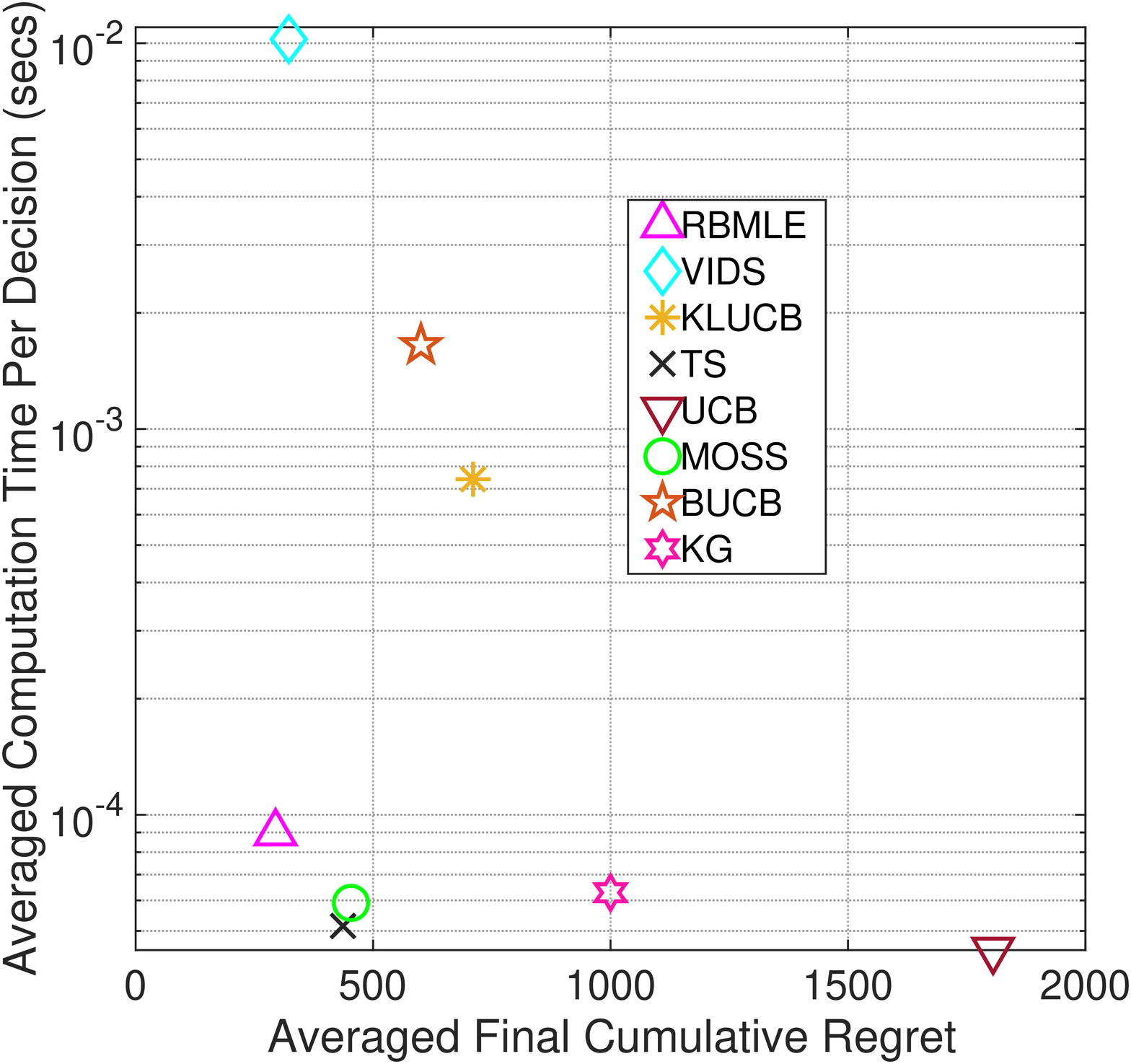}} \label{fig:Exp_ID=14_Scatter_T=1e5}\\ [0.0cm]
    \mbox{\small (a)} & \hspace{-2mm} \mbox{\small (b)} & \hspace{-2mm} \mbox{\small (c)} \\[-0.2cm]
\end{array}$
\caption{Averaged computation time per decision vs. averaged final cumulative regret: (a) Figure \ref{fig:regret all 2}(a); (b) Figure \ref{fig:regret all 2}(b);  (a) Figure \ref{fig:regret all 2}(c).}
\label{fig:time all 2}
\end{figure*}

\begin{figure*}[h]
\centering
$\begin{array}{c c c}
    \multicolumn{1}{l}{\mbox{\bf }} & \multicolumn{1}{l}{\mbox{\bf }} & \multicolumn{1}{l}{\mbox{\bf }} \\ 
    \hspace{-1mm} \scalebox{0.33}{\includegraphics[width=\textwidth]{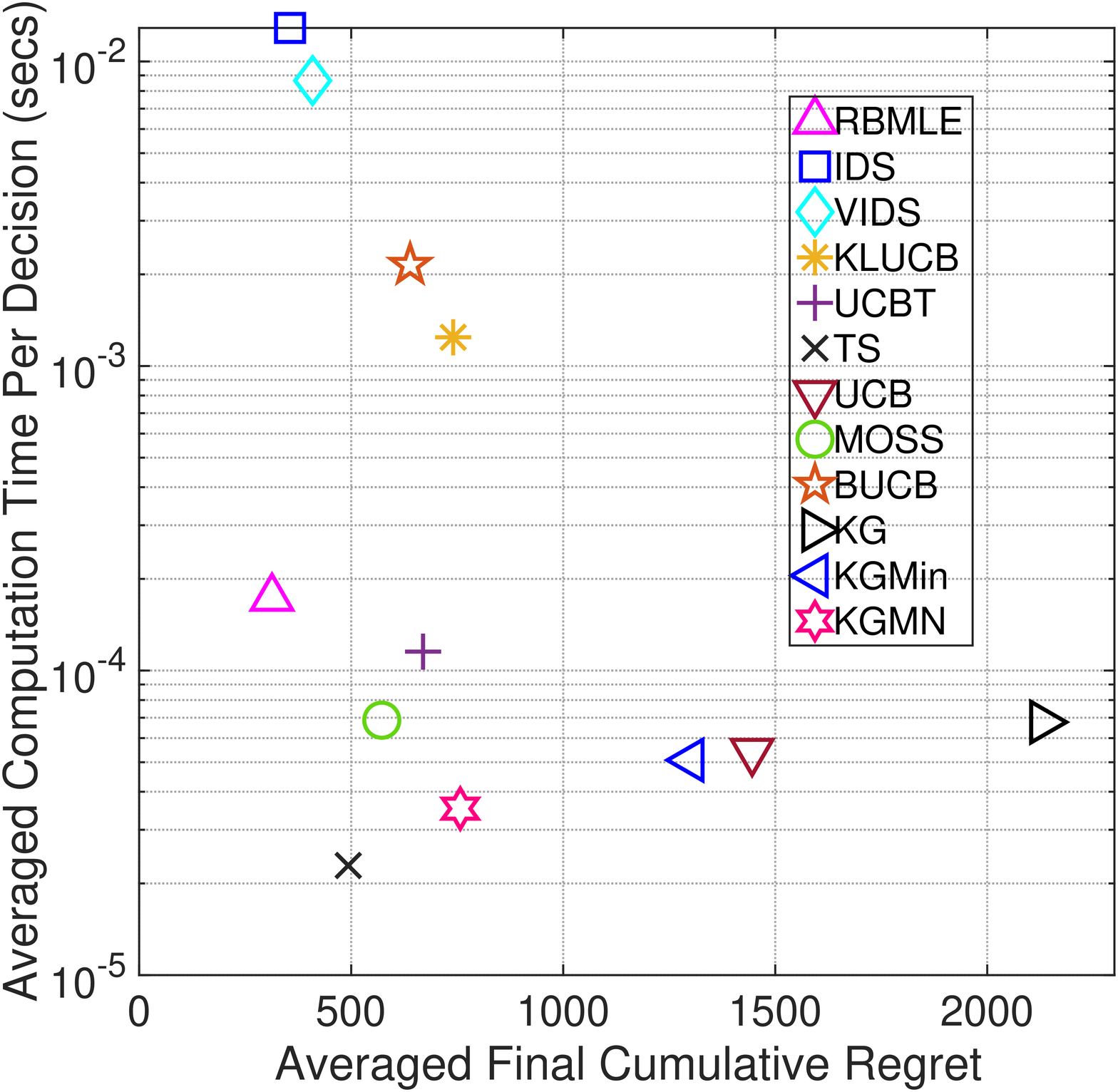}}  \label{fig:Bernoulli_ID=13_Scatter_T=1e5} & 
    \hspace{-3mm} \scalebox{0.33}{\includegraphics[width=\textwidth]{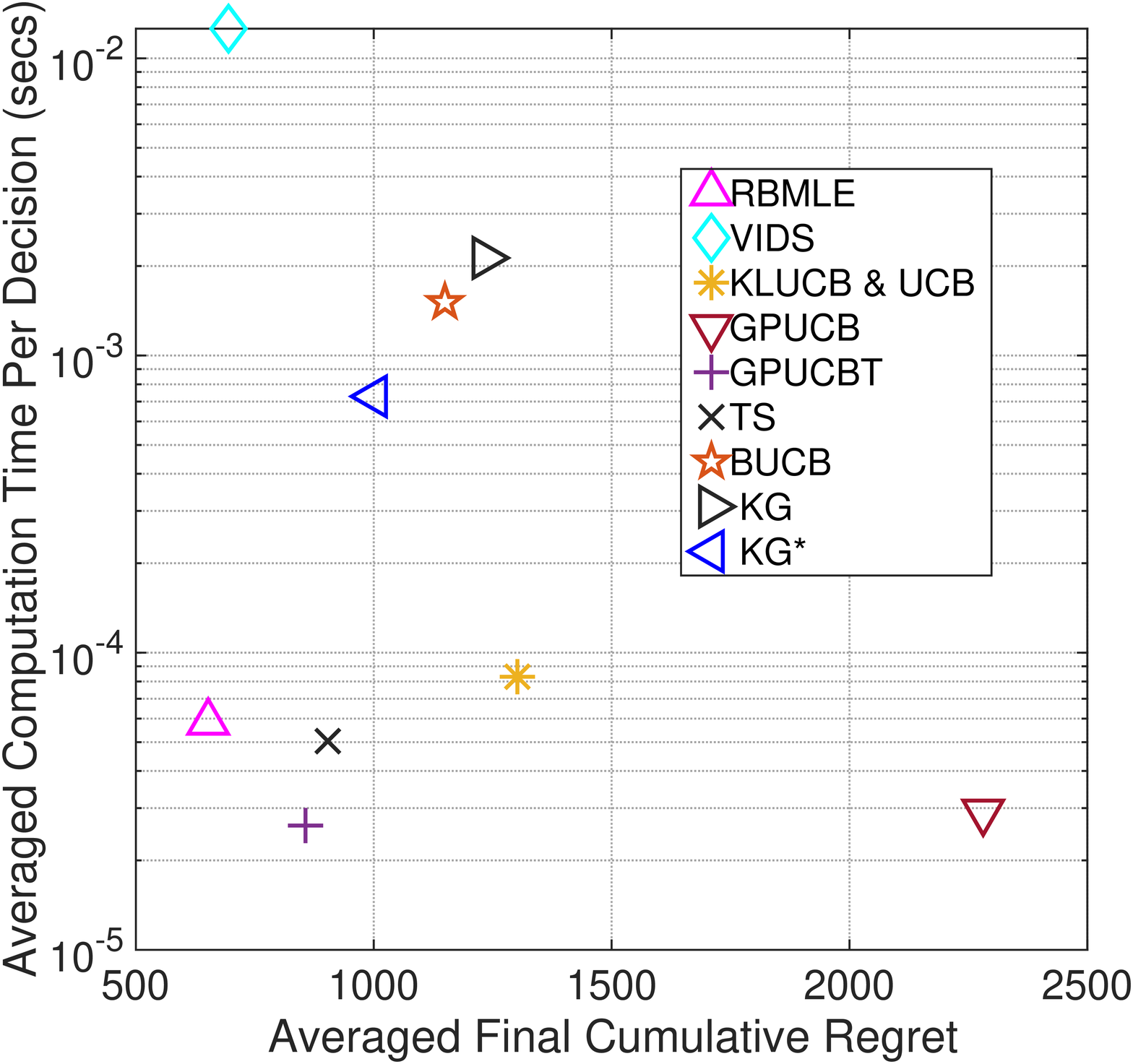}} \label{fig:Gaussian_ID=24_Scatter_T=1e5} & \hspace{-3mm}
    \scalebox{0.33}{\includegraphics[width=\textwidth]{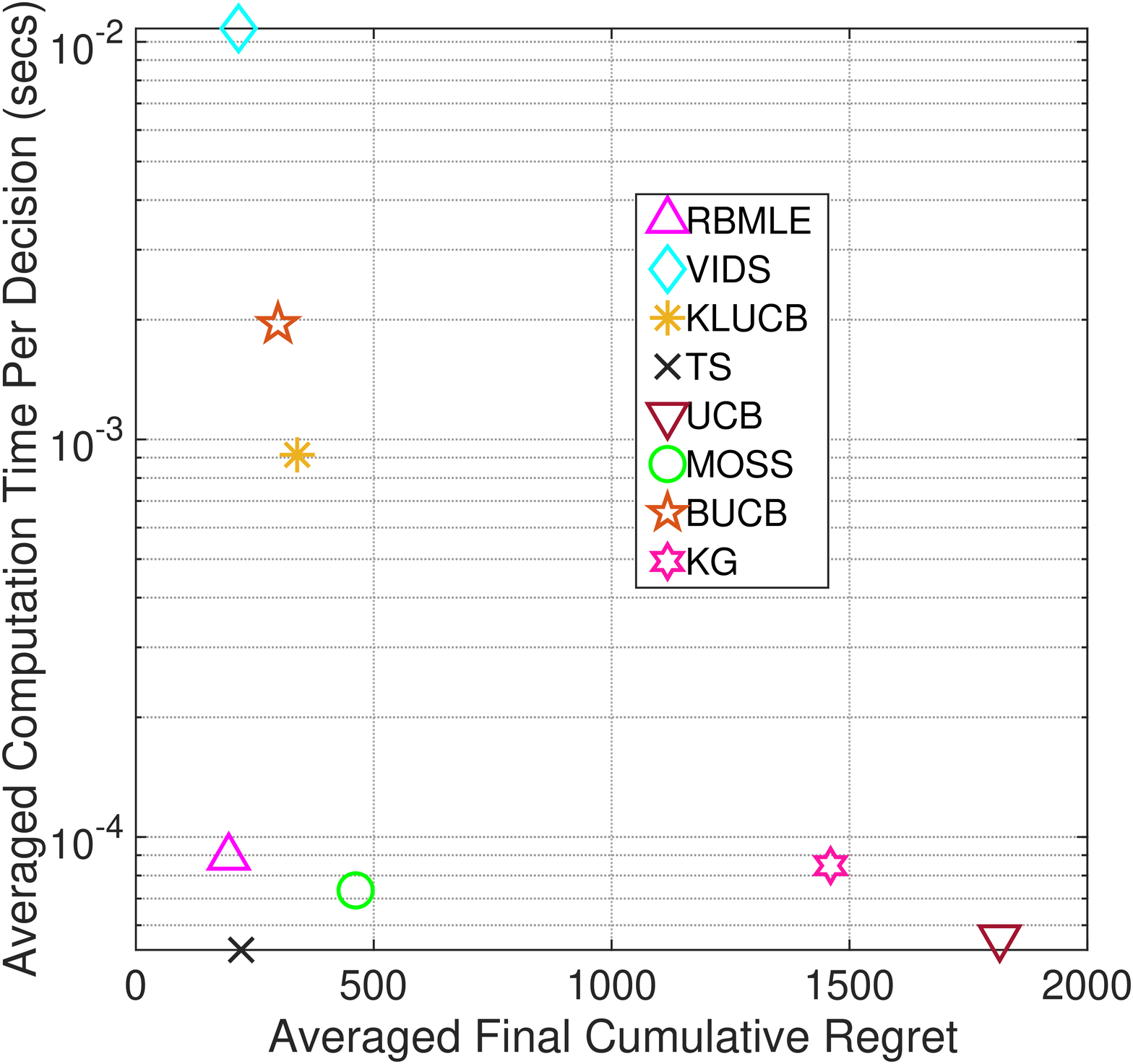}} \label{fig:Exp_ID=17_Scatter_T=1e5}\\ [0.0cm]
    \mbox{\small (a)} & \hspace{-2mm} \mbox{\small (b)} & \hspace{-2mm} \mbox{\small (c)} \\[-0.2cm]
\end{array}$
\caption{Averaged computation time per decision vs. averaged final cumulative regret: (a) Figure \ref{fig:regret all 3}(a); (b) Figure \ref{fig:regret all 3}(b);  (a) Figure \ref{fig:regret all 3}(c).}
\label{fig:time all 3}
\end{figure*}

Figures \ref{fig:time all 2}-\ref{fig:time all 3} present the efficiency of RBMLE in terms of averaged computation time per decision (ACTPD) vs. averaged final cumulative regret. The computation times are measured on a Linux server with (i) an Intel Xeon E7 v4 server operating at a maximal clock rate of 3.60~GHz, and (ii) a total of 528~GB memory. While there are 64 cores in the server, we force the program to run on just one core for a fair comparison. 

\subsection{Scalability}
\label{appendix:more empirical results on scalability}
Tables~\ref{table:Bernoulli_computation_time}-\ref{table:Exponential_computation_time} show the computation time per decision of different policies under varying numbers of arms.

\begin{table}[h]
\caption{Statistics of the final cumulative regret in Figure \ref{fig:regret all}(a). The best in each row is highlighted.}
\label{table:Bernoulli_ID=2}
\begin{center}
{\small
\begin{tabular}{|p{2.3cm}|p{1cm}|p{0.8cm}|p{0.9cm}|p{1.0cm}|p{0.7cm}|p{0.5cm}|p{0.7cm}|p{0.7cm}|p{0.7cm}|p{0.7cm}|p{0.9cm}|p{0.8cm}|}
\hline
\textbf{Algorithm} & \textbf{RBMLE} & \textbf{IDS} & \textbf{VIDS} & \textbf{KLUCB} & \textbf{UCBT} & \textbf{TS} & \textbf{UCB} & \textbf{MOSS} & \textbf{BUCB} & \textbf{KG} & \textbf{KGMin} & \textbf{KGMN}     
\\ \hline
{\tt Mean Regret} & \textbf{\underline{263.5}} & 406.3 & 449.6 & 730.4 & 474.7 & 426.9 & 1809.5 & 464.5 & 580.9 & 2379.5 & 2384.2 & 1814.3
\\ \hline
{\tt Std. Dev.} & 233.5 & 466.7 & 618.2 &	109.3 &	176.3 &	149.3 &	113.0 &	\textbf{\underline{93.1}} &	105.8 &	2163.2 & 355.4 & 344.0
\\ \hline
{\tt Quantile .10} & 142.4 & 74.7 &	54.2 &	584.4 &	309.9 & 283.2	& 1647.3 & 355.8 & 452.9 & \textbf{\underline{3.7}} & 1899.1 &	1344.8
\\  \hline
{\tt Quantile .25} & 161.4 & 113.5 & \textbf{\underline{90.0}} & 661.0 & 362.8 &	326.6 &	1750.4 & 394.2 & 519.2 & 1000.9 &	2103.2 & 1555.3
\\ \hline
{\tt Quantile .50} & 190.6 & 184.3 & \textbf{\underline{134.5}} & 717.7 &	448.7 &	404.0 &	1815.5 & 458.9 & 563.77 & 2001.4 &	2411.6 & 1844.2
\\ \hline
{\tt Quantile .75} & \textbf{\underline{237.8}} & 461.4	& 1043.2	& 804.0	& 543.5 &	489.2 &	1874.3 & 520.8 & 638.0	& 3999.8 &	2620.3 & 2057.7
\\ \hline
{\tt Quantile .90} & \textbf{\underline{430.0}} & 1138.3 & 1116.4 & 860.9 &	655.0 &	595.1 &	1963.1 & 570.8 & 713.8 & 5013.8 & 2809.4	& 2254.1
\\ \hline
{\tt Quantile .95} & 993.1 & 1247.9	& 1247.9 &	926.3	& 759.7 &	647.6 &	1996.6 & \textbf{\underline{615.6}} & 766.0 & 6992.2 & 2911.4 & 2326.5
\\ \hline
\end{tabular}
}
\end{center}
\end{table}

\begin{table}[h]
\caption{Statistics of the final cumulative regret in Figure \ref{fig:regret all 2}(a). The best in each row is highlighted. }
\label{table:Bernoulli_ID=3}
\begin{center}
{\small
\begin{tabular}{|p{2.4cm}|p{1cm}|p{0.8cm}|p{0.9cm}|p{1.0cm}|p{0.7cm}|p{0.5cm}|p{0.7cm}|p{0.7cm}|p{0.7cm}|p{0.7cm}|p{0.9cm}|p{0.8cm}|}
\hline
\textbf{Algorithm} & \textbf{RBMLE} & \textbf{IDS} & \textbf{VIDS} & \textbf{KLUCB} & \textbf{UCBT} & \textbf{TS} & \textbf{UCB} & \textbf{MOSS} & \textbf{BUCB} & \textbf{KG} & \textbf{KGMin} & \textbf{KGMN}     
\\ \hline
{\tt Mean Regret} & \textbf{\underline{361.5}} & 371.3 & 416.7 & 831.9 & 616.9 & 505.8 & 1437.9 & 582.9 & 712.5 & 1637.9 & 1309.0 & 991.1
\\ \hline
{\tt Std. Dev.} & 247.6 & 285.9 & 342.8 &	131.9 &	130.7 &	156.3 &	\textbf{\underline{78.5}} &	169.9 &	120.3 &	1592.7 & 214.1 & 198.4
\\ \hline
{\tt Quantile .10} & 133.0 & 116.3 & 77.2 & 650.3 & 440.6  & 334.4	& 1335.8 & 411.6 & 564.3 & \textbf{\underline{2.3}} & 1013.3 &	741.9
\\  \hline
{\tt Quantile .25} & 165.1 & 164.4 & \textbf{\underline{147.0}} & 732.4 & 532.3 & 385.0 & 1388.3 & 461.1 & 641.2 & 501.5 & 1184.1 & 876.2
\\ \hline
{\tt Quantile .50} & \textbf{\underline{223.5}} & 262.8 & 248.3 & 823.1 & 598.0 & 477.9 & 1436.7 & 532.7 & 715.1 & 1002.6 &	1338.8 & 987.8
\\ \hline
{\tt Quantile .75} & 608.4 & \textbf{\underline{568.9}}	& 593.1	& 930.3	& 693.4 & 575.3 & 1495.6 & 654.8 & 782.6 & 2996.5 &	1447.2 & 1119.8
\\ \hline
{\tt Quantile .90} & \textbf{\underline{661.2}} & 681.7 & 1003.2 & 1020.6 & 779.5 &	698.3 &	1540.4 & 816.1 & 865.3 & 3499.3 & 1538.3 & 1228.4
\\ \hline
{\tt Quantile .95} & \textbf{\underline{722.9}} & 835.1	& 1060.9 & 1062.0 & 857.9 & 793.4 & 1561.2 & 943.2 & 906.4 & 4497.6 & 1620.9 & 1343.0
\\ \hline
\end{tabular}
}
\end{center}
\end{table}

\begin{table}[h]
\caption{Statistics of the final cumulative regret in Figure \ref{fig:regret all 3}(a). The best in each row is highlighted.}
\label{table:Bernoulli_ID=13}
\begin{center}
{\small
\begin{tabular}{|p{2.3cm}|p{1cm}|p{0.8cm}|p{0.9cm}|p{1.0cm}|p{0.7cm}|p{0.5cm}|p{0.7cm}|p{0.7cm}|p{0.7cm}|p{0.7cm}|p{0.9cm}|p{0.8cm}|}
\hline
\textbf{Algorithm} & \textbf{RBMLE} & \textbf{IDS} & \textbf{VIDS} & \textbf{KLUCB} & \textbf{UCBT} & \textbf{TS} & \textbf{UCB} & \textbf{MOSS} & \textbf{BUCB} & \textbf{KG} & \textbf{KGMin} & \textbf{KGMN}     
\\ \hline
{\tt Mean Regret} & \textbf{\underline{313.2}} & 355.7 & 425.5 & 740.5 & 669.5 & 493.0 & 1445.6 & 572.1 & 638.9 & 2131.2 & 1301.0 & 757.2
\\ \hline
{\tt Std. Dev.} & 228.1 & 386.5 & 474.7 &	126.9 &	120.3 &	171.9 &	\textbf{\underline{69.1}} &	132.7 &	127.8 &	1336.5 & 209.8 & 178.3
\\ \hline
{\tt Quantile .10} & 142.2 & 95.1 & \textbf{\underline{70.2}} & 581.4 & 506.7  & 328.9	& 1347.8 & 433.6 & 485.6 & 451.3 & 1024.6 &	531.2
\\  \hline
{\tt Quantile .25} & 169.4 & 133.2 & \textbf{\underline{102.5}} & 651.6 & 573.5 & 374.0 & 1396.6 & 463.5 & 542.7 & 1001.1 & 1174.2 & 628.7
\\ \hline
{\tt Quantile .50} & 203.7 & 179.0 & \textbf{\underline{173.9}} & 725.9 & 680.5 & 463.3 & 1446.4 & 543.0 & 623.3 & 2001.2 &	1322.2 & 754.0
\\ \hline
{\tt Quantile .75} & \textbf{\underline{369.7}} & 543.2	& 589.7	& 806.5	& 751.9 & 534.0 & 1497.7 & 648.6 & 724.3 & 3000.2 &	1442.1 & 897.2
\\ \hline
{\tt Quantile .90} & \textbf{\underline{680.2}} & 695.5 & 1067.4 & 886.3 & 833.6 & 726.2 & 1541.1 & 760.8 & 813.5 & 3999.8 & 1563.4 & 963.8
\\ \hline
{\tt Quantile .95} & \textbf{\underline{720.2}} & 891.4	& 1739.8 & 999.6 & 867.1 & 774.1 & 1554.6 & 830.9 & 867.4 & 4498.3 & 1590.7 & 1039.2
\\ \hline
\end{tabular}
}
\end{center}
\end{table}

\begin{table}[h]
\caption{Statistics of the final cumulative regret in Figure \ref{fig:regret all}(b). The best in each row is highlighted.}
\label{table:Gaussian_ID=16}
\begin{center}
{\small
\begin{tabular}{|p{2.3cm}|p{1cm}|p{0.8cm}|p{2.0cm}|p{1.0cm}|p{1.1cm}|p{0.7cm}|p{0.9cm}|p{0.8cm}|p{0.9cm}|}
\hline
\textbf{Algorithm} & \textbf{RBMLE} & \textbf{VIDS} & \textbf{KLUCB\&UCB} & \textbf{GPUCB} & \textbf{GPUCBT} & \textbf{TS} & \textbf{BUCB} & \textbf{KG} & \textbf{KG*}  
\\ \hline
{\tt Mean Regret} & \textbf{\underline{730.6}} & 775.0 & 1412.2 & 2640.3 & 848.5 & 932.7 & 1222.3 & 1684.3 & 1046.0
\\ \hline
{\tt Std. Dev.} & 827.4 & 678.7 & \textbf{\underline{219.2}} & 227.0 & 314.2 &	282.1 &	231.4 &	2056.8 & 238.9
\\ \hline
{\tt Quantile .10} & 135.3 & 233.9 & 1147.2 & 2382.8 & 529.3  & 657.8 & 960.8 & \textbf{\underline{20.4}} & 788.0
\\  \hline
{\tt Quantile .25} & 160.2 & 336.0 & 1272.1 & 2500.0 & 608.0 & 706.6 & 1036.5 & \textbf{\underline{59.9}} & 891.6
\\ \hline
{\tt Quantile .50} & \textbf{\underline{263.1}} & 544.1 & 1395.9 & 2600.4 & 814.7 & 876.0 & 1205.9 & 1035.8 & 1000.6
\\ \hline
{\tt Quantile .75} & 1140.8 & 1137.7 & 1545.9 & 2787.1	& \textbf{\underline{1001.1}} & 1125.3 & 1390.6 & 2028.0 & 1171.1
\\ \hline
{\tt Quantile .90} & 2107.9 & 1516.5 & 1674.6 & 2916.1 & \textbf{\underline{1228.6}} & 1304.8 & 1512.9 & 4028.8 & 1314.1
\\ \hline
{\tt Quantile .95} & 2157.6 & 1862.0 & 1724.6 & 3024.4 & 1578.7 & 1472.7 & 1565.5 & 7818.3 & \textbf{\underline{1413.7}}
\\ \hline
\end{tabular}
}
\end{center}
\end{table}

\begin{table}[h]
\caption{Statistics of the final cumulative regret in Figure \ref{fig:regret all 2}(b). The best in each row is highlighted.}
\label{table:Gaussian_ID=27}
\begin{center}
{\small
\begin{tabular}{|p{2.3cm}|p{1cm}|p{0.8cm}|p{2.0cm}|p{1.0cm}|p{1.1cm}|p{0.7cm}|p{0.9cm}|p{0.8cm}|p{0.9cm}|}
\hline
\textbf{Algorithm} & \textbf{RBMLE} & \textbf{VIDS} & \textbf{KLUCB\&UCB} & \textbf{GPUCB} & \textbf{GPUCBT} & \textbf{TS} & \textbf{BUCB} & \textbf{KG} & \textbf{KG*}  
\\ \hline
{\tt Mean Regret} & \textbf{\underline{531.1}} & 638.5 & 1102.7 & 2464.2 & 607.7 & 684.3 & 923.6 & 1995.0 & 760.2
\\ \hline
{\tt Std. Dev.} & 469.5 & 1117.0 & 196.9 & 210.8 &	234.1 &	250.1 &	178.7 & 3541.8 & \textbf{\underline{163.8}}
\\ \hline
{\tt Quantile .10} & 145.5 & 143.7 & 859.7 & 2200.1 & 361.4 & 411.1 & 724.5 & \textbf{\underline{21.1}} & 568.4
\\  \hline
{\tt Quantile .25} & 167.4 & 206.6 & 937.4 & 2320.7 & 444.3 & 501.7 & 792.9 & \textbf{\underline{30.2}} & 664.5
\\ \hline
{\tt Quantile .50} & \textbf{\underline{207.7}} & 314.1 & 1093.2 & 2466.4 & 544.8 & 623.1 & 927.2 & 1014.4 & 752.5
\\ \hline
{\tt Quantile .75} & 1131.8 & 889.0 & 1232.0 & 2605.0 & \textbf{\underline{714.6}} & 792.2 & 1042.0 & 1044.3 & 851.4
\\ \hline
{\tt Quantile .90} & 1188.1 & 1183.3 & 1346.8 & 2726.0 & \textbf{\underline{926.2}} & 1058.9 & 1174.1 & 8121.5 & 930.0
\\ \hline
{\tt Quantile .95} & 1204.2 & 1248.6 & 1439.0 & 2804.9 & 1041.8 & 1209.2 & 1193.5 & 9023.5 & \textbf{\underline{959.5}}
\\ \hline
\end{tabular}
}
\end{center}
\end{table}

\begin{table}[h]
\caption{Statistics of the final cumulative regret in Figure \ref{fig:regret all 3}(b). The best in each row is highlighted.}
\label{table:Gaussian_ID=24}
\begin{center}
{\small
\begin{tabular}{|p{2.3cm}|p{1cm}|p{0.8cm}|p{2.0cm}|p{1.0cm}|p{1.1cm}|p{0.7cm}|p{0.9cm}|p{0.8cm}|p{0.9cm}|}
\hline
\textbf{Algorithm} & \textbf{RBMLE} & \textbf{VIDS} & \textbf{KLUCB\&UCB} & \textbf{GPUCB} & \textbf{GPUCBT} & \textbf{TS} & \textbf{BUCB} & \textbf{KG} & \textbf{KG*}  
\\ \hline
{\tt Mean Regret} & \textbf{\underline{652.0}} & 694.7 & 1302.0 & 2281.0 & 856.5 & 903.4 & 1149.5 & 1233.6 & 1001.7
\\ \hline
{\tt Std. Dev.} & 581.8 & 776.1 & \textbf{\underline{164.5}} & 169.5 & 255.8 &	268.2 &	201.0 &	1659.2 & 234.8
\\ \hline
{\tt Quantile .10} & 127.3 & 193.6 & 1100.0 & 2062.5 & 561.1 & 574.8 & 897.0 & \textbf{\underline{24.5}} & 747.2
\\  \hline
{\tt Quantile .25} & 155.7 & 322.9 & 1173.4 & 2156.6 & 665.7 & 715.8 & 1000.4 & \textbf{\underline{72.0}} & 827.9
\\ \hline
{\tt Quantile .50} & \textbf{\underline{265.4}} & 471.9 & 1295.7 & 2262.7 & 814.3 & 849.3 & 1130.5 & 1021.1 & 944.4
\\ \hline
{\tt Quantile .75} & 1116.2 & \textbf{\underline{861.0}} & 1428.3 & 2397.7 & 1007.8 & 1085.6 & 1294.0 & 1987.0 & 1128.1
\\ \hline
{\tt Quantile .90} & 1202.8 & 1236.1 & 1492.8 & 2506.3 & \textbf{\underline{1164.6}} & 1283.0 & 1404.6 & 2028.1 & 1346.7
\\ \hline
{\tt Quantile .95} & 2021.8 & 1467.5 & 1549.4 & 2545.1 & \textbf{\underline{1334.9}} & 1394.5 & 1511.5 & 2055.5 & 1467.2
\\ \hline
\end{tabular}
}
\end{center}
\end{table}

\begin{table}[h]
\caption{Statistics of the final cumulative regret in Figure \ref{fig:regret all}(c). The best in each row is highlighted.}
\label{table:Exponential_ID=21}
\begin{center}
{\small
\begin{tabular}{|p{2.3cm}|p{1cm}|p{0.8cm}|p{1.0cm}|p{0.7cm}|p{1.1cm}|p{0.8cm}|p{0.9cm}|p{0.8cm}|}
\hline
\textbf{Algorithm} & \textbf{RBMLE} & \textbf{VIDS} & \textbf{KLUCB} & \textbf{TS} & \textbf{UCB} & \textbf{MOSS} & \textbf{BUCB} & \textbf{KG}  
\\ \hline
{\tt Mean Regret} & \textbf{\underline{179.6}} & 243.3 & 322.7 & 208.6 & 1504.6 & 379.9 & 288.2 & 961.6
\\ \hline
{\tt Std. Dev.} & 119.4 & 463.1 & 63.9 & 61.3 & 66.1 & \textbf{\underline{44.5}} & 71.9 & 1063.3
\\ \hline
{\tt Quantile .10} & 128.7 & 37.6 & 239.4 & 132.8 & 1430.9 & 329.4 & 196.7 & \textbf{\underline{26.5}}
\\  \hline
{\tt Quantile .25} & 139.7 & 47.9 & 271.3 & 157.7 & 1452.0 & 345.8 & 238.3 & \textbf{\underline{37.2}}
\\  \hline
{\tt Quantile .50} & 155.2 & \textbf{\underline{70.5}} & 331.7 & 202.3 & 1505.4 & 380.1 & 275.1 & 387.2
\\ \hline
{\tt Quantile .75} & 173.4 & \textbf{\underline{103.7}} & 367.2 & 243.4 & 1550.6 & 405.9 & 330.6 & 2450.7
\\ \hline
{\tt Quantile .90} & \textbf{\underline{195.4}} & 1039.9 & 407.0 & 303.1 & 1586.5 & 435.0 & 377.3 & 2509.9
\\ \hline
{\tt Quantile .95} & \textbf{\underline{291.7}} & 1074.1 & 423.2 & 320.1 & 1617.6 & 457.8 & 405.3 & 2522.7
\\ \hline
\end{tabular}
}
\end{center}
\end{table}

\begin{table}[h]
\caption{Statistics of the final cumulative regret in Figure \ref{fig:regret all 2}(c). The best in each row is highlighted. }
\label{table:Exponential_ID=14}
\begin{center}
{\small
\begin{tabular}{|p{2.3cm}|p{1cm}|p{0.8cm}|p{1.0cm}|p{0.7cm}|p{1.1cm}|p{0.8cm}|p{0.9cm}|p{0.8cm}|}
\hline
\textbf{Algorithm} & \textbf{RBMLE} & \textbf{VIDS} & \textbf{KLUCB} & \textbf{TS} & \textbf{UCB} & \textbf{MOSS} & \textbf{BUCB} & \textbf{KG}  
\\ \hline
{\tt Mean Regret} & \textbf{\underline{294.6}} & 322.4 & 710.6 & 436.7 & 1805.6 & 453.5 & 600.8 & 1000.0
\\ \hline
{\tt Std. Dev.} & 301.3 & 352.5 & \textbf{\underline{118.0}} & 168.7 &	126.6 &	147.8 &	126.3 & 1637.9
\\ \hline
{\tt Quantile .10} & 139.8 & 93.3 & 565.4 & 288.1 & 1653.3 & 342.8 & 464.1 & \textbf{\underline{34.9}}
\\  \hline
{\tt Quantile .25} & 148.7 & 116.4 & 609.8 & 335.9 & 1713.9 & 374.8 & 792.9 & \textbf{\underline{30.2}}
\\  \hline
{\tt Quantile .50} & 176.9 & 166.1 & 695.1 & 411.0 & 1789.0 & 419.6 & 592.2 & \textbf{\underline{77.4}}
\\ \hline
{\tt Quantile .75} & \textbf{\underline{237.4}} & 273.8 & 784.9 & 468.3 & 1898.1 & 483.9 & 662.3 & 1050.0
\\ \hline
{\tt Quantile .90} & 919.0 & 1064.9 & 875.6 & 610.0 & 1970.0 & \textbf{\underline{578.0}} & 739.0 & 4920.6
\\ \hline
{\tt Quantile .95} & 1183.3 & 1112.1 & 916.6 & 682.5 & 2035.3 & \textbf{\underline{644.9}} & 789.5 & 5042.0
\\ \hline
\end{tabular}
}
\end{center}
\end{table}

\begin{table}[h]
\caption{Statistics of the final cumulative regret in Figure \ref{fig:regret all 3}(c). The best in each row is highlighted. }
\label{table:Exponential_ID=17}
\begin{center}
{\small
\begin{tabular}{|p{2.3cm}|p{1cm}|p{0.8cm}|p{1.0cm}|p{0.7cm}|p{1.1cm}|p{0.8cm}|p{0.9cm}|p{0.8cm}|}
\hline
\textbf{Algorithm} & \textbf{RBMLE} & \textbf{VIDS} & \textbf{KLUCB} & \textbf{TS} & \textbf{UCB} & \textbf{MOSS} & \textbf{BUCB} & \textbf{KG}  
\\ \hline
{\tt Mean Regret} & \textbf{\underline{195.2}} & 215.9 & 339.1 & 221.3 & 1815.8 & 462.0 & 298.8 & 1460.3
\\ \hline
{\tt Std. Dev.} & 140.2 & 425.2 & 53.6 & 60.3 & 69.2 &	53.3 & \textbf{\underline{45.9}} & 2035.8
\\ \hline
{\tt Quantile .10} & 140.9 & 43.5 & 264.9 & 159.6 & 1729.3 & 402.8 & 247.8 & \textbf{\underline{26.7}}
\\  \hline
{\tt Quantile .25} & 153.1 & 55.9 & 301.4 & 176.9 & 1776.9 & 428.6 & 263.5 & \textbf{\underline{33.7}}
\\  \hline
{\tt Quantile .50} & 166.2 & 70.5 & 335.7 & 211.3 & 1818.0 & 456.7 & 297.7 & \textbf{\underline{58.3}}
\\ \hline
{\tt Quantile .75} & 188.0 & \textbf{\underline{94.1}} & 373.1 & 248.6 & 1863.7 & 480.2 & 326.5 & 3249.7
\\ \hline
{\tt Quantile .90} & \textbf{\underline{225.8}} & 1037.1 & 408.4 & 296.9 & 1897.8 & 532.3 & 365.8 & 4955.2
\\ \hline
{\tt Quantile .95} & \textbf{\underline{291.7}} & 1064.6 & 433.2 & 319.3 & 1934.1 & 563.1 & 383.3 & 4966.9
\\ \hline
\end{tabular}
}
\end{center}
\end{table}

\begin{table}[h]
\caption{Average computation time per decision for Bernoulli bandits, under different numbers of arms. All numbers are averaged over 100 trials with $T=10^{4}$ and in $10^{-4}$ seconds. The best in each row is highlighted.}
\label{table:Bernoulli_computation_time}
\begin{center}
{\small
\begin{tabular}{|p{2.9cm}|p{1cm}|p{0.5cm}|p{0.6cm}|p{1.0cm}|p{0.7cm}|p{0.7cm}|p{0.7cm}|p{0.7cm}|p{0.7cm}|p{0.4cm}|p{0.9cm}|p{0.8cm}|}
\hline
\textbf{\# Arms (Statistics)} & \textbf{RBMLE} & \textbf{IDS} & \textbf{VIDS} & \textbf{KLUCB} & \textbf{UCBT} & \textbf{TS} & \textbf{UCB} & \textbf{MOSS} & \textbf{BUCB} & \textbf{KG} & \textbf{KGMin} & \textbf{KGMN}     
\\ \hline
{\tt 10 (Mean)} & 1.36 & 175 & 123 & 12.8 & 1.53 & 0.225 & 0.712 & 0.895 & 0.855 & 28.7 & 0.649 & 0.453
\\ \hline
{\tt 30 (Mean)} & 3.61 & 1260 & 788 & 49.7 & 4.96 &	0.628 &	2.19 & 2.83 & 2.58 & 97.6 & 1.89 & 1.36
\\ \hline
{\tt 50 (Mean)} & 4.58 & 3630 &	1930 &	80.3 &	7.85 & 0.628 & 3.42 & 4.40 & 4.11 & 159 & 2.95 & 2.14
\\  \hline
{\tt 70 (Mean)} & 7.56 & 6660 & 3590 & 113 & 10.3 &	0.628 &	4.49 & 5.87 & 5.43 & 209 &	3.97 & 2.86
\\ \hline
{\tt 10 (Std. Err.)} & 0.236 & 54.8 & 33.1 & 1.53 &	0.586 &	0.0380 & 0.268 & 0.333 & 0.351 & 10.9 &	0.284 & 0.172
\\ \hline
{\tt 30 (Std. Err.)} & 1.30 & 458 & 232 & 17.3 & 1.52 & 0.106 & 0.646 & 0.844 & 0.714 & 29.2 & 0.557 & 0.408
\\ \hline
{\tt 50 (Std. Err.)} & 2.04 & 972 & 536 & 29.4 & 2.59 & 0.106 & 1.11 & 1.40 & 1.25 & 49.5 & 0.931 & 0.678
\\ \hline
{\tt 70 (Std. Err.)} & 2.70 & 1330 & 883 & 36.6 & 3.63 & 0.106 & 1.53 & 2.00 & 1.76 & 69.3 & 1.34 & 0.962
\\ \hline
\end{tabular}
}
\end{center}
\end{table}

\begin{table}[h]
\caption{Average computation time per decision for Gaussian bandits, under different numbers of arms. All numbers are averaged over 100 trials with $T=10^{4}$ and in $10^{-4}$ seconds. The best in each row is highlighted.}
\label{table:Gaussian_computation_time}
\begin{center}
{\small
\begin{tabular}{|p{2.9cm}|p{1cm}|p{0.8cm}|p{2.0cm}|p{1.0cm}|p{1.2cm}|p{0.7cm}|p{0.9cm}|p{0.8cm}|p{0.9cm}|}
\hline
\textbf{\# Arms (Statistics)} & \textbf{RBMLE} & \textbf{VIDS} & \textbf{KLUCB\&UCB} & \textbf{GPUCB} & \textbf{GPUCBT} & \textbf{TS} & \textbf{BUCB} & \textbf{KG} & \textbf{KG*}  
\\ \hline
{\tt 10 (Mean)} & 0.617 & 135 & 0.341 & 0.346 & 0.318 & 0.451 & 17.9 & 25.1 & 10.9
\\ \hline
{\tt 30 (Mean)} & 1.07 & 1410 & 1.12 & 1.10 & 1.08 & 1.33 & 75.2 & 103 & 21.2
\\  \hline
{\tt 50 (Mean)} & 1.49 & 3580 & 1.22 & 1.79 & 1.76 & 2.44 & 121 & 168 & 33.9
\\ \hline
{\tt 70 (Mean)} & 1.95 & 6610 & 1.67 & 2.24 & 2.22 & 3.16 & 162 & 226 & 45.9
\\ \hline
{\tt 10 (Std. Err.)} & 0.284 & 53.9 & 0.417 & 0.136 & 0.160 & 0.0425 & 6.98 & 9.37 & 2.77
\\ \hline
{\tt 30 (Std. Err.)} & 0.484 & 409 & 1.28 & 0.370 & 0.370 & 0.321 & 26.2 & 35 & 5.61
\\ \hline
{\tt 50 (Std. Err.)} & 0.686 & 866 & 2.14 & 0.563 & 0.563 & 0.562 & 42.1 & 56.1 & 9.77
\\ \hline
{\tt 70 (Std. Err.)} & 0.871 & 1290 & 2.95 & 0.755 & 0.773 & 0.774 & 58.5 & 77.6 & 15.7
\\ \hline
\end{tabular}
}
\end{center}
\end{table}

\begin{table}[h]
\caption{Average computation time per decision for Exponential bandits, under different numbers of arms. All numbers are averaged over 100 trials with $T=10^{4}$ and in $10^{-4}$ seconds. The best in each row is highlighted.}
\label{table:Exponential_computation_time}
\begin{center}
{\small
\begin{tabular}{|p{2.9cm}|p{1cm}|p{0.8cm}|p{1.0cm}|p{0.7cm}|p{0.8cm}|p{0.8cm}|p{0.9cm}|p{0.8cm}|}
\hline
\textbf{\# Arms (Statistics)} & \textbf{RBMLE} & \textbf{VIDS} & \textbf{KLUCB} & \textbf{TS} & \textbf{UCB} & \textbf{MOSS} & \textbf{BUCB} & \textbf{KG}
\\ \hline
{\tt 10 (Mean)} & 1.01 & 133 & 7.26 & 1.38 & 0.420 & 0.548 & 14.9 & 0.519
\\ \hline
{\tt 30 (Mean)} & 1.93 & 1160 & 22.8 & 3.97 & 1.20 & 1.61 & 42.6 & 1.36
\\ \hline
{\tt 50 (Mean)} & 2.97 & 3170 & 36.5 & 6.64 & 1.92 & 2.53 & 75.5 & 2.23
\\  \hline
{\tt 70 (Mean)} & 3.79 & 6430 & 53.7 & 9.30 & 2.67 & 3.59 & 102 & 3.06
\\  \hline
{\tt 10 (Std. Err.)} & 0.435 & 13.6 & 0.884 & 0.316 & 0.0980 & 0.112 & 1.55 & 0.101
\\ \hline
{\tt 30 (Std. Err.)} & 0.890 & 187 & 2.79 & 0.777 & 0.263 & 0.340 & 5.02 & 0.265
\\ \hline
{\tt 50 (Std. Err.)} & 1.24 & 447 & 5.47 & 1.20 & 0.397 & 0.498 & 10.2 & 0.456
\\ \hline
{\tt 70 (Std. Err.)} & 1.56 & 788 & 6.92 & 1.96 & 0.531 & 0.688 & 12.3 & 0.605
\\ \hline
\end{tabular}
}
\end{center}
\end{table}
\medskip
\end{document}